\documentclass[final,3p,times]{elsarticle}

\usepackage{amssymb}
\usepackage{amsmath}
\usepackage{graphicx}
\usepackage{subcaption}
\usepackage{booktabs}
\usepackage{multirow}
\usepackage{tabularx}
\usepackage{makecell}
\usepackage{adjustbox}
\usepackage{threeparttable}
\usepackage{tikz}
\usetikzlibrary{positioning, arrows.meta, shapes.geometric, fit, backgrounds}
\usepackage{xcolor}
\usepackage{pifont}
\usepackage{hyperref}
\usepackage{algorithm}
\usepackage{algpseudocode}
\usepackage{natbib}

\usepackage{caption}
\usepackage{subcaption}
\usepackage{lineno}
\usepackage{hyperref}

\hypersetup{
    colorlinks=true,      
    linkcolor=blue,       
    citecolor=green,      
    urlcolor=cyan,        
    filecolor=magenta,    
    linktoc=all           
}

\newcommand{\cmark}{\textcolor{green}{\ding{51}}} 
\newcommand{\xmark}{\textcolor{red}{\ding{55}}} 

\journal{Neural Networks}

\begin{document}

\begin{frontmatter}

\title{Point-DeepONet: Predicting Nonlinear Fields on Non-Parametric Geometries under Variable Load Conditions}

\nonumnote{Accepted for publication in \textit{Neural Networks} (2026). \\ 
DOI: \href{https://doi.org/10.1016/j.neunet.2026.108560}{10.1016/j.neunet.2026.108560}}

\author[1]{Jangseop Park}
\author[1,2]{Namwoo Kang\corref{cor1}}

\cortext[cor1]{Corresponding author. Email: nwkang@kaist.ac.kr \\
Code and dataset available at \url{https://github.com/jangseop-park/Point-DeepONet}.}

\affiliation[1]{
    organization={Cho Chun Shik Graduate School of Mobility, Korea Advanced Institute of Science and Technology},
    city={Daejeon},
    postcode={34051},
    country={South Korea}
}

\affiliation[2]{
    organization={Narnia Labs},
    city={Daejeon},
    postcode={34051},
    country={South Korea}
}

\begin{abstract}
Nonlinear structural analyses in engineering often require extensive finite element simulations, limiting their applicability in design optimization and real-time control. Conventional deep learning surrogates often struggle with complex, non-parametric three-dimensional (3D) geometries and directionally varying loads. This work presents Point-DeepONet, an operator-learning-based surrogate that integrates PointNet into the DeepONet framework to learn a mapping from non-parametric geometries and variable load conditions to physical response fields. By leveraging PointNet to learn a geometric representation from raw point clouds, our model circumvents the need for manual parameterization. This geometric embedding is then synergistically fused with load conditions within the DeepONet architecture to accurately predict three-dimensional displacement and von Mises stress fields. Trained on a large-scale dataset, Point-DeepONet demonstrates high fidelity, achieving a coefficient of determination (R²) reaching 0.987 for displacement and 0.923 for von Mises stress. Furthermore, to rigorously validate its generalization capabilities, we conducted additional experiments on unseen, randomly oriented load directions, where the model maintained exceptional accuracy. Compared to nonlinear finite element analyses that require about 19.32 minutes per case, Point-DeepONet provides predictions in mere seconds—approximately 400 times faster—while maintaining excellent scalability. These findings, validated through extensive experiments and ablation studies, highlight the potential of Point-DeepONet to enable rapid, high-fidelity structural analyses for complex engineering workflows.
\end{abstract}

\begin{keyword}
Deep operator network (DeepONet) \sep PointNet \sep Surrogate modeling \sep Nonlinear structure analysis \sep Non-parametric 3D geometry
\end{keyword}
\end{frontmatter}

\section{Introduction}
\label{sec:Introduction}

The simulation of complex physical systems is fundamental in engineering and scientific research, enabling the analysis and prediction of phenomena such as structural deformations, fluid dynamics, and thermal processes. High-fidelity numerical methods like the finite element method (FEM) have traditionally been employed to achieve accurate results in these simulations. However, when dealing with nonlinear material behaviors, complex three-dimensional (3D) geometries, and varying boundary conditions, these methods become computationally intensive and time-consuming. In particular, nonlinear analyses require iterative solutions of nonlinear equations and fine discretizations to capture localized effects, which further increase computational demands.

The significant computational demands inherent in simulating complex physical systems present major challenges for applications requiring rapid evaluations across numerous scenarios, such as real-time control, design optimization, uncertainty quantification, and inverse problem solving. Traditional surrogate models based on neural networks often predict solutions only at predefined mesh nodes or within limited parameter ranges. Consequently, even minor variations in inputs—such as changes in loads, boundary conditions, or domain geometry—typically require expensive retraining or transfer learning processes. This limitation underscores the urgent need for more efficient surrogate models that can accurately approximate the behavior of complex physical systems while reducing computational costs and enhancing the ability to generalize to previously unseen cases.

To meet these requirements, various deep learning architectures have been explored for surrogate modeling in computational mechanics. Among them, Convolutional Neural Networks (CNNs) have been widely adopted due to their ability to capture spatial hierarchies in data. In the context of surrogate modeling for physical simulations, CNNs are typically applied to regular grid data or transformed representations of computational domains. For example, some studies have converted irregular geometries into regular grids or images to leverage CNN architectures for flow predictions and structural analyses \citep{SUN2020112732, bao2022physics}. However, CNNs face significant limitations when dealing directly with unstructured meshes and irregular geometries typical in FEM discretizations. Their grid-based nature makes it challenging to process complex 3D geometries without introducing interpolation errors or losing geometric fidelity. Moreover, representing intricate 3D domains with regular grids often results in high-dimensional inputs and computational inefficiencies.

To address the challenges posed by irregular geometries, PointNet \citep{qi2017pointnet} was introduced. PointNet processes point cloud data directly, treating sets of 3D points as inputs and extracting global features through shared multilayer perceptrons (MLPs) combined with symmetric aggregation functions (e.g., max pooling). This approach captures the geometric information of complex shapes without relying on structured grids or mesh connectivity. However, while PointNet effectively encodes global geometric features, it does not explicitly model local or hierarchical structures.

PointNet++ \citep{qi2017pointnet++} extends PointNet by introducing a hierarchical feature learning architecture. By recursively applying PointNet on nested subsets of points and using metric space distances to group points into local regions, PointNet++ learns local features at multiple scales. This hierarchical approach enhances its ability to represent intricate geometries and makes PointNet++ more effective for tasks such as semantic segmentation and object part segmentation.

Graph neural networks (GNNs) have also been proposed to leverage the connectivity inherent in mesh-based representations. Graph neural networks (GNNs) treat mesh nodes as graph vertices and elements or edges as connections, capturing both geometric and topological information \citep{gilmer2017neural, battaglia2018relational, park2023bayesian}. They have shown success in modeling variable geometries by facilitating information propagation among connected nodes and have been applied to various engineering problems, including structural analysis and fluid dynamics. However, as the number of mesh nodes grows, GNNs face reduced computational efficiency, limiting their scalability for large-scale problems.

Physics-informed neural networks (PINNs) have been introduced to solve partial differential equations (PDEs) by embedding governing physical laws directly into the loss function \citep{raissi2019physics}. Physics-informed neural networks (PINNs) leverage both data and physics to predict solution fields without requiring extensive labeled datasets. Despite their promise in accurately solving various PDEs, PINNs are typically tailored to single instances of geometries and boundary conditions. Changes in these factors often require retraining, reducing PINNs' flexibility and scalability in scenarios where multiple geometries or dynamic loading conditions must be considered \citep{lu2021deepxde, zhu2019physics, sirignano2018dgm}.

To address the limitations of these methods, operator learning has emerged as a promising new direction. Operator learning aims to approximate governing operators via neural networks \citep{kovachki2023neural}, thus enabling broader applicability without retraining for every new parameter set. A prominent operator learning framework, the fourier neural operator (FNO), introduced by \citet{li2020fourier}, encodes input functions through multiple Fourier layers. Each layer applies a Fast Fourier Transform (FFT) and filters out high-frequency modes. The fourier neural operator (FNO) and its variants have been successfully applied to various differential equations \citep{li2020fourier} and elastoplastic deformation problems in mechanics \citep{li2023fourier}. However, because FNOs are typically formulated on regular grids, their direct applicability to irregular or complex geometries is limited. Although extensions to handle irregular domains exist \citep{li2023fourier}, they often require additional preprocessing or mesh mapping.

Deep operator networks (DeepONets), developed by \citet{lu2021learning}, mark a significant advancement in operator learning. Inspired by the universal approximation theorem for operators \citep{chen1995universal}, deep operator networks (DeepONets) map unseen parametric functions to solution spaces for both linear and nonlinear PDEs. Since their inception, DeepONets have been applied to materially nonlinear solid mechanics \citep{koric2024deep, lu2023deep}, fracture \citep{goswami2022physics}, aerodynamics \citep{zhao2023learning}, acoustics \citep{xu2023training}, heat transfer \citep{sahin2024deep, koric2023data}, seismology \citep{haghighat2024en-deeponet}, and multiscale modeling of elastic and hyperelastic materials \citep{yin2022interfacing}. Improvements such as physics-informed DeepONet \citep{wang2021learning} enhance prediction accuracy by incorporating PDE constraints, while s-DeepONet \citep{he2024sequential} handles time-dependent inputs. Recent work has integrated ResUNet architectures into DeepONet \citep{he2023novel}, enabling the handling of complex and highly disparate two-dimensional input geometries under parametric loads and elastoplastic material behavior.

Despite these advances, most DeepONet-based approaches remain limited to regular grids, parameterized geometries, or simpler variations in loading conditions. To address these gaps, \citet{he2024geom} introduced Geom-DeepONet, a point-cloud-based DeepONet capable of field predictions on 3D parameterized geometries. By processing point clouds directly, Geom-DeepONet enables predictions on complex 3D shapes. However, it still relies on parameterized geometries and, although it considers variations in load magnitude, does not handle more intricate changes such as load directionality. Consequently, its applicability to non-parametric geometries and directionally varying load conditions remains limited.

Table~\ref{tab:method_comparison} provides a comparative overview of existing deep learning-based surrogate modeling approaches. CNNs rely on structured grids and cannot handle non-parametric geometries or directional load changes, while PointNet and GNNs operate on non-parametric domains but struggle with fully varying load conditions and show declining efficiency as complexity grows. PINNs and FNOs handle load variations yet remain constrained by single-instance setups or regular grids. DeepONet improves efficiency and resolution but still uses parameterized inputs, whereas Geom-DeepONet extends capabilities to point clouds and load magnitude variations, though not load directions. In contrast, our proposed Point-DeepONet accommodates non-parametric 3D geometries, varying load magnitude and direction, maintains high computational efficiency, and produces high-resolution predictions directly on the original finite element mesh.

\begin{table}[h]
    \centering
    \begin{threeparttable}
        \caption{Comparison of deep learning surrogate models for mechanical engineering based on geometry handling, load variations, high-resolution inference, and computational efficiency.}
        \label{tab:method_comparison}
        \begin{tabularx}{\linewidth}{lccc c X}
            \toprule
            Method & 
            \makecell{Handles \\ non-parametric \\ geometries} & 
            \makecell{Handles varying \\ load conditions} & 
            \makecell{High-resolution \\ inference} &
            \makecell{Computational \\ efficiency} &
            Reference \\
            \midrule
            CNNs & \xmark & \xmark & \xmark & \textcolor{red}{\(\blacktriangledown\)} & \citep{SUN2020112732, bao2022physics} \\
            PointNet, PointNet++ & \cmark & \xmark & \xmark & \textcolor{red}{\(\blacktriangledown\)} & \citep{qi2017pointnet, qi2017pointnet++, kashefi2021point} \\
            GNNs & \cmark & \xmark & \xmark & \textcolor{red}{\(\blacktriangledown\)} & \citep{gilmer2017neural, battaglia2018relational, park2023bayesian} \\
            PINNs & \xmark & \cmark & \xmark & \textcolor{red}{\(\blacktriangledown\)} & \citep{raissi2019physics} \\
            FNO & \xmark & \cmark & \xmark & \textcolor{red}{\(\blacktriangledown\)} & \citep{li2020fourier, li2023fourier} \\
            DeepONet & \xmark & \cmark & \cmark & \textcolor{green}{\(\blacktriangle\)} & \citep{lu2021learning} \\
            Geom-DeepONet & \xmark & \cmark & \cmark & \textcolor{green}{\(\blacktriangle\)} & \citep{he2024geom} \\
            \makecell[l]{Point-DeepONet \\ (Our proposed)} & \cmark & \cmark & \cmark & \textcolor{green}{\(\blacktriangle\)} & - \\
            \bottomrule
        \end{tabularx}
        \begin{tablenotes}
            \footnotesize
            \item \cmark: Capable of performing the task, \xmark: Limited or unable to perform the task, \(\blacktriangle\): Higher efficiency, \(\blacktriangledown\): Lower efficiency.
        \end{tablenotes}
    \end{threeparttable}
\end{table}

Notably, while recent methods like Geom-DeepONet \citep{he2024geom} have advanced the field by processing point clouds, they are primarily designed for \textit{parameterized} geometries where design variables (e.g., length, thickness, radius) are explicitly known and fed into the branch network. This approach is powerful for design optimization within a defined parameter space but is not applicable to \textit{non-parametric} geometries, such as those resulting from topology optimization, which lack simple descriptive parameters. Furthermore, its application has been limited to variations in load magnitude. The limitations of existing methods motivated the development of Point-DeepONet. Our work addresses these critical gaps. First, by integrating PointNet, it learns a comprehensive geometric representation directly from the point cloud, eliminating the need for any parameterization. Second, it accommodates variations in both load magnitude and direction, making it a more generalized framework for real-world engineering analysis. This fundamental difference in problem definition and architectural design is why a direct quantitative comparison is challenging, and it highlights the unique contribution of our work in tackling a broader class of engineering problems.

In this work, we present \textit{Point-DeepONet}, a novel neural network architecture that seamlessly integrates PointNet into the DeepONet framework. This integration enables fully nonlinear analyses on non-parametric 3D geometries under both magnitude- and direction-varying load conditions, without relying on explicit parameterizations or mesh connectivity. By incorporating PointNet into the branch network, our model encodes complex geometric information from point clouds, facilitating generalization to a wide variety of shapes. Additionally, an auxiliary branch network is introduced to incorporate global physical parameters such as boundary conditions and mass. Combining these global inputs with the geometric encoding from PointNet allows our model to capture local geometric details and the influences of varying boundary conditions, enabling accurate predictions in nonlinear analyses.

To the best of our knowledge, this is the first operator learning-based surrogate model capable of performing fully nonlinear analyses on non-parametric three-dimensional domains while accommodating both magnitude and directional variations in load conditions. Our proposed \textit{Point-DeepONet} offers several key advantages:

\begin{enumerate}
    \item \textbf{Flexible handling of non-parametric 3D geometries:} By integrating PointNet, our approach directly processes arbitrary three-dimensional domains represented as point clouds. This flexibility eliminates the need for parameterizations or regular grids, thereby facilitating the modeling of complex, non-parametric geometries.
    \item \textbf{Versatile adaptation to varying load conditions:} Unlike existing methods such as Geom-DeepONet \citep{he2024geom}, our model seamlessly accommodates both changes in load magnitude and direction. As a result, it can be readily applied to a broad range of engineering scenarios with non-uniform and evolving loading conditions.
    \item \textbf{High-fidelity field predictions at the source mesh:} Our model predicts essential field variables, including displacement components ($u_x$, $u_y$, $u_z$) and von Mises stress, directly on the original finite element mesh. This approach ensures precise, high-resolution outputs and minimizes the need for additional interpolation or re-processing.
    \item \textbf{Reduced computational overhead:} By learning operator mappings that generalize across multiple configurations, \textit{Point-DeepONet} mitigates the necessity for repetitive mesh generation and re-analysis. Consequently, the computational cost and complexity associated with exploring diverse design spaces are significantly lowered.
    \item \textbf{Accelerated decision-making processes:} Through its ability to capture intricate geometric details and boundary condition variations, our framework enables swift evaluations of numerous design alternatives. This expedited analysis supports more informed decision-making in shorter timeframes, enhancing overall engineering productivity.
    \item \textbf{Robust scalability with increasing data and complexity:} As demonstrated in subsequent sections, our model's performance and efficiency improve with larger datasets and more sampling points. This scalability ensures that \textit{Point-DeepONet} remains effective even as problem sizes and design spaces continue to grow, making it a durable solution for evolving engineering challenges.
\end{enumerate}

Our work is inspired by \citet{he2023novel}, who integrated a ResUNet-based branch network into DeepONet for complex elastoplastic deformations in two-dimensional structures. By extending their concept to three dimensions and leveraging PointNet, we significantly advance the capability to model nonlinear phenomena on non-parametric 3D geometries under complex, directionally varying load conditions.

In Section~\ref{sec:Methodology}, we present the methodology, including data generation, data preprocessing, and the proposed neural network architecture (Point-DeepONet) that integrates PointNet into DeepONet for handling nonlinear material behaviors. Section~\ref{sec:results_discussion} then reports and discusses the numerical experiment results, demonstrating the effectiveness of the proposed model in predicting displacement and von Mises stress fields for nonlinear problems. Finally, Section~\ref{sec:Conclusion_futurework} concludes the paper by summarizing the main findings, addressing current limitations, and suggesting avenues for future work.

\section{Methodology}
\label{sec:Methodology}


\subsection{Dataset Generation}
\label{subsec:DatasetGeneration}

We utilize the DeepJEB dataset \citep{hong2025deepjeb}, a synthetic dataset specifically designed for 3D deep learning applications in structural mechanics, focusing on jet engine brackets. This dataset includes various bracket geometries subjected to different load cases—vertical, horizontal, and diagonal—providing a diverse range of scenarios to train and evaluate deep learning models for predicting field values. While the original DeepJEB dataset offers solutions from linear static analyses, in this study we extend its applicability by performing our own nonlinear static finite element analyses to predict displacement fields ($u_x$, $u_y$, $u_z$) and von Mises stress under varying geometric and loading conditions.

Finite element analyses (FEA) are conducted using Altair OptiStruct \citep{altair2023optistruct} to simulate the structural response under nonlinear static conditions. Each bracket geometry is discretized using second-order tetrahedral elements with an average element size of 2~mm, enhancing the precision of the displacement and stress predictions. The material properties for the brackets are based on Ti–6Al–4V, specified with a density of $4.47 \times 10^{-3}$~g/mm$^3$, a Young's modulus ($E$) of 113.8~GPa, and a Poisson's ratio ($\nu$) of 0.342, representing realistic behavior under the applied loads.

\begin{figure}[!h]
    \centering
    \begin{subfigure}{0.25\textwidth}
        \centering
        \includegraphics[width=\linewidth]{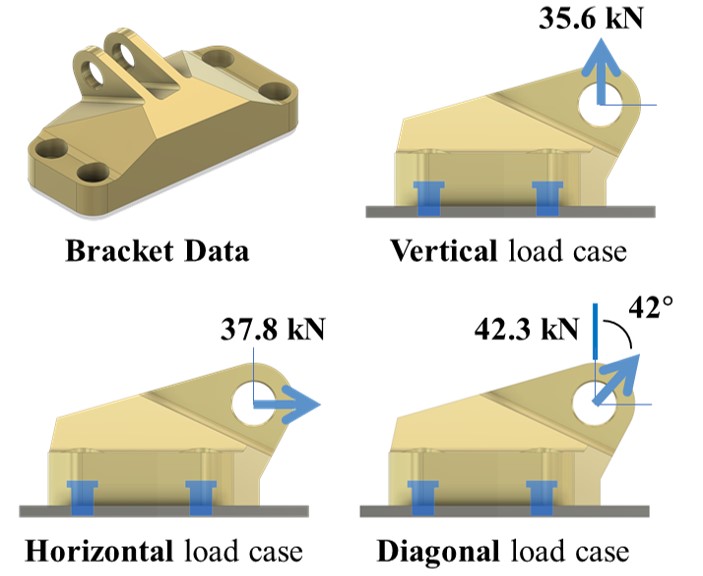}
        \caption{}
        \label{fig:figure_1_a}
    \end{subfigure}
    \begin{subfigure}{0.29\textwidth}
        \centering
        \includegraphics[width=\linewidth]{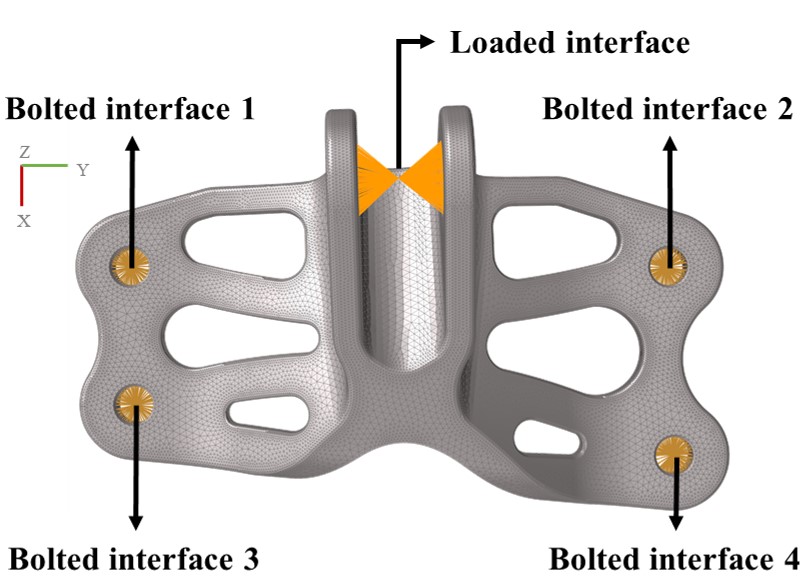}
        \caption{}
        \label{fig:figure_1_b}
    \end{subfigure}%
    \begin{subfigure}{0.24\textwidth}
        \centering
        \includegraphics[width=\linewidth]{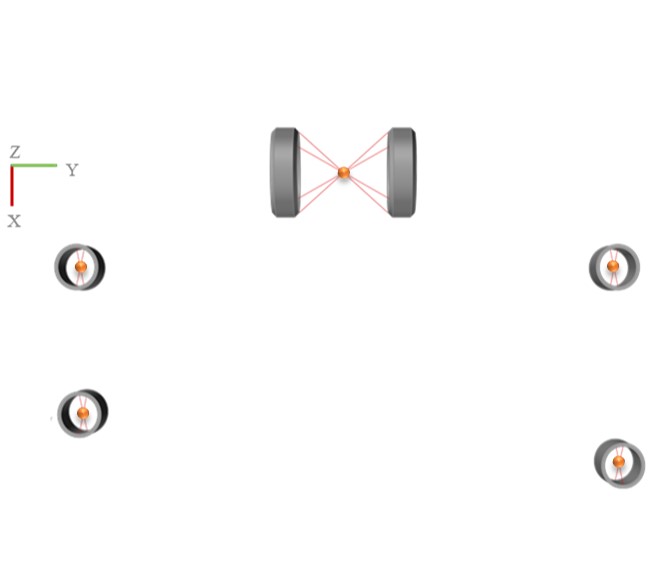}
        \caption{}
        \label{fig:figure_1_c}
    \end{subfigure}%
    \caption{Illustration of the nonlinear static analysis setup for the bracket under different load cases. (a) Bracket geometry and load direction overview: vertical, horizontal, and diagonal load cases. (b) Identification of bolted and loaded interfaces in the bracket. (c) Schematic of boundary conditions and constraints applied in the analysis.}
    \label{fig:figure_1}
\end{figure}

An elastic–plastic material model with linear isotropic hardening is employed to capture the nonlinear response, characterized by a yield stress of 227.6~MPa and a hardening modulus of 355.56~MPa. The nonlinear analysis settings include a maximum iteration limit of 10 and a convergence tolerance of 1\%, ensuring accurate simulation of the structural response to complex loading conditions. Figure~\ref{fig:figure_1} illustrates the nonlinear static analysis setup, showing the bracket geometry and load directions (a), bolted and loaded interfaces (b), and the boundary conditions and constraints applied in the analysis (c).

\begin{figure}[!h]
    \centering    
    \begin{subfigure}[b]{0.4\textwidth}
        \centering
        \includegraphics[width=\linewidth]{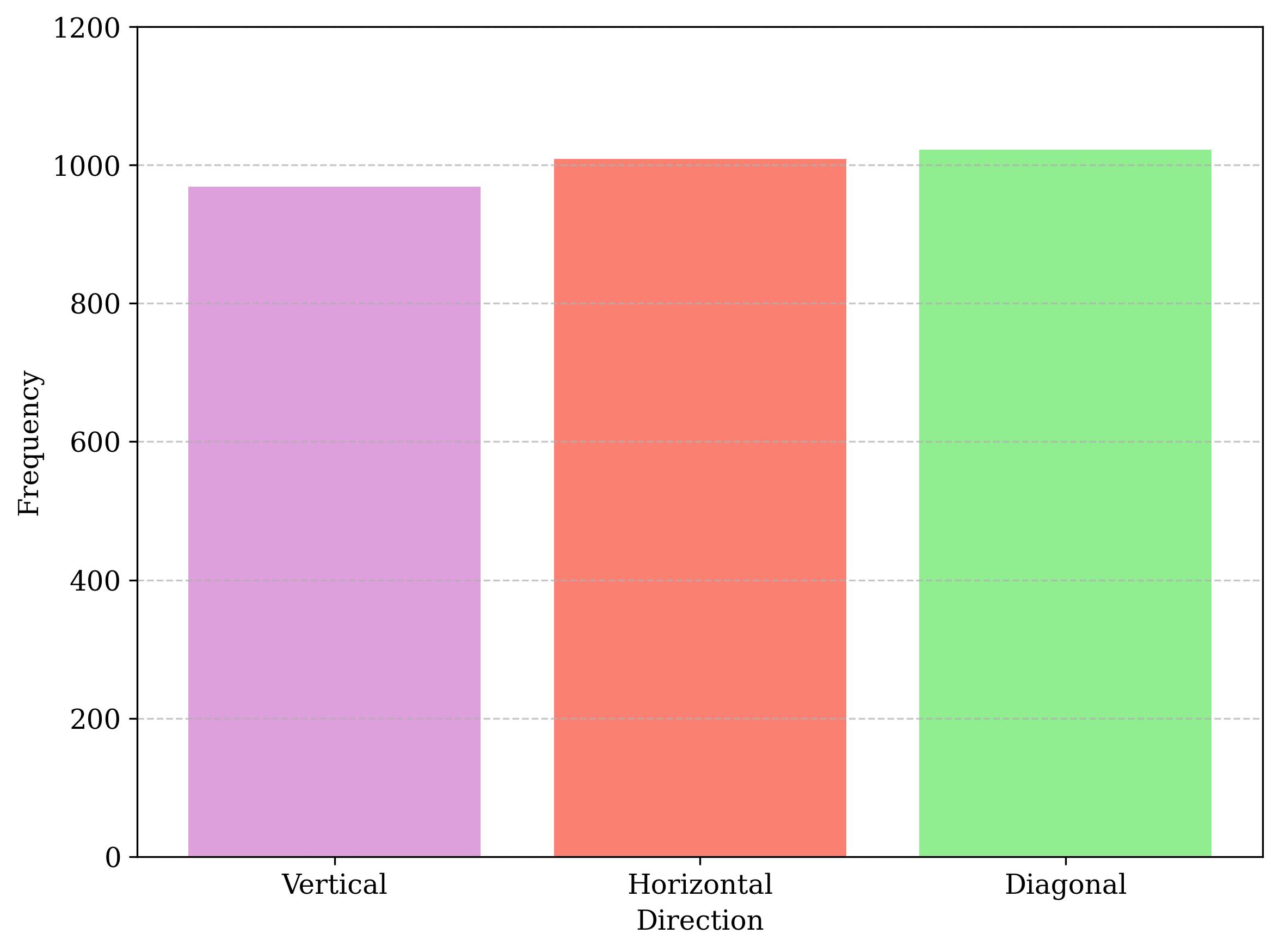}
        \caption{}
        \label{fig:figure_2_a}
    \end{subfigure}
    \begin{subfigure}[b]{0.4\textwidth}
        \centering
        \includegraphics[width=\linewidth]{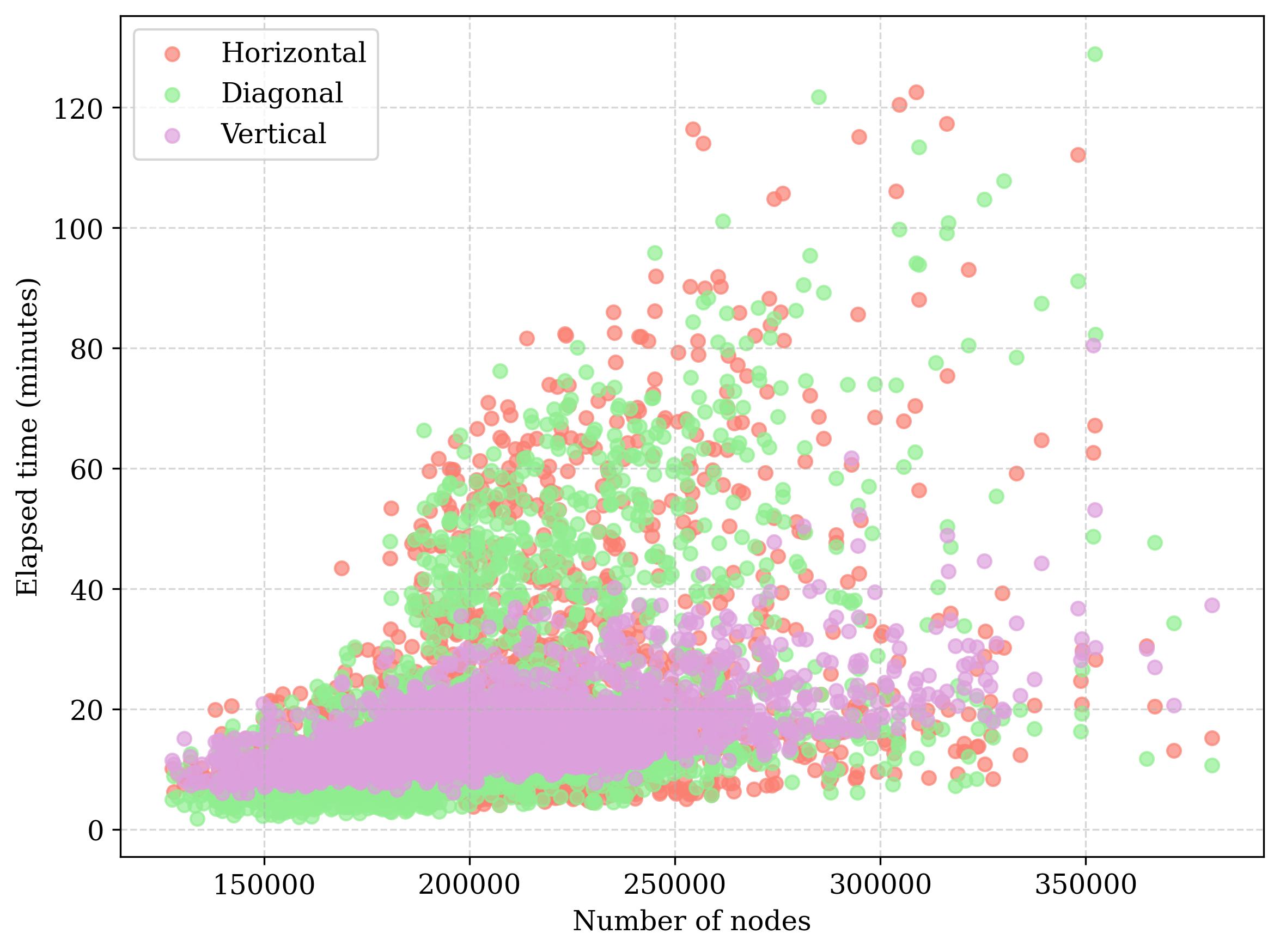}
        \caption{}
        \label{fig:figure_2_b}
    \end{subfigure}
    \caption{Visualization of dataset distribution by load direction. (a) Frequency distribution of load cases across vertical, horizontal, and diagonal directions. (b) Scatter plot showing elapsed time versus node count for vertical, horizontal, and diagonal load cases.}
    \label{fig:figure_2}
\end{figure}

The distribution of the dataset across load case directions and the FE analysis time are illustrated based on dataset size. Figure~\ref{fig:figure_2_a} presents the dataset size in terms of direction-specific distributions (vertical, horizontal, diagonal) and unique geometries. The total dataset size used in this study is 3,000, with 969 vertical, 1,009 horizontal, and 1,022 diagonal load cases. There are 1,785 unique geometries across these cases. Each bar corresponds to a mixed dataset size, with the count in parentheses representing the unique geometries, clarifying how dataset composition varies with respect to load case directions. Furthermore, in Section~\ref{subsec:influence_dataset}, we investigate the influence of dataset size by progressively reducing it and comparing the resulting performance and training time of the deep learning models.

The computational time for each load case direction varies due to differences in structural responses. The experiments were conducted on a system equipped with an AMD EPYC 9654 processor featuring 96 cores and 192 threads, ensuring that the analysis tasks could be efficiently distributed across multiple parallel processing units. As shown in Figure~\ref{fig:figure_2_b}, for vertical load cases, the elapsed time ranges from a minimum of 6.07~minutes to a maximum of 80.45~minutes, with an average time of 16.23~minutes. Horizontal load cases show a broader time range, from 3.87~minutes to 122.57~minutes, with a mean of 20.89~minutes. Diagonal load cases, similarly complex, exhibit times between 1.83~minutes and 128.85~minutes, averaging 20.84~minutes. Overall, the entire dataset's elapsed time spans from 1.83~minutes to 128.85~minutes, with a mean time of 19.32~minutes. This distribution of analysis times highlights the computational complexity associated with different load cases and provides insight into the variations in processing time based on dataset composition.

\subsection{Data Preprocessing}
\label{subsec:DataPreprocessing}

\begin{figure}[!h]
    \centering
    \begin{subfigure}[b]{0.4\textwidth}
        \centering
        \includegraphics[width=\linewidth]{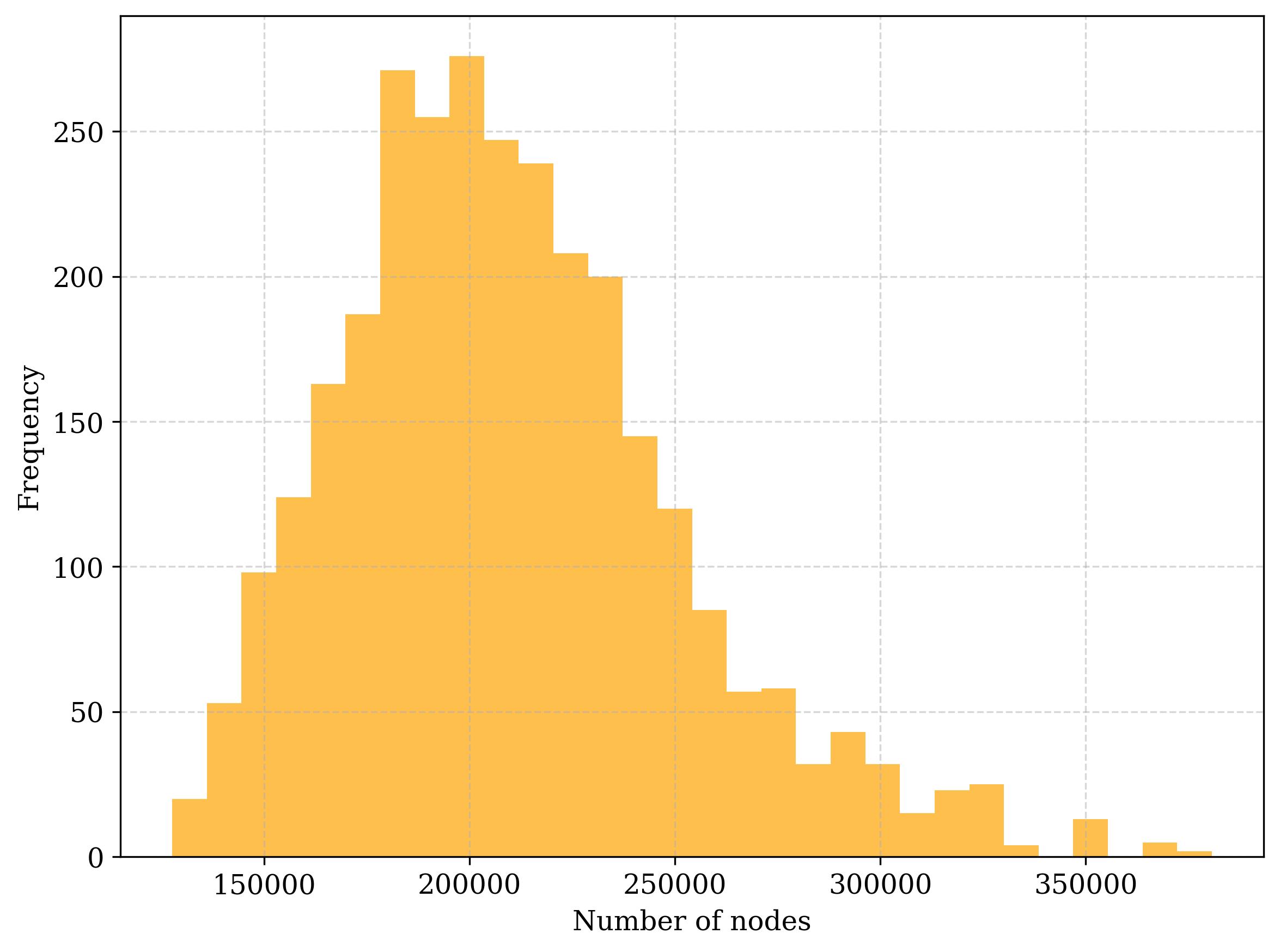}
        \caption{}
        \label{fig:figure_3_a}
    \end{subfigure}
    \begin{subfigure}[b]{0.4\textwidth}
        \centering
        \includegraphics[width=\linewidth]{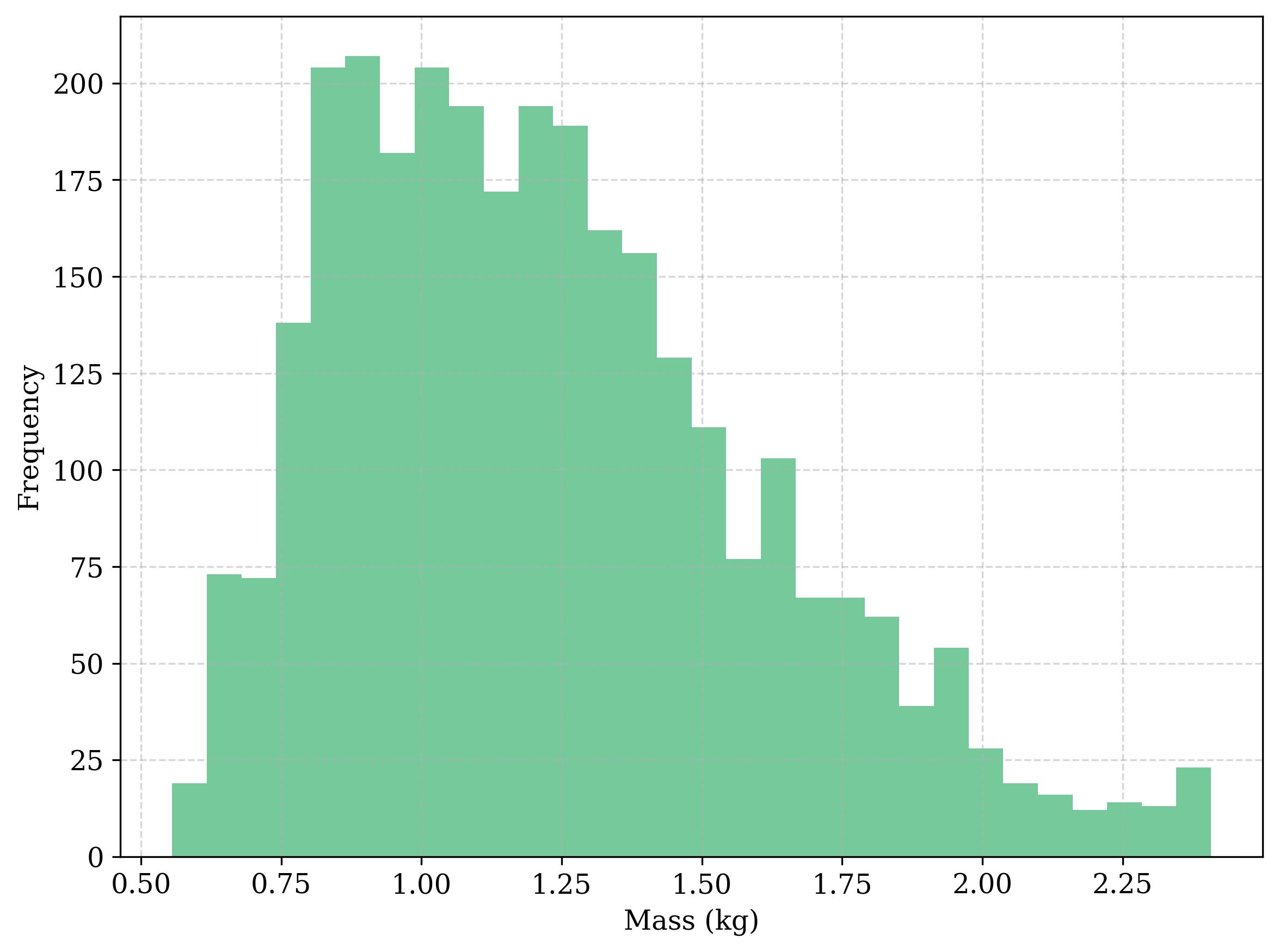}
        \caption{}
        \label{fig:figure_3_b}
    \end{subfigure}
    \caption{Distribution of structural properties in the dataset: (a) Node count distribution, representing the complexity of the mesh geometries used in FEA analysis; (b) Mass distribution, capturing the range of structural weights across different bracket geometries.}
    \label{fig:figure_3}
\end{figure}

The dataset used in this study comprises a range of jet engine bracket geometries with varying structural properties and masses, as illustrated in Figure~\ref{fig:figure_3}. Figure~\ref{fig:figure_3_a} shows the distribution of node counts, while Figure~\ref{fig:figure_3_b} presents the mass distribution across all geometries. This distribution provides insights into the diversity of the dataset, with the number of nodes ranging from 127,634 to 380,714 and the mass spanning from 0.56~kg to 2.41~kg, as summarized in Table~\ref{tab:table_2}. The average values for the dataset's structural properties are 209,974 nodes, 787,658 edges, and a mass of 1.23~kg.

While Point-DeepONet demonstrates a significant speedup in analysis, it is important to consider the computational overhead of preprocessing steps, specifically the generation of the signed distance function (SDF). The SDF for each geometry is computed only once as a one-time setup cost. To provide a clear perspective on its impact, Table~\ref{tab:cost_comparison} compares the average time required for the initial FEA simulation, one-time SDF generation, and a single inference pass of our trained model.

\begin{table}[h]
    \centering
    \caption{Quantitative comparison of computational time per geometry.}
    \label{tab:cost_comparison}
    \begin{tabular}{lc}
        \toprule
        Task & Average Time \\
        \midrule
        Nonlinear FEA Simulation & \textasciitilde 19.32 minutes \\
        SDF Generation (One-time Preprocessing) & \textasciitilde 45 seconds \\
        Point-DeepONet Inference (Full Mesh) & \textasciitilde 3.0 seconds \\
        \bottomrule
    \end{tabular}
\end{table}

As shown in Table~\ref{tab:cost_comparison}, the time required for SDF generation is negligible compared to the full FEA simulation. In practical applications such as design optimization or uncertainty quantification, where a single geometry is evaluated under numerous load conditions, the one-time SDF cost is quickly amortized. The nearly 400-fold speedup of our model refers to the inference stage, which is the critical bottleneck in such iterative analyses. Therefore, the SDF preprocessing does not undermine the overall computational efficiency of the Point-DeepONet framework.

\begin{table}[!h]
    \centering
    \caption{Summary of FEA model characteristics, showing the minimum, maximum, and average values for node count, edge count, cell count, and mass across all bracket geometries. These metrics provide insights into the computational complexity and physical properties of the dataset.}
    \label{tab:table_2}
    \begin{tabularx}{\linewidth}{Xccc}
        \toprule
        Metric & Minimum & Maximum & Average \\
        \midrule
        Number of nodes & 127,634 & 380,714 & 209,974 \\
        Number of edges & 468,708 & 1,453,872 & 787,658 \\ 
        Number of cells & 78,118 & 242,312 & 131,276 \\ 
        Mass (kg) & 0.56  & 2.41 & 1.23 \\
        \bottomrule
    \end{tabularx}
\end{table}

To facilitate effective model training and evaluation, the dataset was divided into training and validation subsets, with 80\% allocated for training and 20\% reserved for validation. This approach ensures that the model is trained on a representative variety of bracket geometries while maintaining a separate set for evaluating generalization performance.

\begin{figure}[!h]
    \centering
    \begin{subfigure}[b]{0.4\textwidth}
        \centering
        \includegraphics[width=\textwidth]{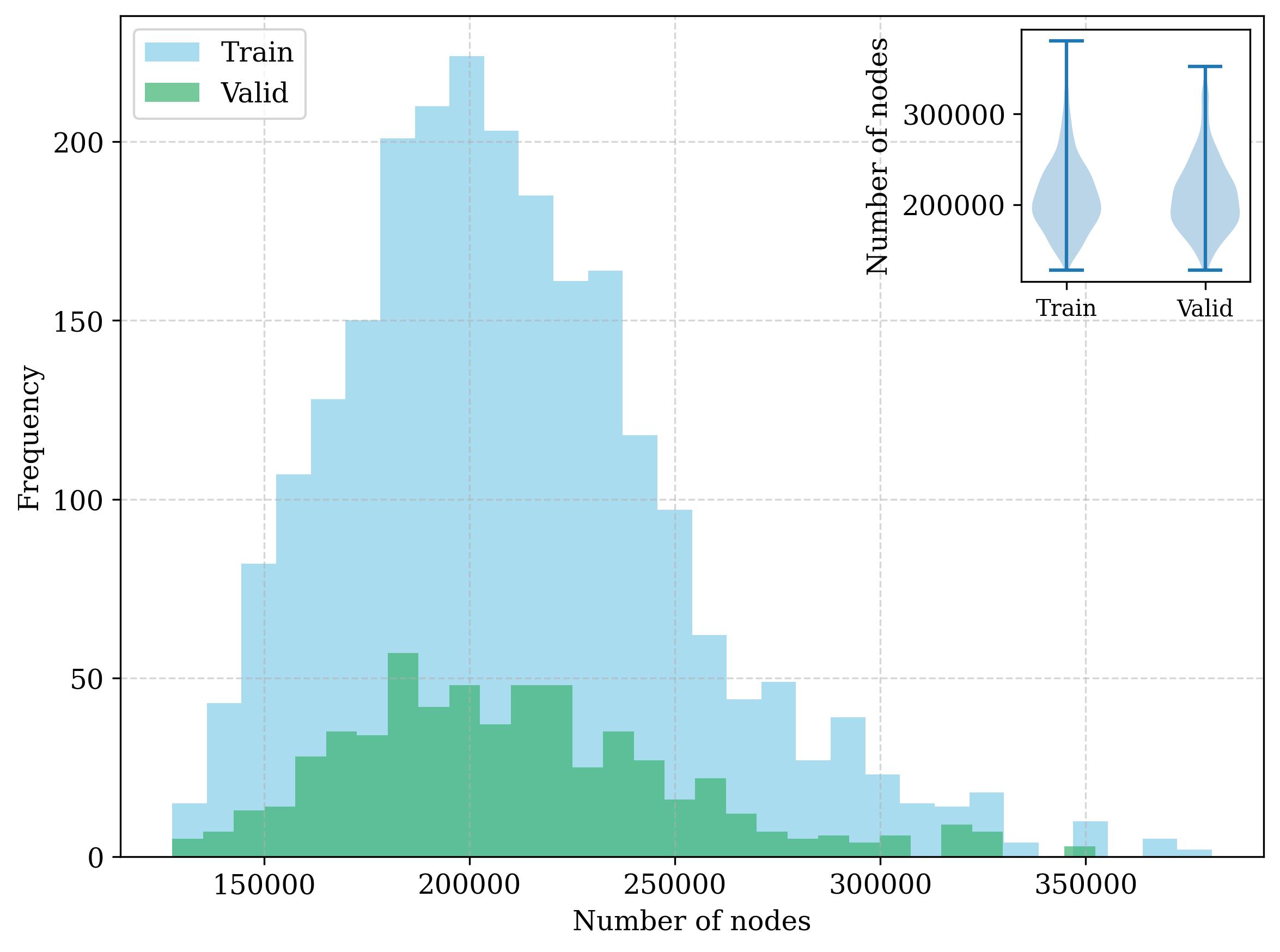}
        \caption{}
        \label{fig:figure_4_a}
    \end{subfigure}
    \begin{subfigure}[b]{0.4\textwidth}
        \centering
        \includegraphics[width=\textwidth]{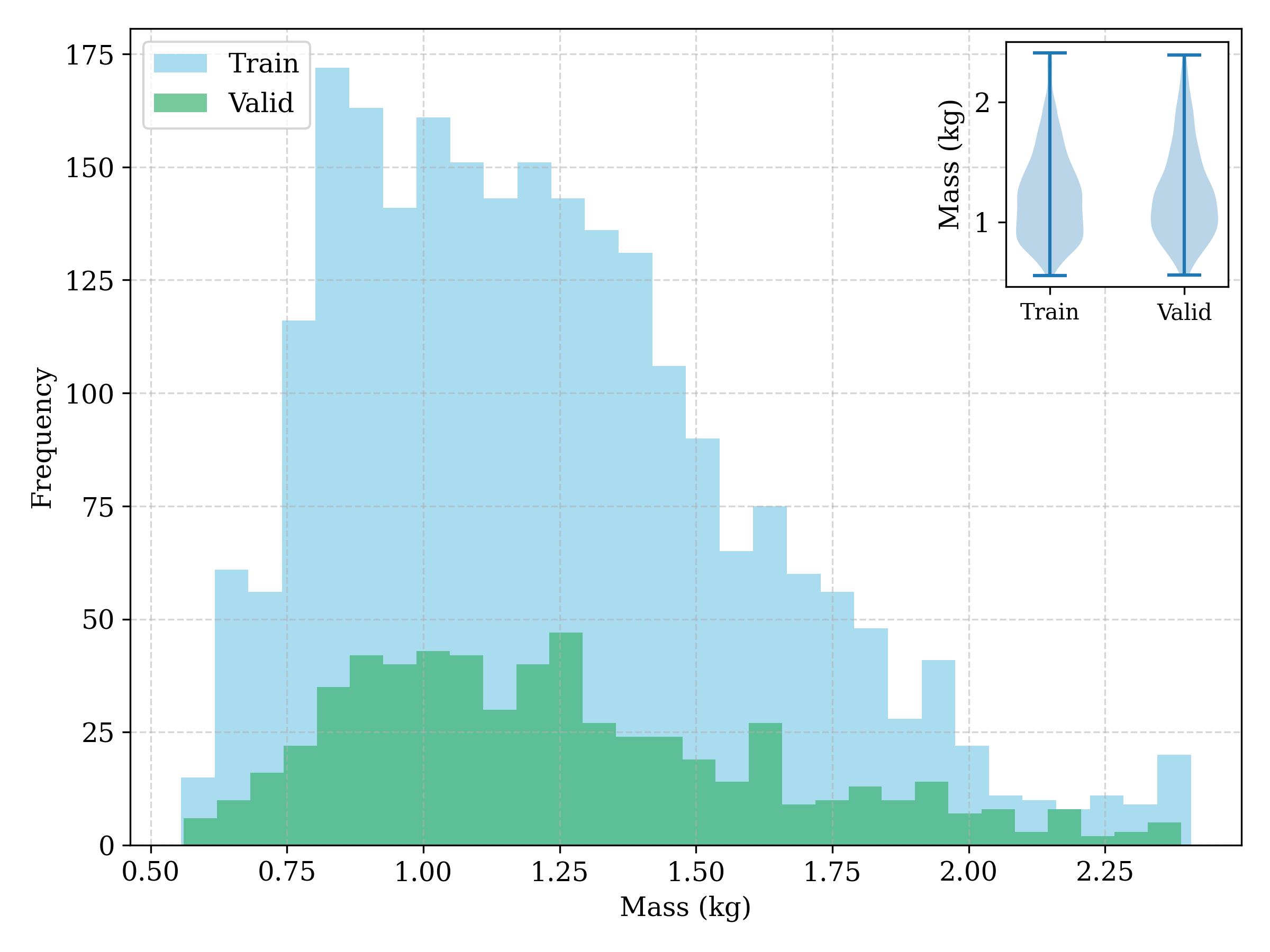}
        \caption{}
        \label{fig:figure_4_b}
    \end{subfigure}
    \begin{subfigure}[b]{1\textwidth}
        \centering
        \includegraphics[width=\textwidth]{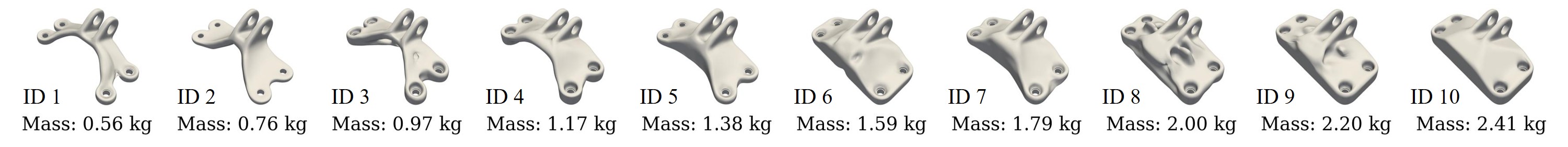}
        \caption{}
        \label{fig:figure_4_c}
    \end{subfigure}
    \begin{subfigure}[b]{1\textwidth}
        \centering
        \includegraphics[width=\textwidth]{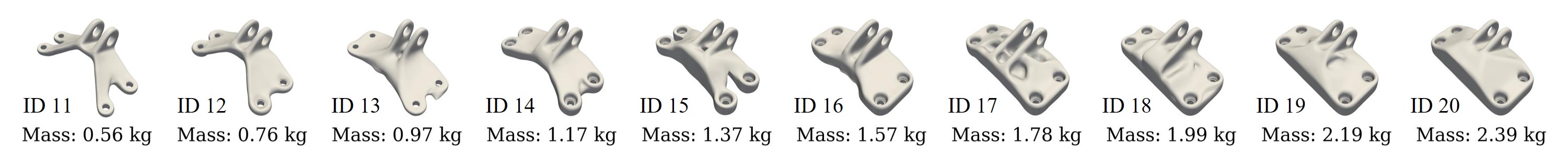}
        \caption{}
        \label{fig:figure_4_d}
    \end{subfigure}
    \caption{Dataset characteristics for training and validation sets. (a) Node count distribution with violin plot insets showing spread. (b) Mass distribution with violin plot insets showing variability. (c) Sampled images from the training set by mass. (d) Sampled images from the validation set by mass. The 20 geometries shown in (c) and (d) are assigned IDs from 1 to 20 (left to right, top to bottom) for reference in the error analysis in Section 3.5.}
    \label{fig:figure_4}
\end{figure}

In the training and validation sets, the dataset was further analyzed based on node count and mass distribution, as depicted in Figure~\ref{fig:figure_4}. Figures~\ref{fig:figure_4_a} and \ref{fig:figure_4_b} illustrate the distributions of node count and mass, respectively, with violin plot insets showing the variability within each subset. This visualization highlights the spread and central tendencies of these characteristics, ensuring that the dataset is representative across different structural sizes and weights. The distribution in each subset confirms that the dataset split maintains a similar statistical profile in terms of structural complexity and mass, which is critical for ensuring consistent performance during training and evaluation.

To ensure a comprehensive understanding of structural variations, sample images from the training and validation sets are shown in Figures~\ref{fig:figure_4_c} and \ref{fig:figure_4_d}, organized by mass. These images demonstrate the diversity of bracket geometries, with mass values ranging from 0.56~kg to 2.41~kg. Figures~\ref{fig:figure_4_c} and \ref{fig:figure_4_d} specifically display 10 evenly spaced samples from the smallest to the largest mass in the training and validation datasets, respectively. The selected samples illustrate the range of structural forms that the model encounters, contributing to robust model training and validation through varied and representative data points. Such visual inspection allows for validation of the dataset's suitability in capturing the range of geometric and mass-related variability needed for accurate predictive modeling.

\subsection{Model Architecture}
\label{subsec:ModelArchitecture}

This study explores three primary architectures—PointNet, DeepONet, and the proposed Point-DeepONet—to effectively model displacement and stress fields in non-parametric 3D geometries under varying load conditions. 

\begin{figure}[!h]
    \centering
    \includegraphics[width=0.85\textwidth]{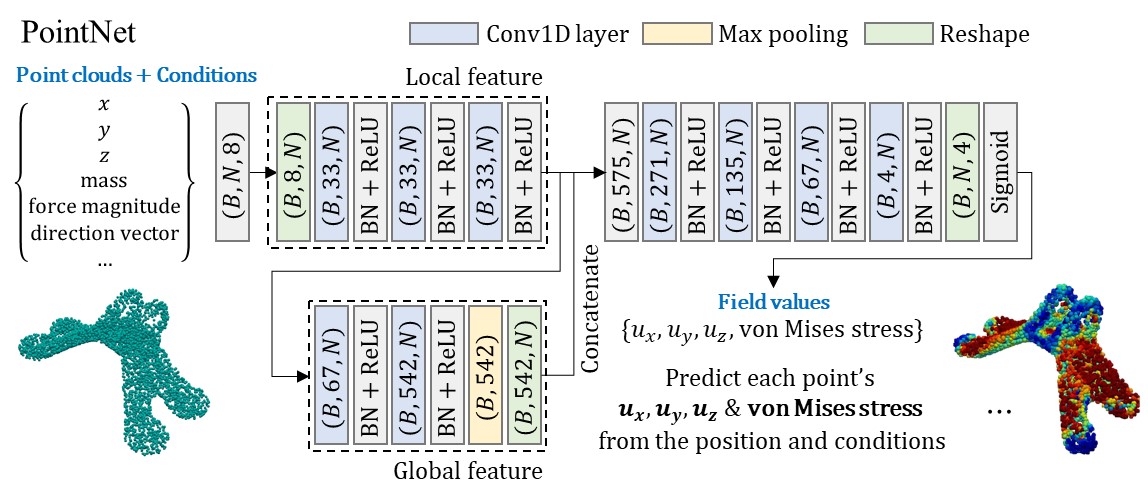}
    \caption{PointNet architecture.}
    \label{fig:figure_5}
\end{figure}

\paragraph{PointNet Architecture}
The PointNet architecture, shown in Figure~\ref{fig:figure_5}, is adapted from the original framework by \citet{qi2017pointnet} with modifications similar to those made by \citet{kashefi2021point} to accommodate variable and irregular geometries. PointNet is used to predict displacement fields ($u_x$, $u_y$, $u_z$) and von Mises stress from 3D point cloud data under various load conditions.

In this setup, $B$ represents the batch size, and $N$ is the number of points (e.g., 5,000) in each input point cloud. The input to PointNet includes several components: point coordinates $(x, y, z)$ with shape $(B, N, 3)$; scalar quantities such as mass $(B, N, 1)$; force magnitude $(B, N, 1)$ and a unit direction vector $(B, N, 3)$ that defines the loading direction.

The architecture begins with a sequence of one-dimensional convolution (Conv1D) layers, each followed by batch normalization (BN) and the ReLU activation function. BN standardizes the inputs of each layer, improving training stability and accelerating convergence, while ReLU adds nonlinearity without saturating gradients. This combination of Conv1D+BN+ReLU acts as a shared multilayer perceptron (MLP) applied to each point independently, enabling the extraction of localized geometric and condition-dependent features. After extracting local features, these per-point embeddings are aggregated into a global feature vector via max pooling, allowing PointNet to capture the overall geometric characteristics of the input domain.

Once the global feature is formed, it is concatenated with the local features and passed through additional Conv1D layers with BN and ReLU activations, effectively integrating both local and global information. Finally, the output layer employs a Sigmoid activation function, producing bounded predictions for each point's $u_x$, $u_y$, $u_z$, and von Mises stress.

To maintain a fair comparison with other architectures and reduce computational overhead, we applied a model scaling factor of 0.53 \citep{kashefi2021point}, resulting in a total of 250,927 trainable parameters. By directly handling irregular point clouds and incorporating load conditions, PointNet delivers accurate high-resolution field predictions, making it a suitable baseline for evaluating more advanced operator-learning frameworks.

\begin{figure}[htbp]
    \centering
    \includegraphics[width=0.8\textwidth]{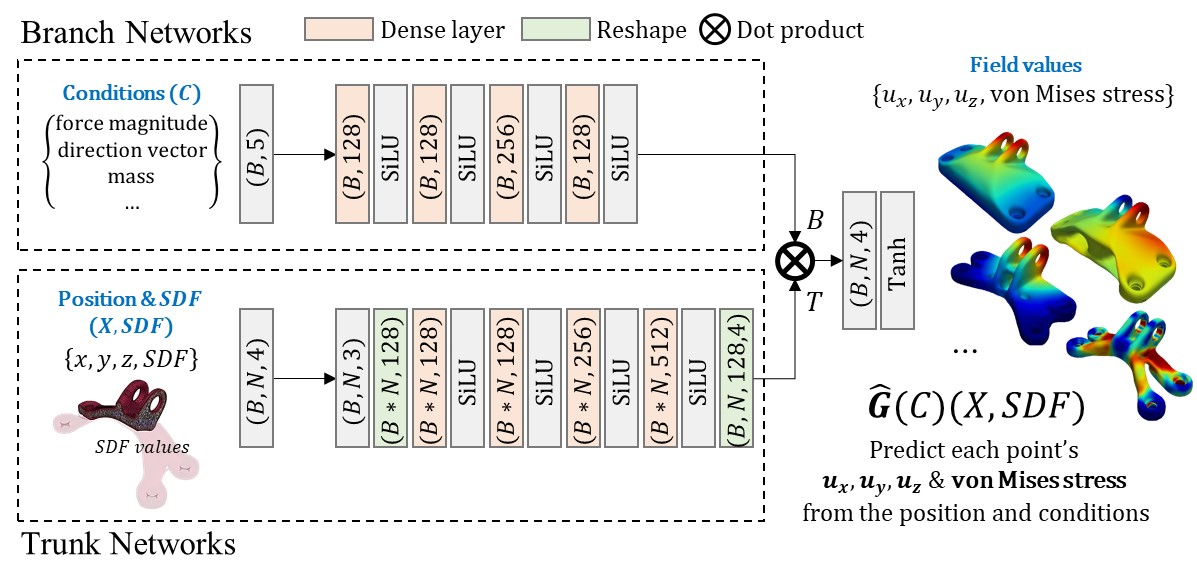}
    \caption{DeepONet architecture: The branch network encodes condition-based inputs into $B$, while the trunk network encodes positional and SDF inputs into $T$. A dot product between $B$ and $T$, followed by a Tanh activation, predicts $u_x$, $u_y$, $u_z$, and von Mises stress.}
    \label{fig:figure_6}
\end{figure}

\paragraph{DeepONet Architecture}
The DeepONet architecture, illustrated in Figure~\ref{fig:figure_6}, is adapted from the framework presented by \citet{lu2021learning} and is designed to predict field quantities (e.g., displacement and stress) within complex, 3D geometries under various loading conditions. DeepONet consists of two main components: a branch network and a trunk network, each tailored to process distinct input types.

The branch network takes in global condition-based inputs, such as force magnitude, direction vectors, and mass. It uses a series of fully connected layers with SiLU activations \citep{Elfwing2018} to encode these condition parameters into a compact latent representation, denoted by $B$. This latent code effectively captures the global physical influences of the applied conditions.

Simultaneously, the trunk network processes point-based inputs consisting of spatial coordinates $(x, y, z)$ and signed distance function (SDF) values \citep{park2019deepsdf}, which provide geometric context by indicating each point's proximity to internal or external boundaries. The trunk network, also utilizing SiLU activations, encodes this geometry-aware information into a latent representation $T$ for each point.

After obtaining the $B$ and $T$ embeddings, a dot product is applied to produce predictions for each point's displacement components ($u_x$, $u_y$, $u_z$) and von Mises stress:
\begin{equation}
\hat{G}(C)(X, SDF) = \tanh\left(\sum_{h=1}^{128} B_h T_h\right),
\label{eq:deeponet_dot}
\end{equation}
where $C$ denotes the global conditions, $B_h$ and $T_h$ are the $h$-th components of the encoded branch and trunk feature vectors, respectively, and $h=1,\dots,128$ indexes the latent feature dimension. A Tanh activation function is applied at the output to ensure bounded and smooth predictions, which can help stabilize training and improve the quality of results.

During training, a key challenge arises from the variability in mesh resolutions. Different geometries, discretized with varying numbers of nodes and elements, produce input data arrays of non-uniform sizes, hindering batched training. To address this, we randomly resample all nodal coordinates and their corresponding outputs to a fixed size $N$, introducing repetition when necessary. This resampling approach enables batched training, significantly improving computational efficiency and facilitating comparisons across diverse geometries. Unlike previous works \citep{shen2023deep, li2024geometry}, where subsampling was used mainly for dataset size reduction, this strategy directly tackles the inherent discrepancies in mesh node counts while maintaining prediction accuracy. During inference, the model remains flexible and can handle arbitrary node counts without retraining, as depicted in Figure~\ref{fig:figure_6}.

We implemented DeepONet with 264,931 parameters using the DeepXDE framework \citep{lu2021deepxde} and the PyTorch backend \citep{NEURIPS2019_bdbca288}. By incorporating geometry via SDF values, employing SiLU activations for stable and expressive layer transformations, and applying Tanh at the output for bounded predictions, this architecture robustly captures how changing load conditions influence the complex geometry's mechanical response.

\begin{figure}[!h]
    \centering
    \includegraphics[width=0.9\textwidth]{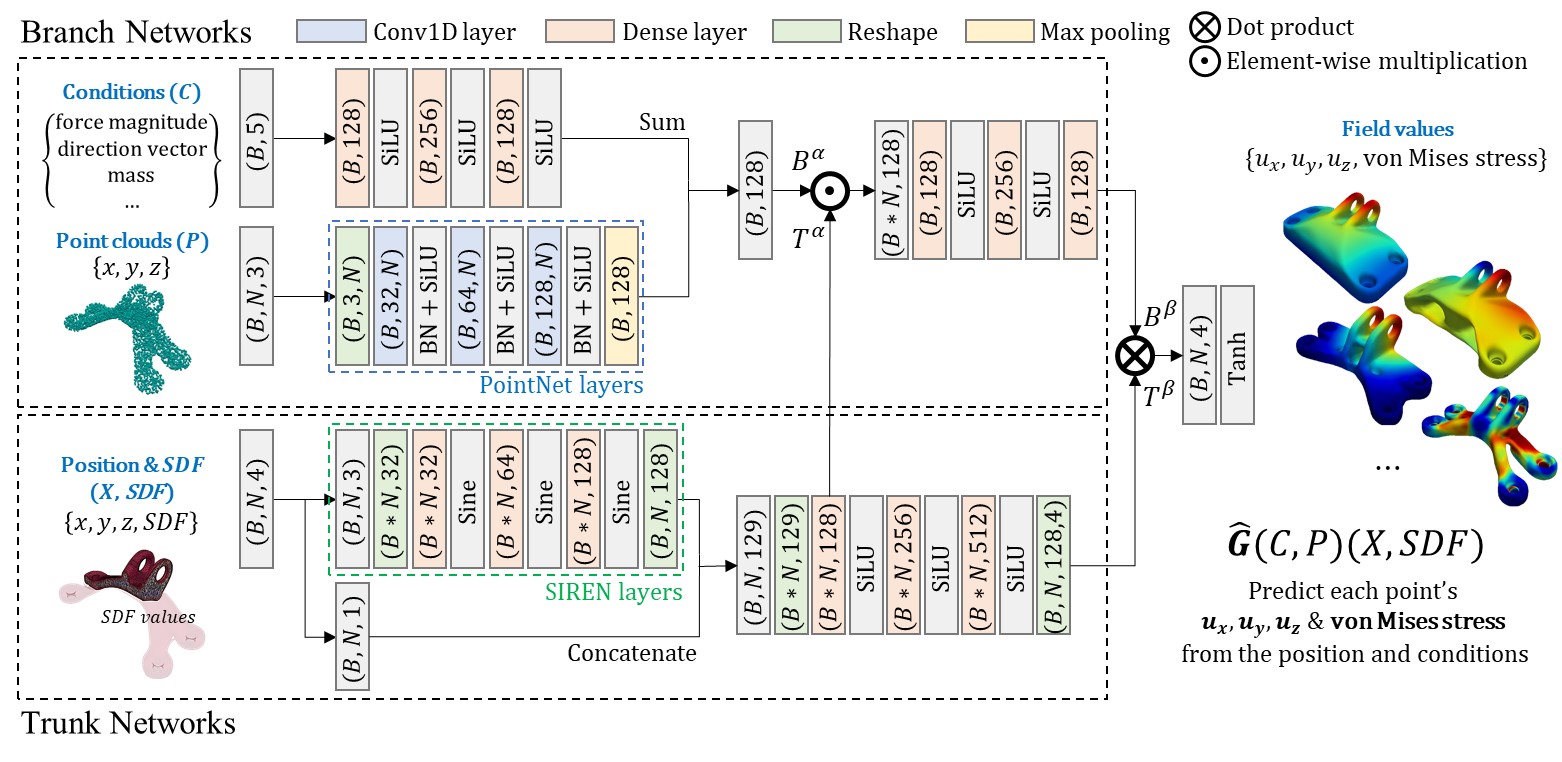}
    \caption{Proposed Point-DeepONet architecture: Condition-based inputs and point cloud data are encoded in the branch network to form $B^\alpha$, while $(x, y, z, SDF)$ coordinates are processed through SIREN layers in the trunk network to form $T^\alpha$. After element-wise multiplication of $B^\alpha$ and $T^\alpha$, subsequent layers yield $B^\beta$ and $T^\beta$. Finally, a dot product produces the predicted vector field (e.g., $u_x, u_y, u_z$, von Mises stress).}
    \label{fig:figure_7}
\end{figure}

\paragraph{Proposed Point-DeepONet Architecture}
The integration in Point-DeepONet is not a simple stacking of modules but a synergistic fusion designed to create a powerful, geometry-aware operator. The architecture, illustrated in Figure~\ref{fig:figure_7}, leverages each component for a distinct purpose. The PointNet module in the branch network acts as a learnable feature extractor for the input geometry function, which is a significant advancement over methods that rely on fixed geometric parameters. It processes the raw point cloud $P \in \mathbb{R}^{B,N,3}$ to produce a global geometric embedding $B^{(P)} \in \mathbb{R}^{B,128}$. This embedding captures the holistic shape of the object. Concurrently, a separate MLP-based sub-network processes the global physical conditions $C$ (e.g., load magnitude, direction, mass) into a condition embedding $B^{(C)} \in \mathbb{R}^{B,128}$. These two latent vectors are then combined to form a unified context vector for the branch network:
\begin{equation}
    B^\alpha_{b,h} = B^{(C)}_{b,h} + B^{(P)}_{b,h}.
\end{equation}
On the trunk side, the input is enriched by incorporating signed distance function (SDF) values in addition to the coordinates $X=(x,y,z)$. We feed $(X,SDF) \in \mathbb{R}^{B,N,4}$ into sinusoidal representation network (SIREN) layers \citep{sitzmann2020implicit}, which utilize sine-based activation functions to effectively represent high-frequency details in the solution field:
\begin{equation}
    T^\alpha_{b,i,h} = f_{\text{SIREN}}(X_{b,i}, SDF_{b,i}).
\end{equation}
The key innovation lies in the fusion mechanism. The original DeepONet architecture fuses information only at the final stage via a single dot product, limiting the interaction between global and local features. Our approach, inspired by recent advancements \citep{wang2022improved, he2023novel}, incorporates an element-wise multiplication between the branch context vector $B^\alpha$ and the trunk's point-wise feature vector $T^\alpha$. This operation,
\begin{equation}
    F_{b,i,h} = B^\alpha_{b,h} \cdot T^\alpha_{b,i,h},
\end{equation}
allows the global geometric and physical context to modulate the learned basis functions at each point in space. From a physical perspective, this element-wise modulation enables the global context (geometry and load conditions encoded in $B^\alpha$) to locally adapt the basis functions (encoded in $T^\alpha$) at each spatial location. In structural mechanics, stress and displacement fields depend on both global boundary conditions and local geometric features—a thin section responds differently than a thick section under the same global load. The element-wise multiplication naturally captures this physics: each component of $B^\alpha$ acts as a spatially-varying coefficient that scales the corresponding local basis function. This mechanism is conceptually analogous to how material properties or geometric parameters locally modify the Green's function in classical elasticity theory \citep{stakgold2011green}. In contrast, a single dot product aggregates all information into a scalar, losing the ability to represent such position-dependent modulation. This richer interaction allows the model to better approximate the true physical operator, particularly in problems with strong local-global coupling such as stress concentration near geometric discontinuities.

Subsequent MLP layers refine $F_{b,i,h}$ into two sets of features, $B^\beta$ and $T^\beta$, from which we derive the displacement and stress fields—specifically, three displacement components ($u_x,u_y,u_z$) and the von Mises stress. A final dot product along the latent feature dimension yields:
\begin{equation}
    \hat{G}(C,P)(X,SDF)_{b,i,m} = \tanh\left(\sum_{h=1}^{128} B^\beta_{b,i,h} \cdot T^\beta_{b,i,h,m}\right),
\end{equation}
where $b=1,\dots,B$ indexes the batch, $i=1,\dots,N$ indexes the finite element nodes, $h=1,\dots,128$ spans the latent feature dimension, and $m=1,\dots,4$ indexes the field variables.

By uniting DeepONet's operator-learning framework with PointNet's robust geometric feature extraction, augmented by SDF-based shape encoding, SIREN layers, early feature fusion, BN layers for stable training, SiLU for enhanced nonlinear expressiveness, and Tanh for bounded outputs, Point-DeepONet delivers accurate, high-resolution predictions directly on arbitrary meshes under complex, nonlinear loading scenarios. Implemented with 251,936 trainable parameters using DeepXDE \citep{lu2021deepxde} and PyTorch \citep{NEURIPS2019_bdbca288}, this approach is both scalable and computationally efficient, serving as a next-generation surrogate model for advanced structural analyses.

\section{Results and Discussion}
\label{sec:results_discussion}

In this section, we present a comprehensive evaluation of our proposed Point-DeepONet architecture. The models were trained using an NVIDIA A100 GPU with 80GB of HBM2e memory. Our objective is to assess the efficacy of each model in capturing the complex nonlinear elastoplastic behavior inherent in these structures. We begin by detailing the optimization strategies and training procedures employed to ensure a fair and consistent comparison among the models. We then present detailed results from our extensive validation experiments, including an ablation study to justify our architectural choices and tests on unseen load conditions to demonstrate the model's generalization capabilities.


\subsection{Prediction Results}
\label{subsec:prediction_results}

\paragraph{Training Details}
To ensure an equitable comparison, all models were trained under standardized conditions using the AdamW optimizer \citep{loshchilov2017decoupled}, which incorporates weight decay regularization to mitigate overfitting by penalizing large weights. The initial learning rate was set to $1 \times 10^{-3}$, and the mean squared error (MSE) loss function was employed to quantify the discrepancy between the predicted outputs and the ground truth values obtained from finite element analysis (FEA). Training was conducted with a batch size of 16 to balance computational efficiency and convergence stability.

The PointNet model was trained for 4,000 iterations, whereas DeepONet and our proposed Point-DeepONet were trained for 40,000 iterations each. The extended training duration for DeepONet and Point-DeepONet accounts for their more complex architectures and the necessity for additional iterations to achieve convergence. To ensure reproducibility and eliminate randomness in the training process, a fixed random seed was utilized across all experiments.

Model performance was quantitatively assessed using three key metrics: mean absolute error (MAE), root mean square error (RMSE), and the coefficient of determination ($R^2$). These metrics were computed on both the training and validation datasets to evaluate the models' accuracy and generalization capabilities. The MAE provides a straightforward measure of the average magnitude of errors, RMSE emphasizes larger errors due to the squared term, and $R^2$ indicates the proportion of variance in the dependent variable predictable from the independent variables. By maintaining consistent hyperparameters and training configurations, we ensured that any differences in performance could be attributed to the inherent capabilities of the architectures rather than external factors.

Mathematically, the performance metrics are defined as follows:

\begin{equation}
\text{MAE} = \frac{1}{n} \sum_{i=1}^{n} \left| y_i - \hat{y}_i \right|,
\label{eq:mae}
\end{equation}

\begin{equation}
\text{RMSE} = \sqrt{ \frac{1}{n} \sum_{i=1}^{n} \left( y_i - \hat{y}_i \right)^2 },
\label{eq:rmse}
\end{equation}

\begin{equation}
R^2 = 1 - \frac{\sum_{i=1}^{n} \left( y_i - \hat{y}_i \right)^2}{\sum_{i=1}^{n} \left( y_i - \bar{y} \right)^2},
\label{eq:r_squared}
\end{equation}

where $y_i$ are the true values from FEA, $\hat{y}_i$ are the predicted values from the models, $\bar{y}$ is the mean of the true values, and $n$ is the total number of samples.

\begin{figure}[h]
    \centering
    \begin{subfigure}[b]{0.32\textwidth}
        \centering
        \includegraphics[width=\textwidth]{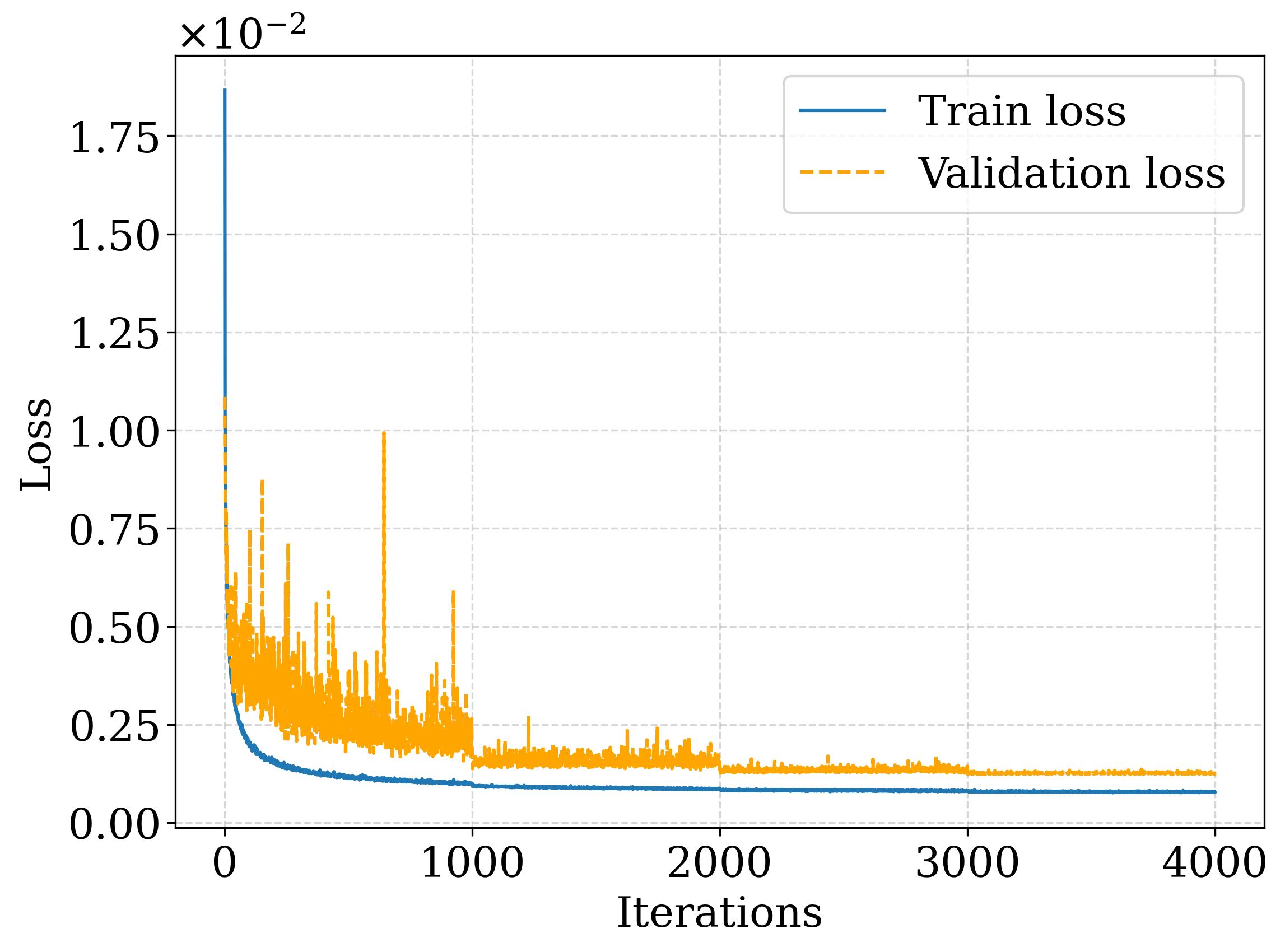}
        \caption{}
        \label{fig:figure_8_a}
    \end{subfigure}
    \begin{subfigure}[b]{0.32\textwidth}
        \centering
        \includegraphics[width=\textwidth]{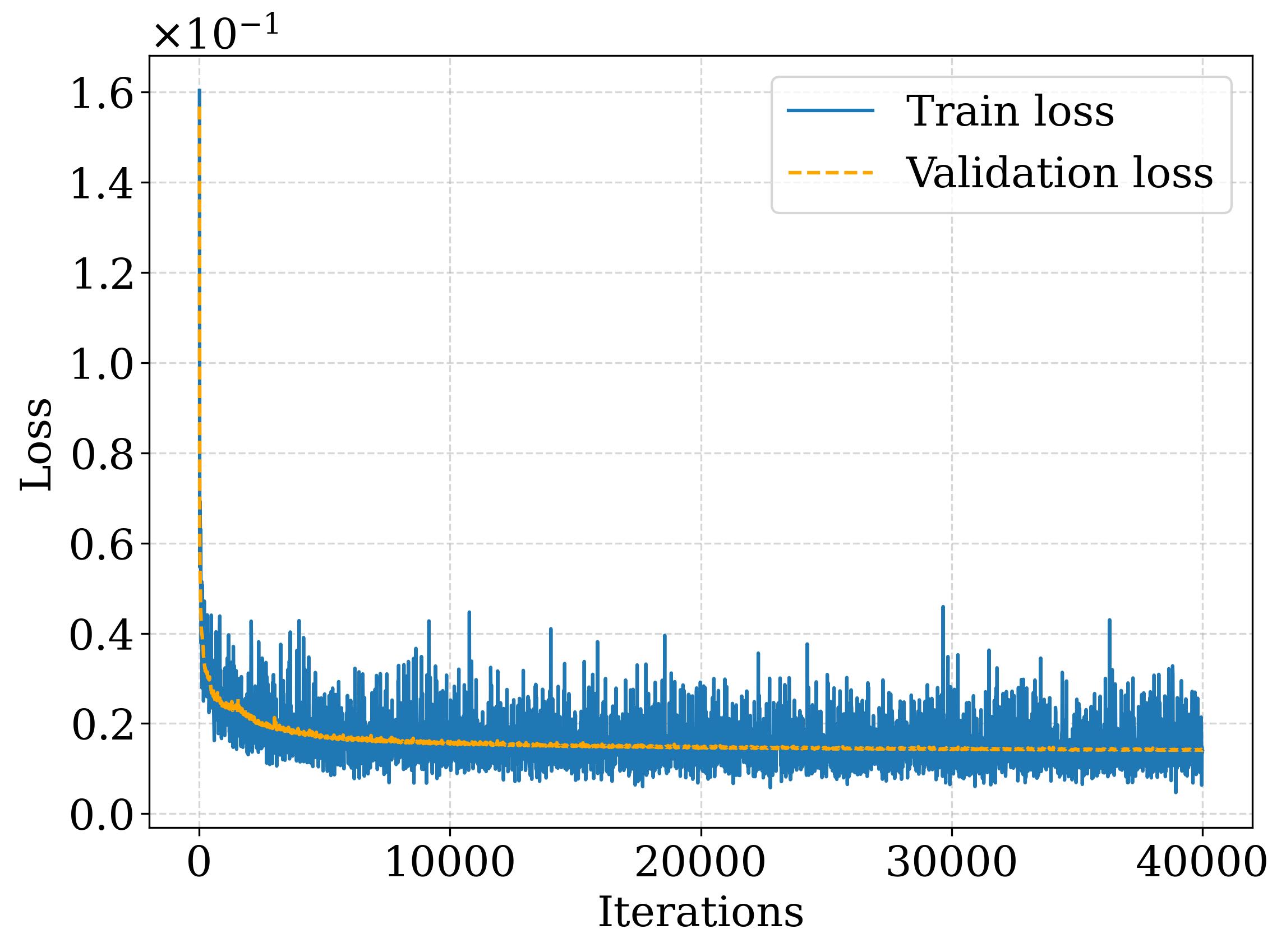}
        \caption{}
        \label{fig:figure_8_b}
    \end{subfigure}
    \begin{subfigure}[b]{0.32\textwidth}
        \centering
        \includegraphics[width=\textwidth]{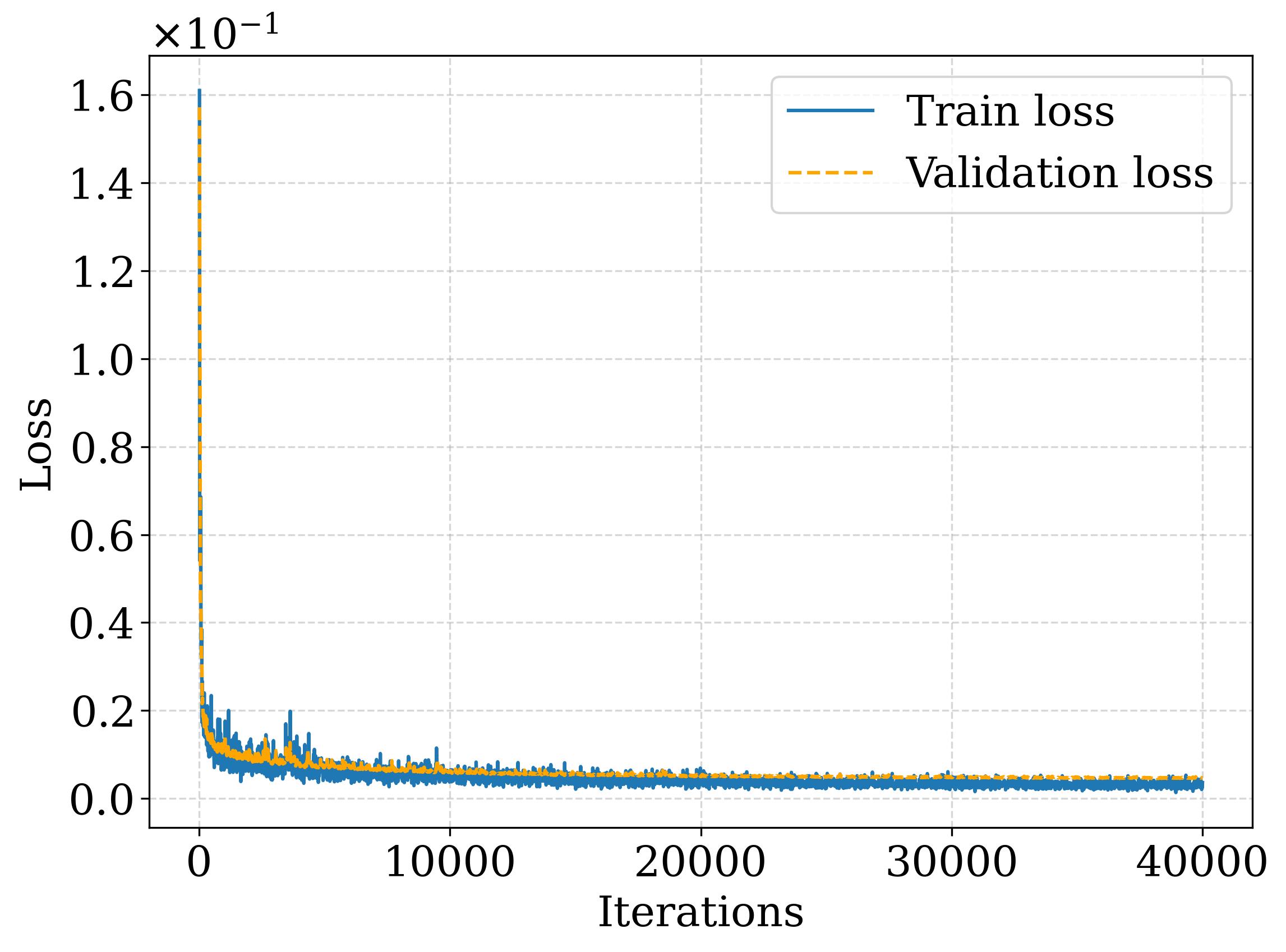}
        \caption{}
        \label{fig:figure_8_c}
    \end{subfigure}
    \caption{
        Training and validation loss curves for (a) PointNet, (b) DeepONet, and (c) Point-DeepONet. The curves illustrate the convergence characteristics of each model during the training process. Notably, Point-DeepONet achieves smoother and more stable convergence compared to the baseline models.
    }
    \label{fig:figure_8}
\end{figure}

Figure~\ref{fig:figure_8} illustrates the training and validation loss curves for the PointNet, DeepONet, and Point-DeepONet models. The PointNet model (Figure~\ref{fig:figure_8_a}) exhibits a rapid decrease in training loss during the initial stages; however, it shows signs of overfitting, as evidenced by the divergence between training and validation losses in the later stages.

The DeepONet model (Figure~\ref{fig:figure_8_b}) demonstrates improved stability in both training and validation losses compared to PointNet. Nonetheless, oscillations are observed during the training process, indicating potential instability when handling complex data.

In contrast, the Point-DeepONet model (Figure~\ref{fig:figure_8_c}) achieves the most stable and smooth convergence in both training and validation losses. The minimal gap between these losses underscores the model's ability to effectively learn complex data while maintaining generalization performance. These results highlight the suitability of Point-DeepONet as a robust framework for nonlinear structural analysis problems.

\begin{table}[h]
    \caption{Performance comparison of PointNet, DeepONet, and Point-DeepONet models using sampled and full data for displacement and von Mises stress metrics}
    \label{tab:selected_model_metrics_comparison_full_subset}
    \centering
    \scriptsize
    \setlength{\tabcolsep}{2pt}
    \begin{adjustbox}{width=\textwidth}
        \begin{threeparttable}
            \begin{tabular}{@{}lc|cc|cc|cc|cc|cc|cc|cc|cc@{}}
                \toprule
                \multirow{3}{*}{Model} & \multirow{3}{*}{Component} 
                & \multicolumn{4}{c|}{$u_x$ (mm)} 
                & \multicolumn{4}{c|}{$u_y$ (mm)} 
                & \multicolumn{4}{c|}{$u_z$ (mm)} 
                & \multicolumn{4}{c}{von Mises stress (MPa)} \\
                \cmidrule(l){3-18}
                &  & \multicolumn{2}{c|}{Sampled data} & \multicolumn{2}{c|}{Full data} 
                & \multicolumn{2}{c|}{Sampled data} & \multicolumn{2}{c|}{Full data} 
                & \multicolumn{2}{c|}{Sampled data} & \multicolumn{2}{c|}{Full data} 
                & \multicolumn{2}{c|}{Sampled data} & \multicolumn{2}{c}{Full data} \\
                \cmidrule(l){3-18}
                &  & MAE & $R^2$ & MAE & $R^2$ 
                & MAE & $R^2$ & MAE & $R^2$ 
                & MAE & $R^2$ & MAE & $R^2$ 
                & MAE & $R^2$ & MAE & $R^2$ \\
                \midrule
                \multirow{3}{*}{PointNet} 
                & Vertical   
                    & 0.008 & 0.953 & - & - 
                    & 0.003 & 0.927 & - & - 
                    & 0.013 & 0.952 & - & - 
                    & 11.666 & 0.866 & - & - \\
                & Horizontal 
                    & 0.006 & 0.983 & - & - 
                    & 0.002 & 0.878 & - & - 
                    & 0.008 & 0.980 & - & - 
                    & 9.073 & 0.904 & - & - \\
                & Diagonal   
                    & 0.004 & 0.884 & - & - 
                    & 0.002 & 0.963 & - & - 
                    & 0.006 & 0.975 & - & - 
                    & 7.639 & 0.859 & - & - \\
                \midrule
                \multirow{3}{*}{DeepONet} 
                & Vertical   
                    & 0.026 & 0.626 & 0.018 & 0.731 
                    & 0.007 & 0.709 & 0.006 & 0.800 
                    & 0.048 & 0.654 & 0.031 & 0.834 
                    & 21.842 & 0.592 & 17.289 & 0.691 \\
                & Horizontal 
                    & 0.024 & 0.694 & 0.017 & 0.793 
                    & 0.004 & 0.503 & 0.003 & 0.575 
                    & 0.026 & 0.699 & 0.019 & 0.764 
                    & 20.091 & 0.572 & 16.370 & 0.657 \\
                & Diagonal   
                    & 0.006 & 0.578 & 0.006 & 0.542 
                    & 0.005 & 0.753 & 0.004 & 0.781 
                    & 0.016 & 0.776 & 0.013 & 0.807 
                    & 12.555 & 0.645 & 11.172 & 0.706 \\
                \midrule
                \multirow{3}{*}{\makecell{Point-DeepONet \\ (Our proposed)}} 
                & Vertical   
                    & 0.007 & 0.964 & 0.006 & 0.979 
                    & 0.003 & 0.923 & 0.003 & 0.949 
                    & 0.012 & 0.961 & 0.011 & 0.981 
                    & 10.541 & 0.882 & 10.139 & 0.881 \\
                & Horizontal 
                    & 0.005 & 0.985 & 0.005 & 0.987 
                    & 0.002 & 0.875 & 0.002 & 0.882 
                    & 0.006 & 0.985 & 0.006 & 0.985 
                    & 7.935 & 0.923 & 7.690 & 0.916 \\
                & Diagonal   
                    & 0.003 & 0.903 & 0.003 & 0.890 
                    & 0.002 & 0.953 & 0.002 & 0.955 
                    & 0.005 & 0.976 & 0.005 & 0.978 
                    & 7.090 & 0.873 & 6.866 & 0.876 \\
                \bottomrule
            \end{tabular}
        \end{threeparttable}
    \end{adjustbox}
\end{table}

Table~\ref{tab:selected_model_metrics_comparison_full_subset} presents a performance comparison of the PointNet, DeepONet, and Point-DeepONet models using both sampled and full data for displacement and von Mises stress metrics. The sampled data refers to predictions made using 5,000 points, consistent with the training configuration for PointNet. In contrast, the full data inference involves predictions across all nodes in the finite element mesh, showcasing each model's ability to generalize beyond the initial sampling.

From the table, it is evident that PointNet is limited to predicting only the sampled data, as indicated by the absence of results for the full data. This constraint highlights PointNet's scalability limitation. In contrast, both DeepONet and our proposed Point-DeepONet exhibit the capability to infer over the full mesh without requiring retraining, demonstrating superior scalability and adaptability to high-resolution datasets.

Evaluating the MAE and $R^2$ values, Point-DeepONet consistently outperforms DeepONet across most load directions and components, especially when considering full data. For instance, under the horizontal load case for von Mises stress, Point-DeepONet attains lower MAE values and higher $R^2$ scores than DeepONet, underscoring its improved accuracy and robustness.

In terms of inference time, while both DeepONet and Point-DeepONet process sampled data faster than full mesh data, Point-DeepONet's more complex architecture results in slightly longer inference times compared to DeepONet. For example, DeepONet may take around 0.4~seconds per case for sampled data and 2.5~seconds for full mesh predictions, whereas Point-DeepONet might require approximately 0.6~seconds for sampled data and 3.0~seconds for full data. Although marginally slower than DeepONet, Point-DeepONet's inference still remains substantially faster than conventional FEA.

As previously described in Section~\ref{subsec:DatasetGeneration}, the finite element simulations are computationally intensive, averaging around 19.32~minutes (approximately 1,159~seconds) per load case. Compared to these simulations, Point-DeepONet inference for the full mesh—on the order of a few seconds—achieves more than $4 \times 10^2$ times faster computation. This dramatic speedup significantly accelerates the design exploration process, enabling rapid evaluation of numerous configurations and fostering near-real-time structural analysis capabilities.

To provide a comprehensive time metric for practical deployment, we quantify the total computational cost when 500 structural evaluations are required (Table~\ref{tab:optimization_time_comparison}). This scenario is representative of iterative engineering applications including design optimization cycles, uncertainty quantification studies, load case analyses, and parametric investigations. Traditional FEA would consume approximately 161 hours (500 × 19.32 minutes) for these evaluations. In contrast, Point-DeepONet's workflow comprises a one-time setup—dataset generation (966 hours, amortized across multiple projects), model training (0.95 hours), and per-geometry SDF generation (0.75 minutes)—followed by rapid 3.0-second inference per evaluation. The 500 evaluations thus require only 0.42 hours (25 minutes) of iterative time, yielding a 383× speedup. Importantly, the upfront costs are one-time investments reusable across different load scenarios, design variants, and uncertainty studies. For organizations conducting repeated structural analyses, the amortized cost becomes negligible, and the effective speedup approaches 400×. The break-even point occurs after approximately 50 evaluations, confirming Point-DeepONet's cost-effectiveness for iterative engineering workflows.

\begin{table}[h]
    \centering
    \caption{Comprehensive comparison of total computational time for 500 structural evaluations, including all setup and iterative costs.}
    \label{tab:optimization_time_comparison}
    
    \begin{tabular}{lcc}
        \toprule
        Task & Traditional FEA & Point-DeepONet \\
        \midrule
        Training dataset generation (3,000 cases) & 966 hours & 966 hours \\
        Model training & N/A & 0.95 hours \\
        SDF generation per geometry (one-time) & N/A & 0.75 minutes \\
        Evaluation per iteration & 19.32 minutes & 3.0 seconds \\
        \midrule
        \textbf{Total time for 500 evaluations} & \textbf{161 hours} & 
        \textbf{967.2 hours (one-time)} \\
        & & \textbf{+ 0.42 hours (iterations)} \\
        \midrule
        \textbf{Amortized time after initial setup} & \textbf{161 hours} & 
        \textbf{0.42 hours} \\
        \textbf{Speedup factor} & \textbf{1×} & \textbf{383×} \\
        \bottomrule
    \end{tabular}
    \begin{tablenotes}
        \footnotesize
        \item Note: The one-time setup cost (dataset generation + training + SDF) is amortized across multiple analyses. For subsequent evaluations using the same trained model, only the iterative evaluation time (0.42 hours for 500 cases) is relevant.
    \end{tablenotes}
\end{table}

\begin{figure}[htbp]
    \centering
    \begin{subfigure}{0.78\textwidth}
        \centering
        \begin{subfigure}{0.19\textwidth}
            \centering
            \includegraphics[width=\linewidth]{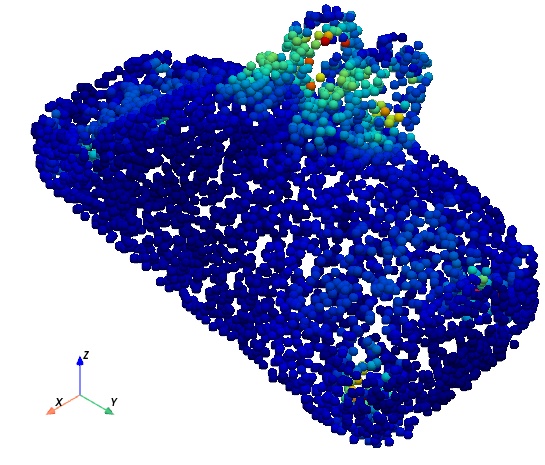}
            \captionsetup[sub]{font=small}
            \caption{FE, best}
            \label{fig:figure_9_a}
        \end{subfigure}
        \hfill
        \begin{subfigure}{0.19\textwidth}
            \centering
            \includegraphics[width=\linewidth]{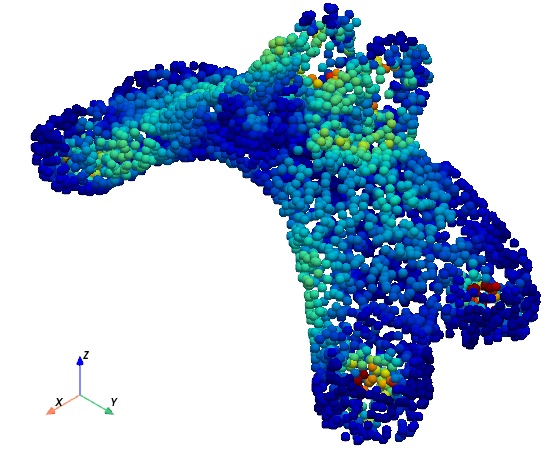}
            \captionsetup[sub]{font=small}
            \caption{FE, 25$^{th}$ pct.}
            \label{fig:figure_9_b}
        \end{subfigure}
        \hfill
        \begin{subfigure}{0.19\textwidth}
            \centering
            \includegraphics[width=\linewidth]{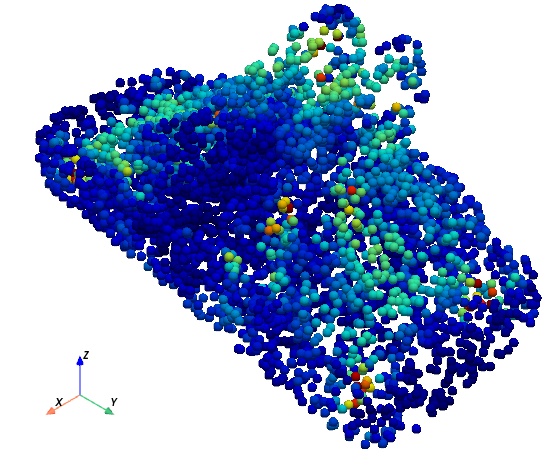}
            \captionsetup[sub]{font=small}
            \caption{FE, 50$^{th}$ pct.}
            \label{fig:figure_9_c}
        \end{subfigure}
        \hfill
        \begin{subfigure}{0.19\textwidth}
            \centering
            \includegraphics[width=\linewidth]{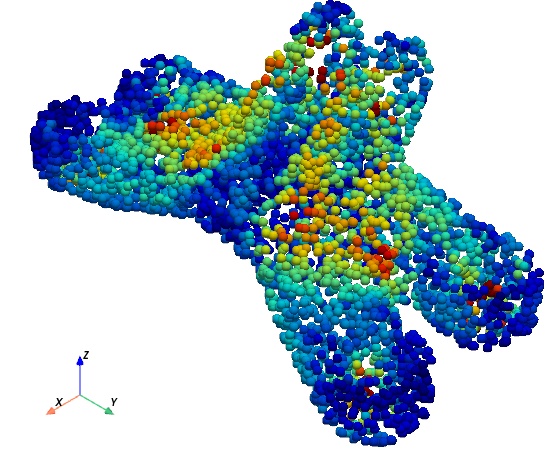}
            \captionsetup[sub]{font=small}
            \caption{FE, 75$^{th}$ pct.}
            \label{fig:figure_9_d}
        \end{subfigure}
        \hfill
        \begin{subfigure}{0.19\textwidth}
            \centering
            \includegraphics[width=\linewidth]{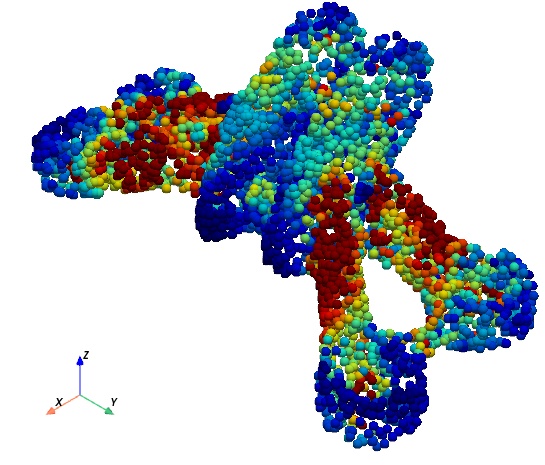}
            \captionsetup[sub]{font=small}
            \caption{FE, worst}
            \label{fig:figure_9_e}
        \end{subfigure}
        
        \vspace{1mm}
        \begin{subfigure}{0.19\textwidth}
            \centering
            \includegraphics[width=\linewidth]{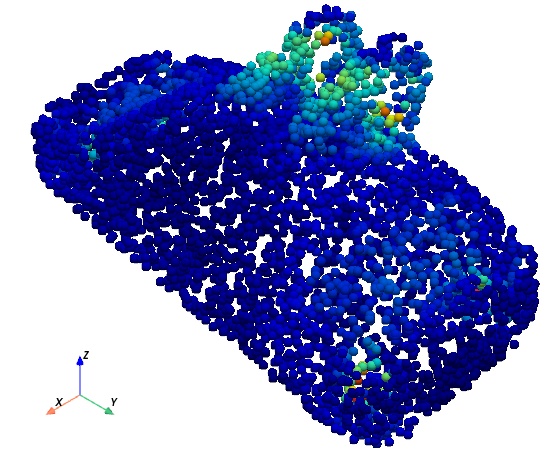}
            \captionsetup[sub]{font=small}
            \caption{Pred., best}
            \label{fig:figure_9_f}
        \end{subfigure}
        \hfill
        \begin{subfigure}{0.19\textwidth}
            \centering
            \includegraphics[width=\linewidth]{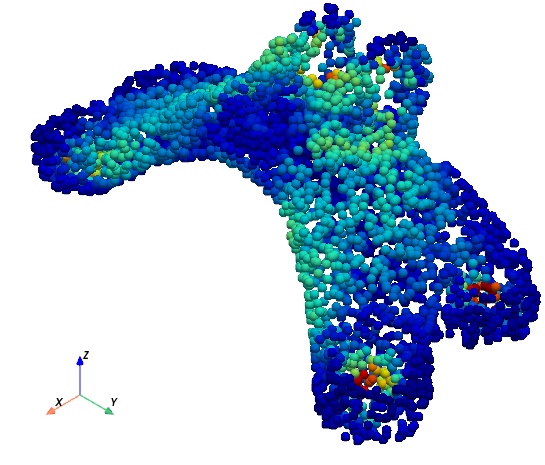}
            \captionsetup[sub]{font=small}
            \caption{Pred., 25$^{th}$ pct.}
            \label{fig:figure_9_g}
        \end{subfigure}
        \hfill
        \begin{subfigure}{0.19\textwidth}
            \centering
            \includegraphics[width=\linewidth]{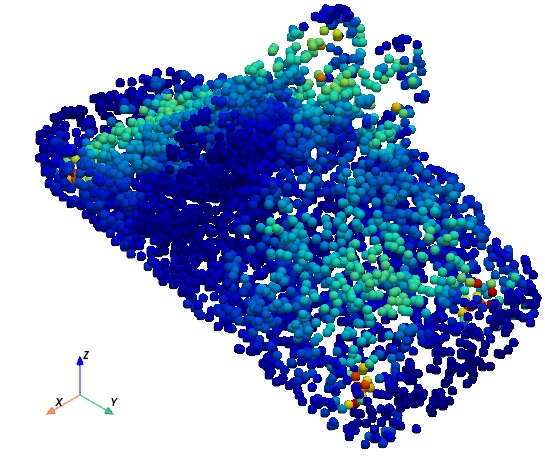}
            \captionsetup[sub]{font=small}
            \caption{Pred., 50$^{th}$ pct.}
            \label{fig:figure_9_h}
        \end{subfigure}
        \hfill
        \begin{subfigure}{0.19\textwidth}
            \centering
            \includegraphics[width=\linewidth]{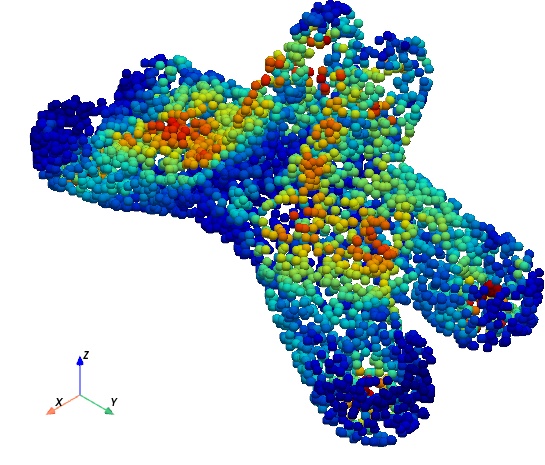}
            \captionsetup[sub]{font=small}
            \caption{Pred., 75$^{th}$ pct.}
            \label{fig:figure_9_i}
        \end{subfigure}
        \hfill
        \begin{subfigure}{0.19\textwidth}
            \centering
            \includegraphics[width=\linewidth]{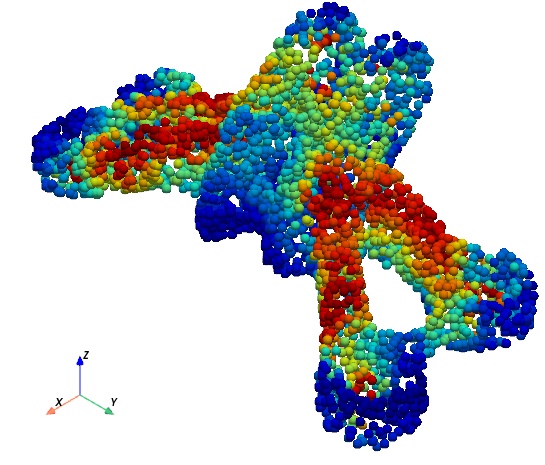}
            \captionsetup[sub]{font=small}
            \caption{Pred., worst.}
            \label{fig:figure_9_j}
        \end{subfigure}
        
        \vspace{1mm}
        \begin{subfigure}{0.19\textwidth}
            \centering
            \includegraphics[width=\linewidth]{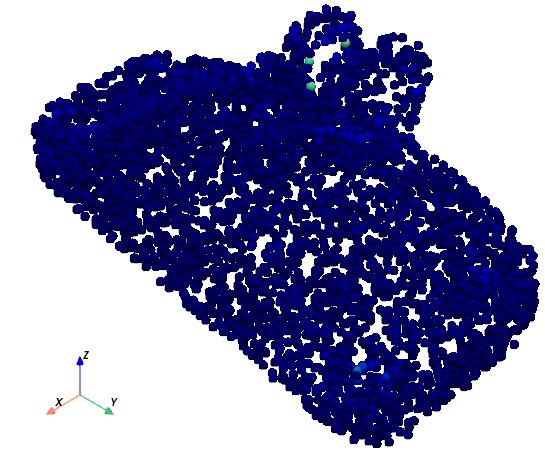}
            \captionsetup[sub]{font=small}
            \caption{MAE: 3.79}
            \label{fig:figure_9_k}
        \end{subfigure}
        \hfill
        \begin{subfigure}{0.19\textwidth}
            \centering
            \includegraphics[width=\linewidth]{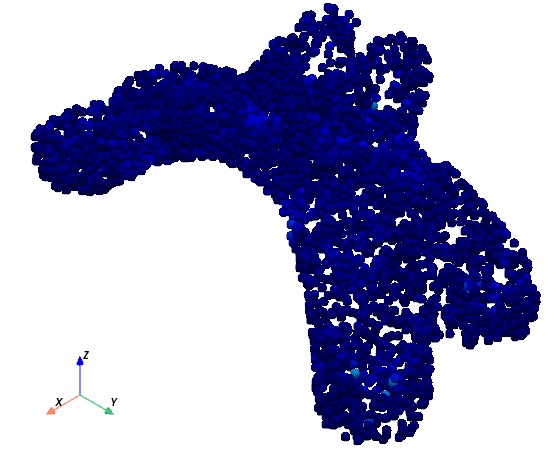}
            \captionsetup[sub]{font=small}
            \caption{MAE: 5.98}
            \label{fig:figure_9_l}
        \end{subfigure}
        \hfill
        \begin{subfigure}{0.19\textwidth}
            \centering
            \includegraphics[width=\linewidth]{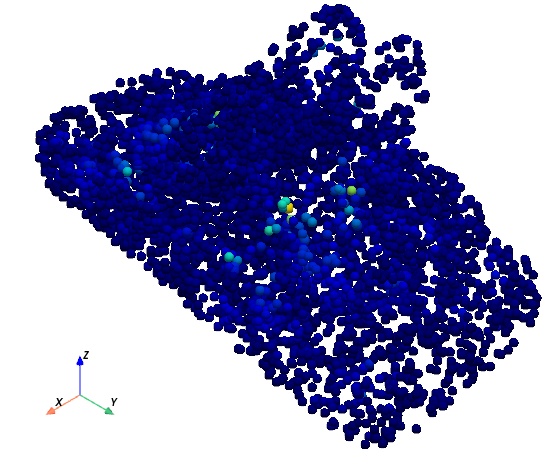}
            \captionsetup[sub]{font=small}
            \caption{MAE: 8.02}
            \label{fig:figure_9_m}
        \end{subfigure}
        \hfill
        \begin{subfigure}{0.2\textwidth}
            \centering
            \includegraphics[width=\linewidth]{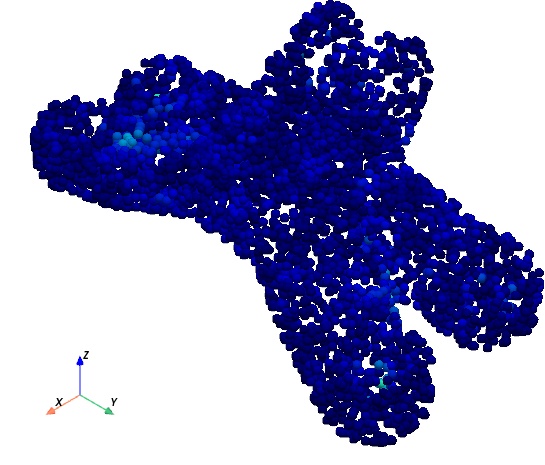}
            \captionsetup[sub]{font=small}
            \caption{MAE: 10.64}
            \label{fig:figure_9_n}
        \end{subfigure}
        \hfill
        \begin{subfigure}{0.19\textwidth}
            \centering
            \includegraphics[width=\linewidth]{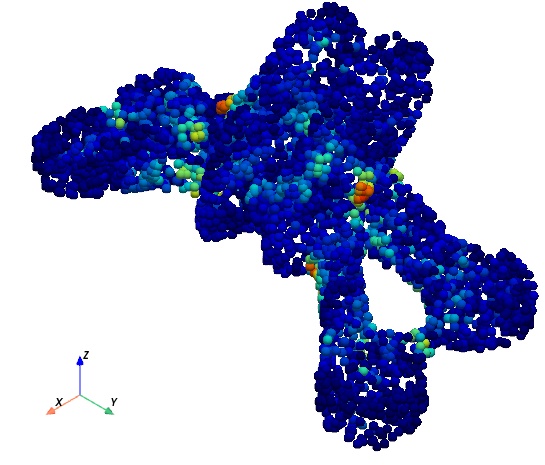}
            \captionsetup[sub]{font=small}
            \caption{MAE: 29.26}
            \label{fig:figure_9_o}
        \end{subfigure}
    \end{subfigure}
    
    \includegraphics[width=0.7\textwidth]{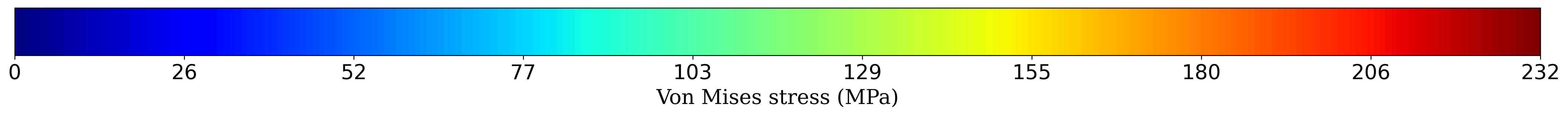}
    
    \caption{Comparison of PointNet predictions and FEA results under horizontal load (von Mises stress). Top row: FEA results; middle row: PointNet predictions; bottom row: absolute errors with MAE values.}
    \label{fig:figure_9}
\end{figure}

\begin{figure}[htbp]
    \centering
    \begin{subfigure}{0.78\textwidth}
        \centering
        \begin{subfigure}{0.19\textwidth}
            \centering
            \includegraphics[width=\linewidth]{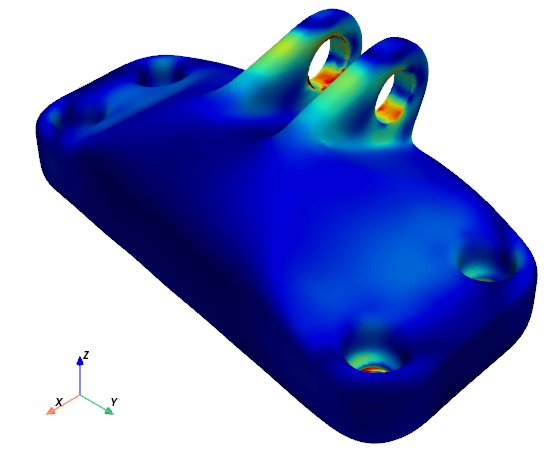}
            \captionsetup{font=small}
            \caption{FE, best.}
            \label{fig:figure_10_a}
        \end{subfigure}
        \hfill
        \begin{subfigure}{0.19\textwidth}
            \centering
            \includegraphics[width=\linewidth]{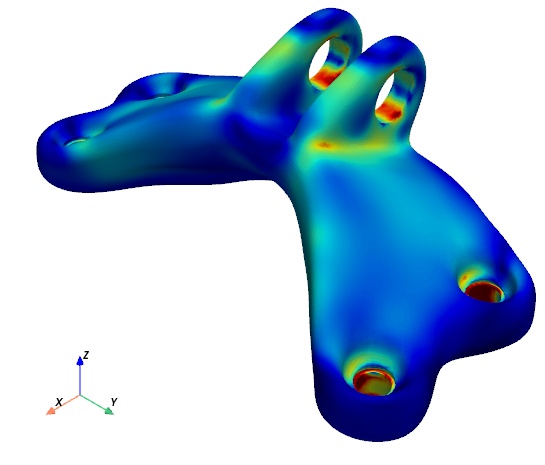}
            \captionsetup{font=small}
            \caption{FEA 25th Pct.}
            \label{fig:figure_10_b}
        \end{subfigure}
        \hfill
        \begin{subfigure}{0.19\textwidth}
            \centering
            \includegraphics[width=\linewidth]{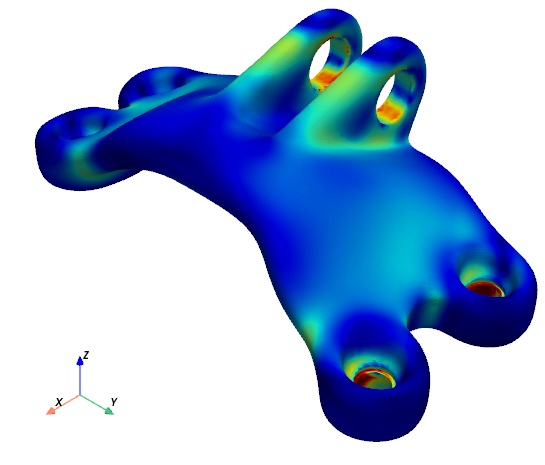}
            \captionsetup{font=small}
            \caption{FE, 50$^{th}$ pct.}
            \label{fig:figure_10_c}
        \end{subfigure}
        \hfill
        \begin{subfigure}{0.19\textwidth}
            \centering
            \includegraphics[width=\linewidth]{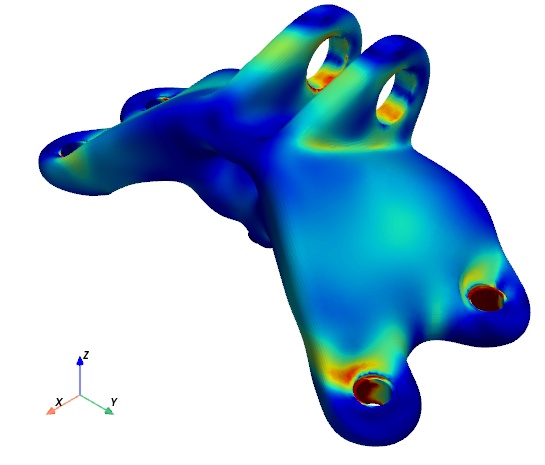}
            \captionsetup{font=small}
            \caption{FE, 75$^{th}$ pct.}
            \label{fig:figure_10_d}
        \end{subfigure}
        \hfill
        \begin{subfigure}{0.19\textwidth}
            \centering
            \includegraphics[width=\linewidth]{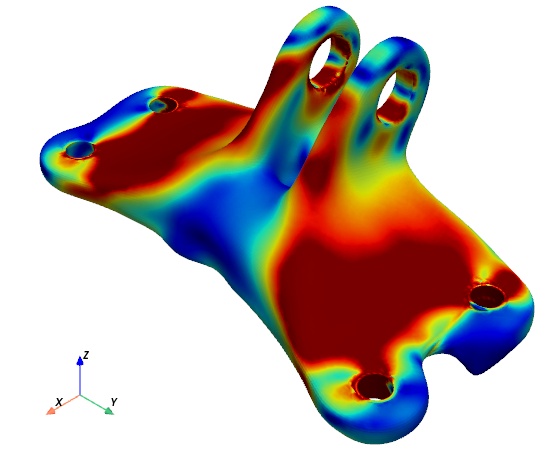}
            \captionsetup{font=small}
            \caption{FE, worst.}
            \label{fig:figure_10_e}
        \end{subfigure}
        
        \begin{subfigure}{0.19\textwidth}
            \centering
            \includegraphics[width=\linewidth]{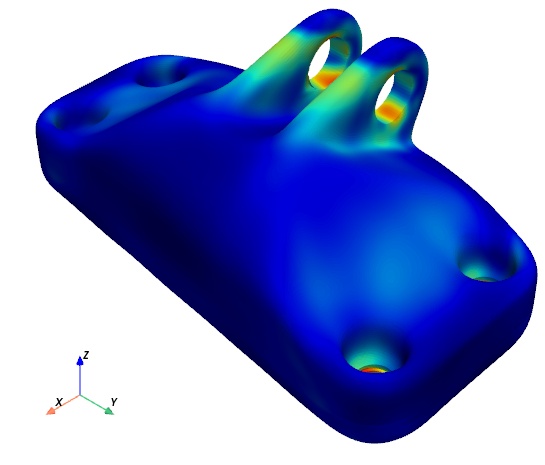}
            \captionsetup{font=small}
            \caption{Pred., best.}
            \label{fig:figure_10_f}
        \end{subfigure}
        \hfill
        \begin{subfigure}{0.19\textwidth}
            \centering
            \includegraphics[width=\linewidth]{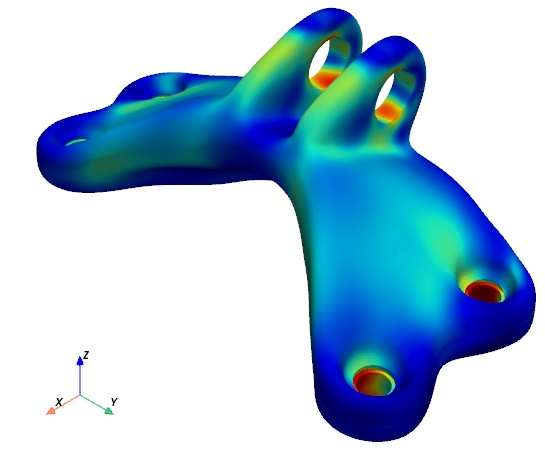}
            \captionsetup{font=small}
            \caption{Pred., 25$^{th}$ pct.}
            \label{fig:figure_10_g}
        \end{subfigure}
        \hfill
        \begin{subfigure}{0.19\textwidth}
            \centering
            \includegraphics[width=\linewidth]{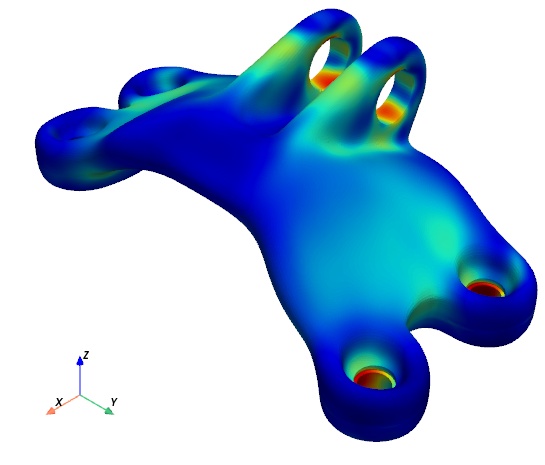}
            \captionsetup{font=small}
            \caption{Pred., 50$^{th}$ pct.}
            \label{fig:figure_10_h}
        \end{subfigure}
        \hfill
        \begin{subfigure}{0.19\textwidth}
            \centering
            \includegraphics[width=\linewidth]{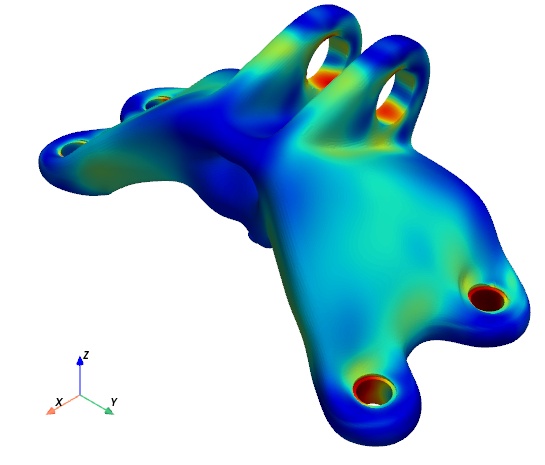}
            \captionsetup{font=small}
            \caption{Pred., 75$^{th}$ pct.}
            \label{fig:figure_10_i}
        \end{subfigure}
        \hfill
        \begin{subfigure}{0.19\textwidth}
            \centering
            \includegraphics[width=\linewidth]{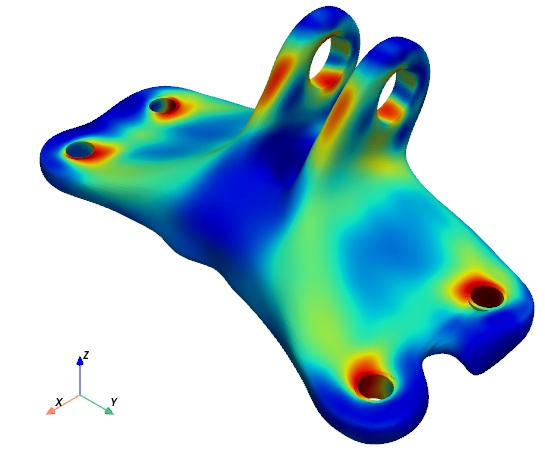}
            \captionsetup{font=small}
            \caption{Pred., worst.}
            \label{fig:figure_10_j}
        \end{subfigure}
        
        \begin{subfigure}{0.185\textwidth}
            \centering
            \includegraphics[width=\linewidth]{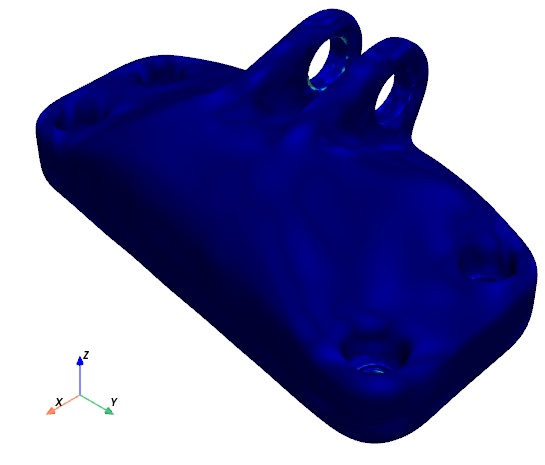}
            \captionsetup{font=small}
            \caption{MAE: 3.35}
            \label{fig:figure_10_k}
        \end{subfigure}
        \hfill
        \begin{subfigure}{0.19\textwidth}
            \centering
            \includegraphics[width=\linewidth]{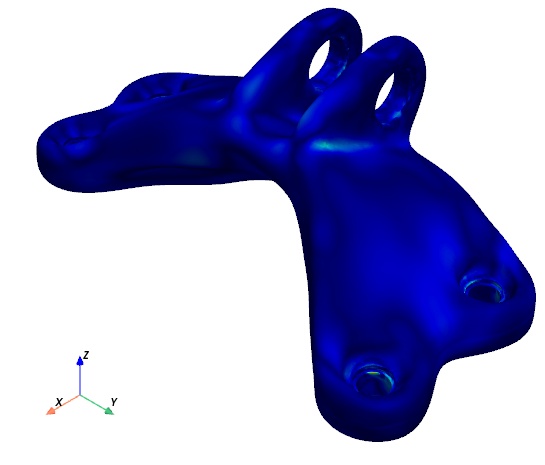}
            \captionsetup{font=small}
            \caption{MAE: 10.64}
            \label{fig:figure_10_l}
        \end{subfigure}
        \hfill
        \begin{subfigure}{0.2\textwidth}
            \centering
            \includegraphics[width=\linewidth]{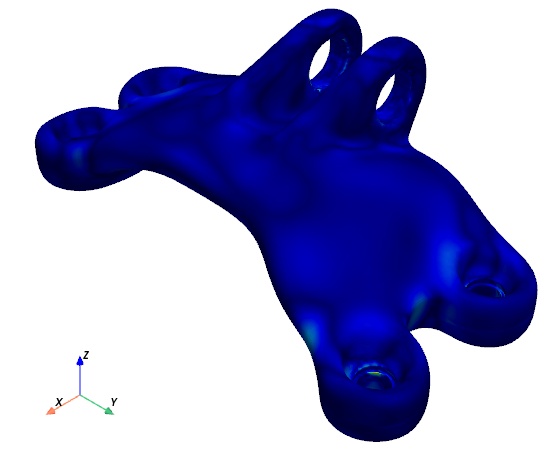}
            \captionsetup{font=small}
            \caption{MAE: 14.69}
            \label{fig:figure_10_m}
        \end{subfigure}
        \hfill
        \begin{subfigure}{0.195\textwidth}
            \centering
            \includegraphics[width=\linewidth]{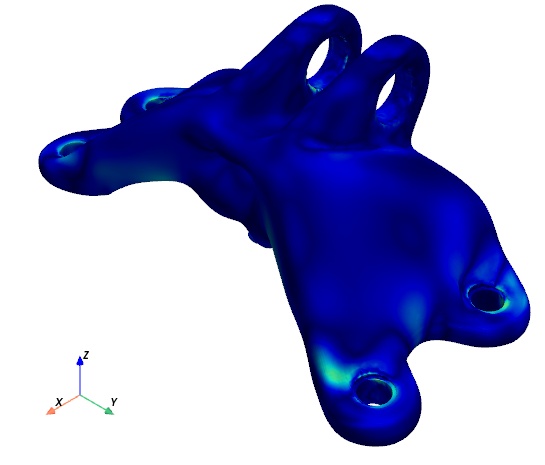}
            \captionsetup{font=small}
            \caption{MAE: 20.78}
            \label{fig:figure_10_n}
        \end{subfigure}
        \hfill
        \begin{subfigure}{0.19\textwidth}
            \centering
            \includegraphics[width=\linewidth]{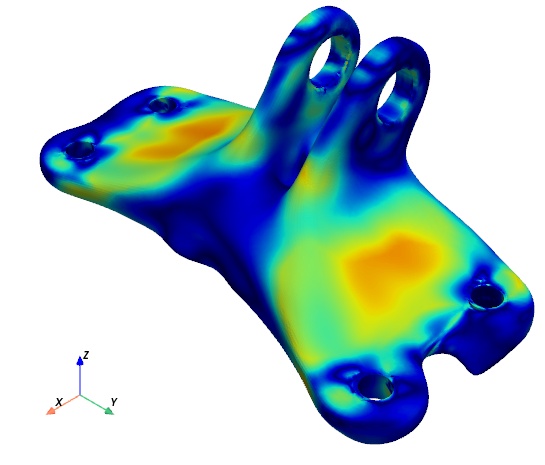}
            \captionsetup{font=small}
            \caption{MAE: 50.07}
            \label{fig:figure_10_o}
        \end{subfigure}
    \end{subfigure}
    
    \includegraphics[width=0.7\textwidth]{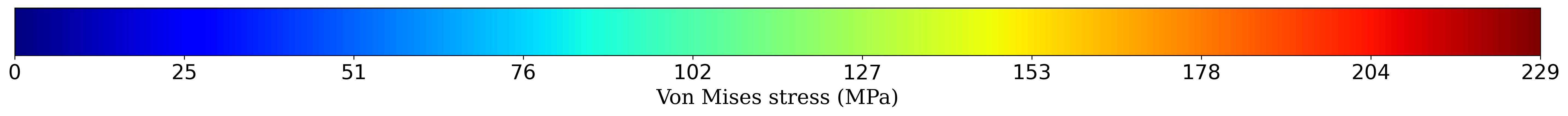}
    
    \caption{Comparison of DeepONet predictions and FEA results under horizontal load (von Mises stress). Top row: FEA results; middle row: DeepONet predictions; bottom row: absolute errors with MAE values.}
    \label{fig:figure_10}
\end{figure}

\begin{figure}[htbp]
    \centering
    \begin{subfigure}{0.78\textwidth}
        \centering
        \begin{subfigure}{0.19\textwidth}
            \centering
            \includegraphics[width=\linewidth]{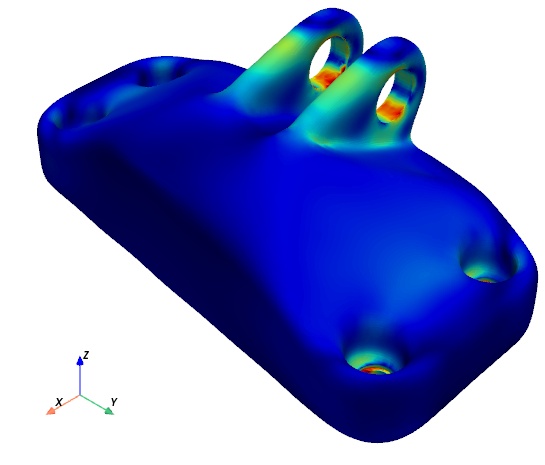}
            \captionsetup{font=small}
            \caption{FE, best.}
            \label{fig:figure_11_a}
        \end{subfigure}
        \hfill
        \begin{subfigure}{0.19\textwidth}
            \centering
            \includegraphics[width=\linewidth]{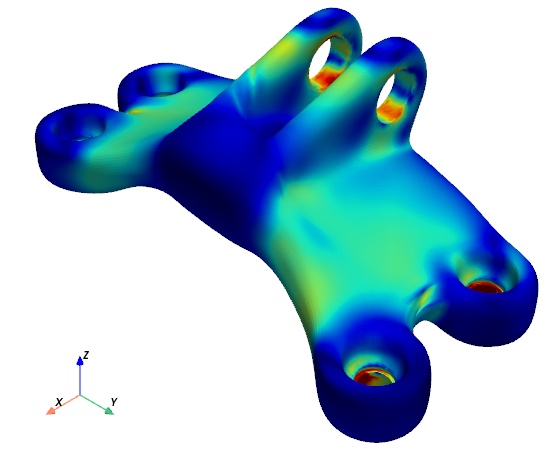}
            \captionsetup{font=small}
            \caption{FE, 25$^{th}$ pct.}
            \label{fig:figure_11_b}
        \end{subfigure}
        \hfill
        \begin{subfigure}{0.19\textwidth}
            \centering
            \includegraphics[width=\linewidth]{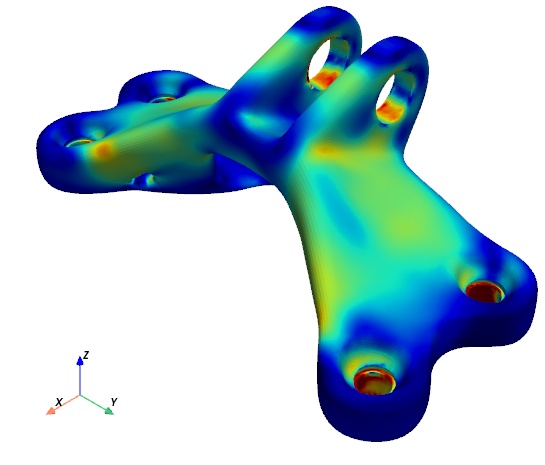}
            \captionsetup{font=small}
            \caption{FE, 50$^{th}$ pct.}
            \label{fig:figure_11_c}
        \end{subfigure}
        \hfill
        \begin{subfigure}{0.2\textwidth}
            \centering
            \includegraphics[width=\linewidth]{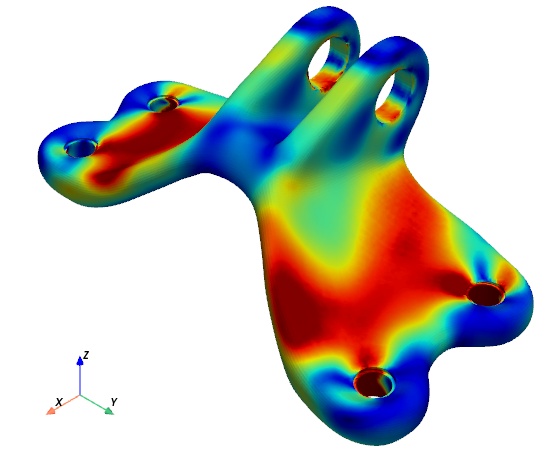}
            \captionsetup{font=small}
            \caption{FE, 75$^{th}$ pct.}
            \label{fig:figure_11_d}
        \end{subfigure}
        \hfill
        \begin{subfigure}{0.19\textwidth}
            \centering
            \includegraphics[width=\linewidth]{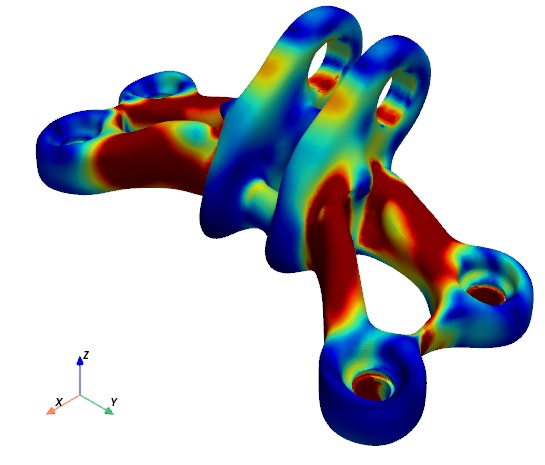}
            \captionsetup{font=small}
            \caption{FE, worst.}
            \label{fig:figure_11_e}
        \end{subfigure}
        
        \begin{subfigure}{0.19\textwidth}
            \centering
            \includegraphics[width=\linewidth]{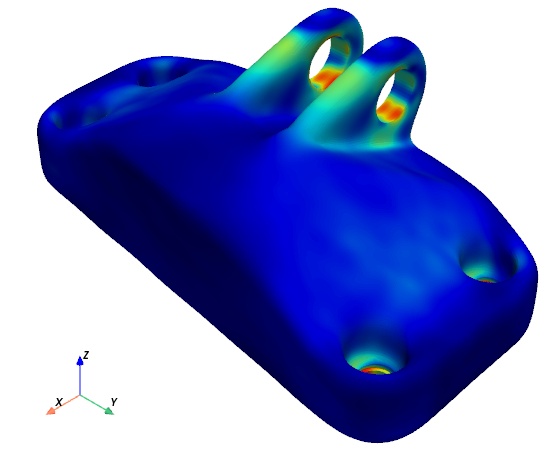}
            \captionsetup{font=small}
            \caption{Pred., best.}
            \label{fig:figure_11_f}
        \end{subfigure}
        \hfill
        \begin{subfigure}{0.19\textwidth}
            \centering
            \includegraphics[width=\linewidth]{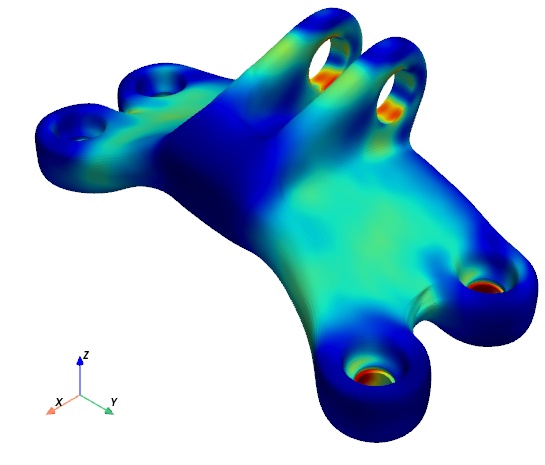}
            \captionsetup{font=small}
            \caption{Pred., 25$^{th}$ pct.}
            \label{fig:figure_11_g}
        \end{subfigure}
        \hfill
        \begin{subfigure}{0.19\textwidth}
            \centering
            \includegraphics[width=\linewidth]{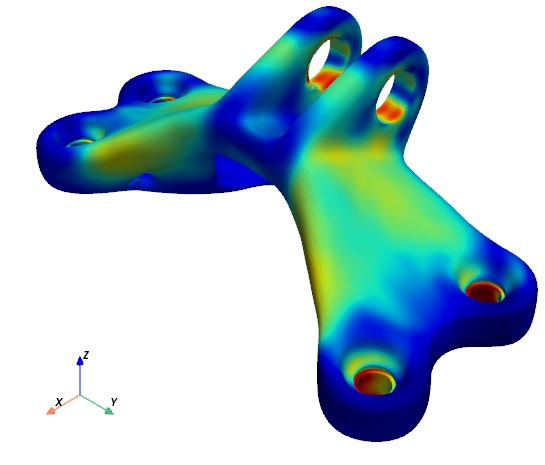}
            \captionsetup{font=small}
            \caption{Pred., 50$^{th}$ pct.}
            \label{fig:figure_11_h}
        \end{subfigure}
        \hfill
        \begin{subfigure}{0.19\textwidth}
            \centering
            \includegraphics[width=\linewidth]{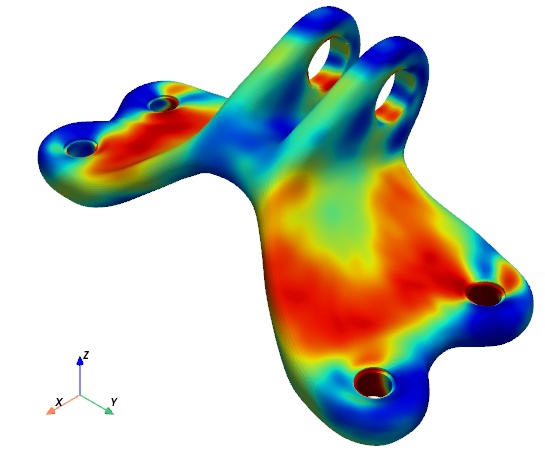}
            \captionsetup{font=small}
            \caption{Pred. 75th Pct.}
            \label{fig:figure_11_i}
        \end{subfigure}
        \hfill
        \begin{subfigure}{0.19\textwidth}
            \centering
            \includegraphics[width=\linewidth]{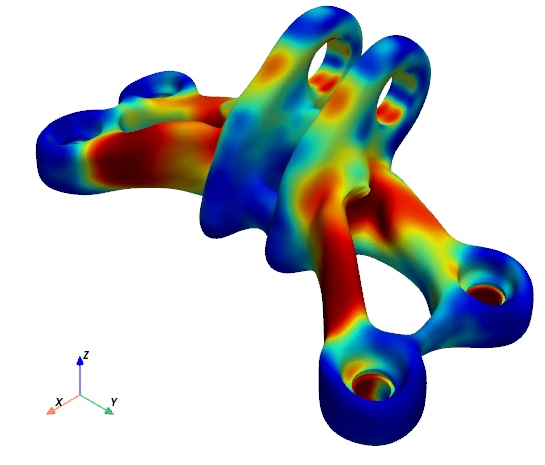}
            \captionsetup{font=small}
            \caption{Pred., worst.}
            \label{fig:figure_11_j}
        \end{subfigure}
        
        \begin{subfigure}{0.19\textwidth}
            \centering
            \includegraphics[width=\linewidth]{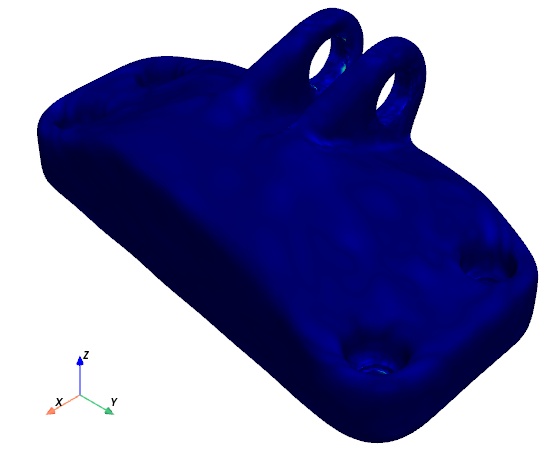}
            \captionsetup{font=small}
            \caption{MAE: 2.56}
            \label{fig:figure_11_k}
        \end{subfigure}
        \hfill
        \begin{subfigure}{0.19\textwidth}
            \centering
            \includegraphics[width=\linewidth]{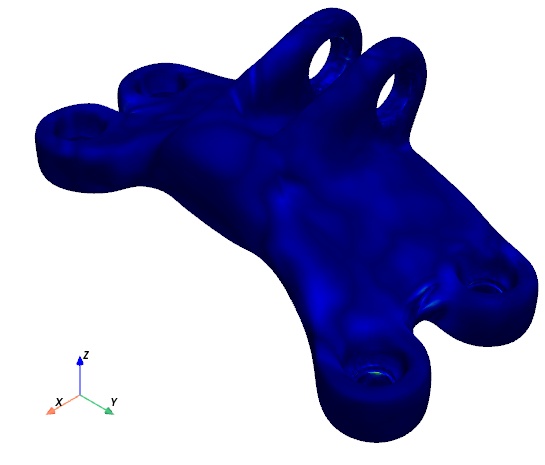}
            \captionsetup{font=small}
            \caption{MAE: 4.76}
            \label{fig:figure_11_l}
        \end{subfigure}
        \hfill
        \begin{subfigure}{0.19\textwidth}
            \centering
            \includegraphics[width=\linewidth]{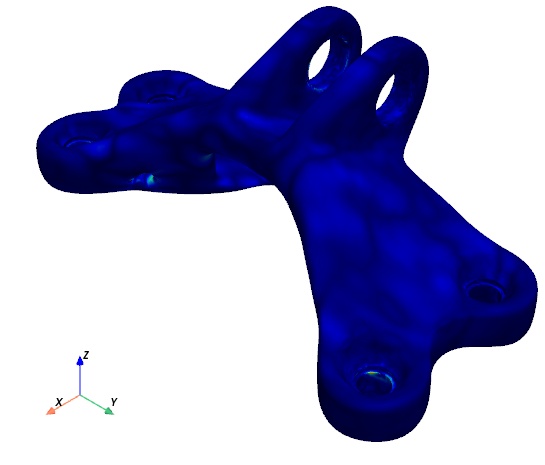}
            \captionsetup{font=small}
            \caption{MAE: 6.65}
            \label{fig:figure_11_m}
        \end{subfigure}
        \hfill
        \begin{subfigure}{0.2\textwidth}
            \centering
            \includegraphics[width=\linewidth]{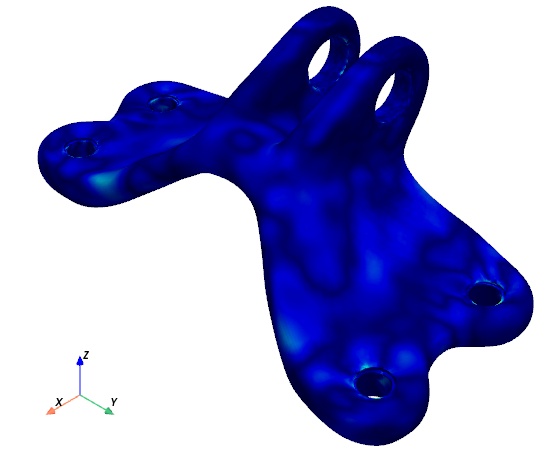}
            \captionsetup{font=small}
            \caption{MAE: 9.94}
            \label{fig:figure_11_n}
        \end{subfigure}
        \hfill
        \begin{subfigure}{0.19\textwidth}
            \centering
            \includegraphics[width=\linewidth]{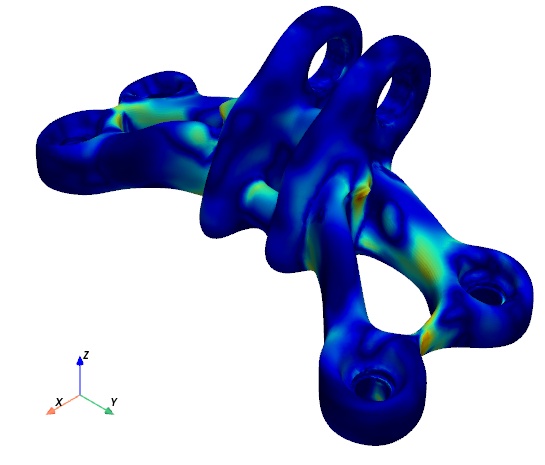}
            \captionsetup[sub]{font=small}
            \caption{MAE: 28.15}
            \label{fig:figure_11_o}
        \end{subfigure}
    \end{subfigure}
    
    \includegraphics[width=0.7\textwidth]{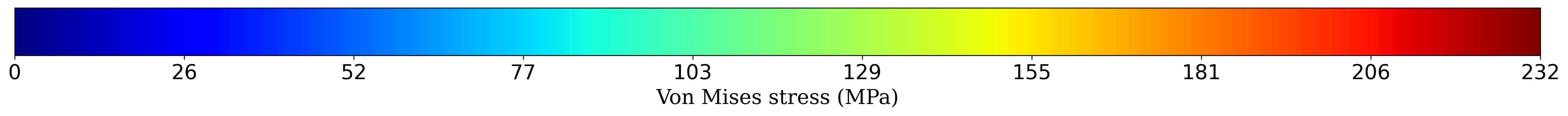}
    
    \caption{Comparison of Point-DeepONet predictions and FEA results under horizontal load (von Mises stress). Top row: FEA results; middle row: Point-DeepONet predictions; bottom row: absolute errors with MAE values.}
    \label{fig:figure_11}
\end{figure}

To provide a visual comparison of the models' predictions, Figures~\ref{fig:figure_9} to \ref{fig:figure_11} present detailed visualizations of the finite element analysis (FEA) results and the corresponding predictions from Point-DeepONet and DeepONet for selected cases. Each figure focuses on a specific load case and stress component, illustrating the FEA results for cases corresponding to the $0^{th}$(best), $25^{th}$, $50^{th}$, $75^{th}$, and $100^{th}$(worst) percentiles of the prediction error distribution. The model predictions for these cases are shown alongside the FEA results, facilitating direct comparison.

Additionally, each figure includes a common colorbar representing the von Mises stress in megapascals (MPa), which is consistent across the FEA results, model predictions, and absolute error distributions. This uniform color scale aids in accurately interpreting and comparing the stress values across different cases. The mean absolute error (MAE) is reported for each case, providing a quantitative measure of prediction accuracy. These visualizations reveal that Point-DeepONet closely approximates the FEA results across a range of error levels, with higher errors typically occurring in regions with complex stress gradients or geometric features.

Figures~\ref{fig:figure_9}, \ref{fig:figure_10}, and \ref{fig:figure_11} specifically illustrate the performance of the PointNet, DeepONet, and Point-DeepONet models under a horizontal load case for von Mises stress. Although the models were trained on a comprehensive set of 12 distinct cases—covering three load directions (vertical, horizontal, and diagonal) and four components ($u_x$, $u_y$, $u_z$, and von Mises stress)—these figures focus on the horizontal load scenario to provide a clear example of the models' predictive capabilities.

Figure~\ref{fig:figure_9} compares the PointNet predictions with the FEA results. The top row displays the FEA simulations at various percentiles, the middle row presents the corresponding PointNet predictions, and the bottom row shows the absolute error distributions with MAE values. The results indicate that PointNet captures the von Mises stress distribution reasonably well across most percentiles. However, higher errors are observed in regions with extreme stress concentrations, suggesting areas for potential improvement.

An analysis of the error plots in Figures 9-11 indicates that the largest prediction errors are consistently localized in regions of sharp geometric discontinuity and high stress concentration. This highlights a limitation in resolving high-frequency physical phenomena, a common challenge for neural operators and an important direction for future work.

\subsection{Influence of Resampling Size on Displacement and Stress Distribution}
\label{subsec:resampling_effects}

\begin{figure}[!h]
    \centering
    \begin{subfigure}{0.8\textwidth}
        \centering
        \begin{subfigure}[b]{0.19\textwidth}
            \centering
            \includegraphics[width=\textwidth]{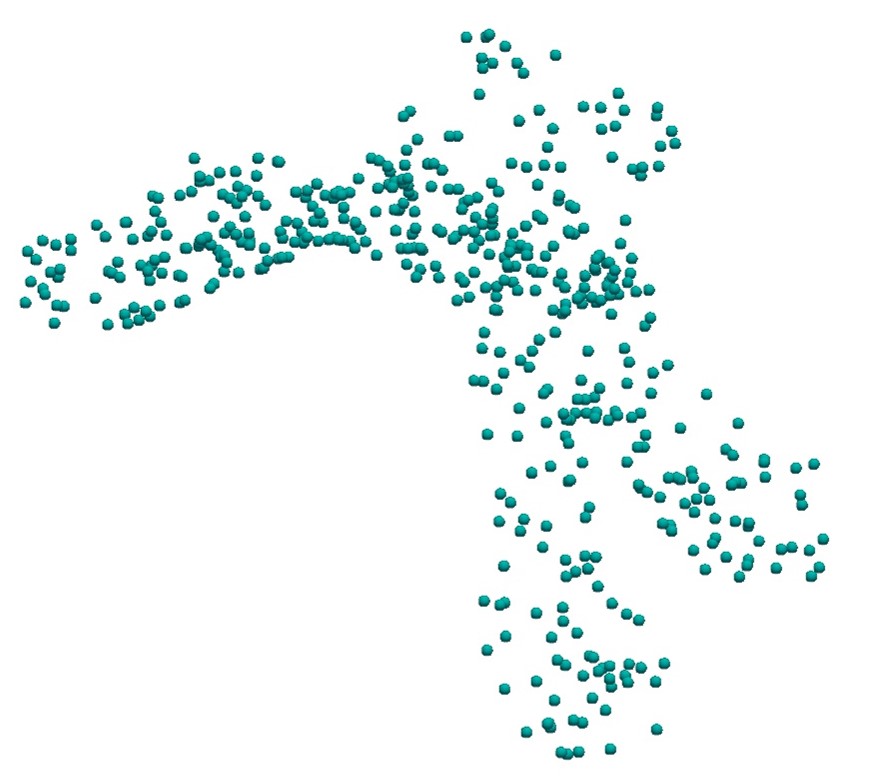}
            \caption{}
            \label{fig:figure_12_a}
        \end{subfigure}
        \begin{subfigure}[b]{0.19\textwidth}
            \centering
            \includegraphics[width=\textwidth]{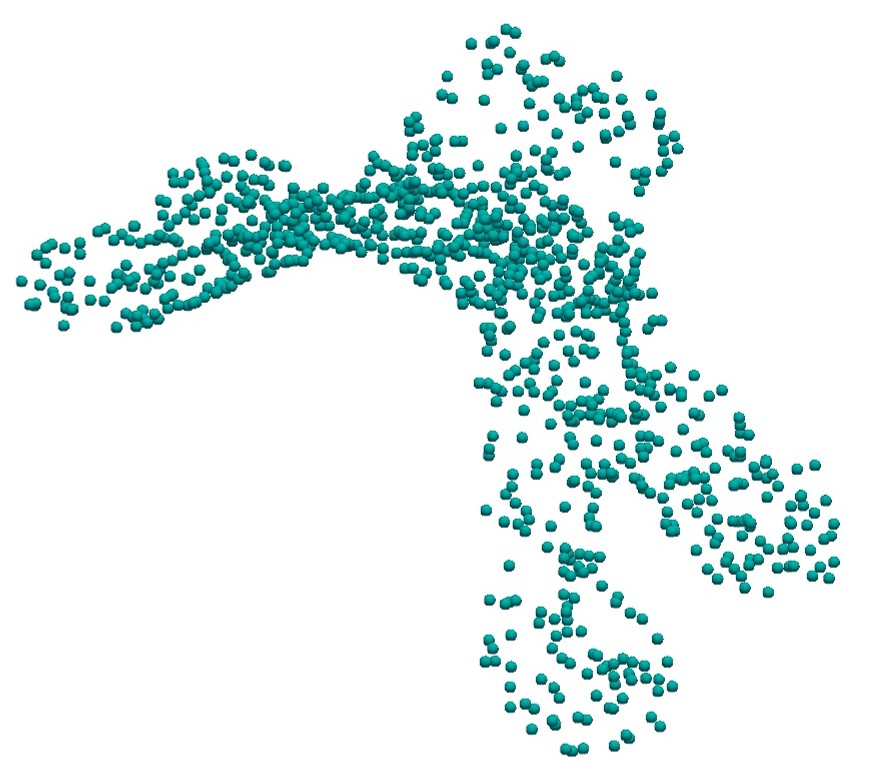}
            \caption{}
            \label{fig:figure_12_b}
        \end{subfigure}
        \begin{subfigure}[b]{0.19\textwidth}
            \centering
            \includegraphics[width=\textwidth]{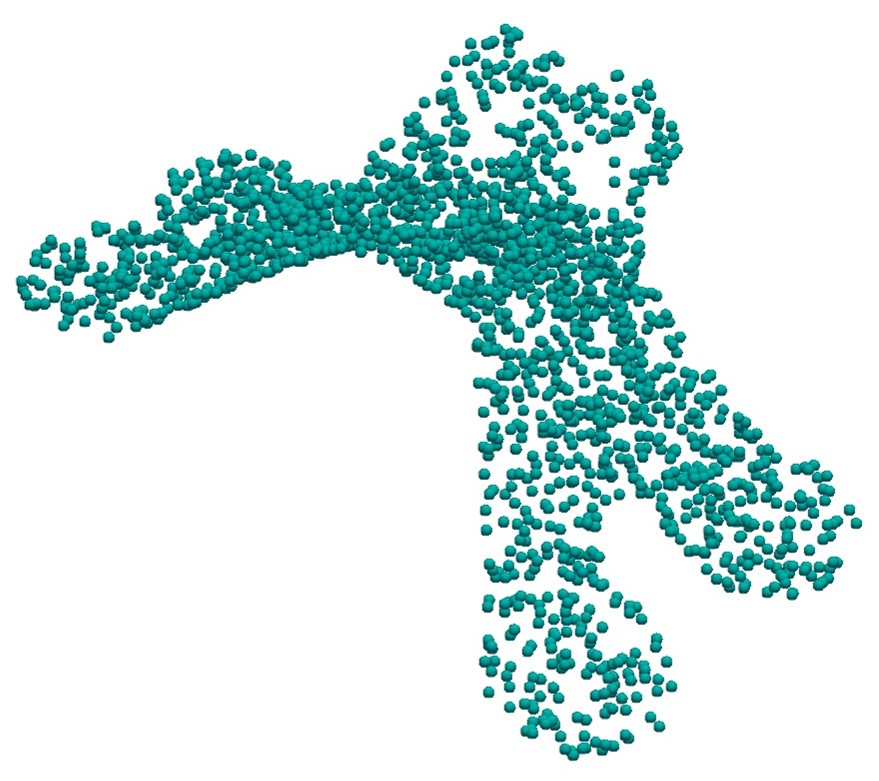}
            \caption{}
            \label{fig:figure_12_c}
        \end{subfigure}
        \begin{subfigure}[b]{0.19\textwidth}
            \centering
            \includegraphics[width=\textwidth]{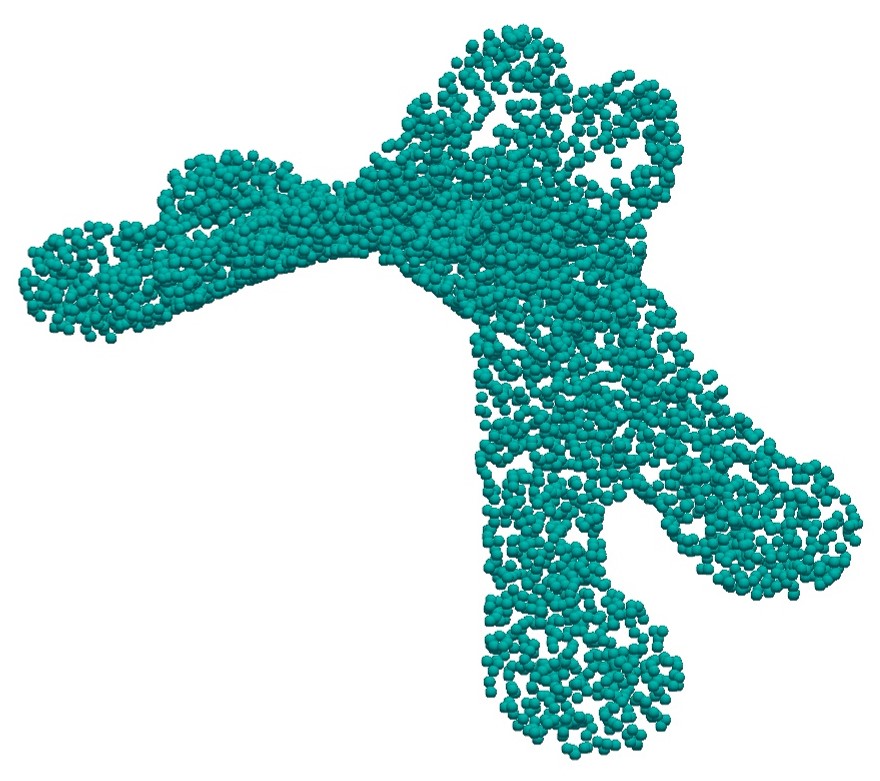}
            \caption{}
            \label{fig:figure_12_d}
        \end{subfigure}
        \begin{subfigure}[b]{0.19\textwidth}
            \centering
            \includegraphics[width=\textwidth]{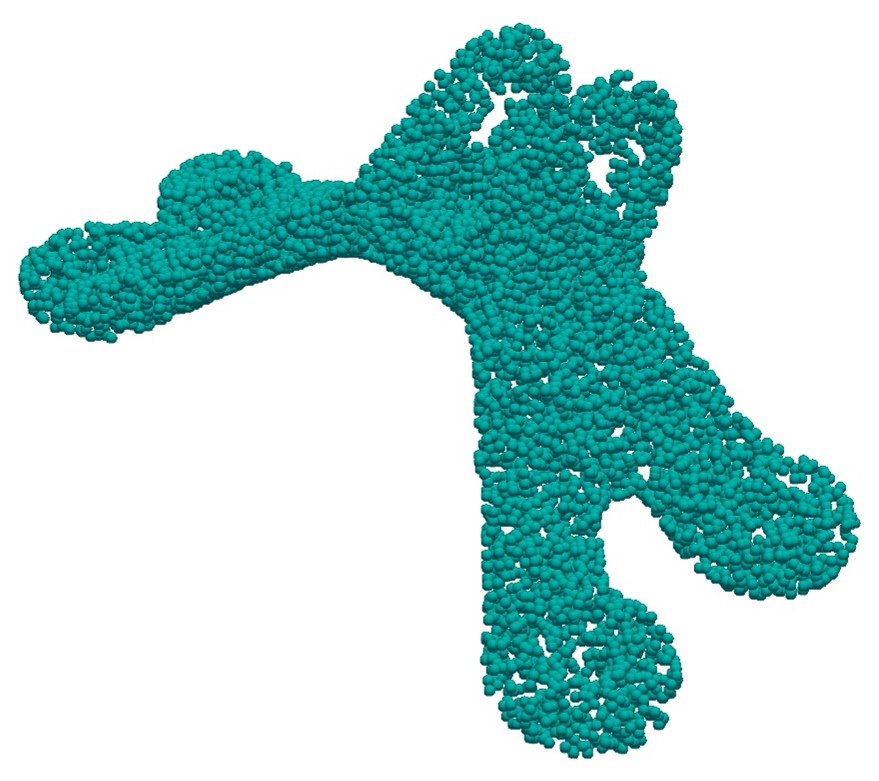}
            \caption{}
            \label{fig:figure_12_e}
        \end{subfigure}
    \end{subfigure}
    
    \caption{Dataset with an average node count of 200,000 in the original nonlinear finite element analysis mesh, sampled at various levels for reduced complexity. (a) $N=500$ (0.25\%). (b) $N=1,000$ (0.5\%). (c) $N=2,000$ (1\%). (d) $N=5,000$ (2.5\%). (e) $N=10,000$ (5\%).}
    \label{fig:figure_12}
\end{figure}

To reduce computational complexity while preserving essential geometric details, the original high-resolution meshes were downsampled at various levels, as shown in Figure~\ref{fig:figure_12}. Starting from an average node count of approximately 200,000 in the original finite element analysis (FEA) mesh, samples were created at different percentages of the total nodes: 500 nodes (0.25\%), 1,000 nodes (0.5\%), 2,000 nodes (1\%), 5,000 nodes (2.5\%), and 10,000 nodes (5\%). This multilevel sampling approach provides a balance between model accuracy and computational efficiency, enabling effective learning across a range of mesh complexities. As explored in Section~\ref{subsec:resampling_effects}, we further analyze the impact of these different sampling levels on the deep learning model's performance and training time, demonstrating the trade-offs between computational cost and predictive accuracy.

\begin{figure}[!h]
    \centering
    \begin{subfigure}{0.245\textwidth}
        \centering
        \includegraphics[width=\linewidth]{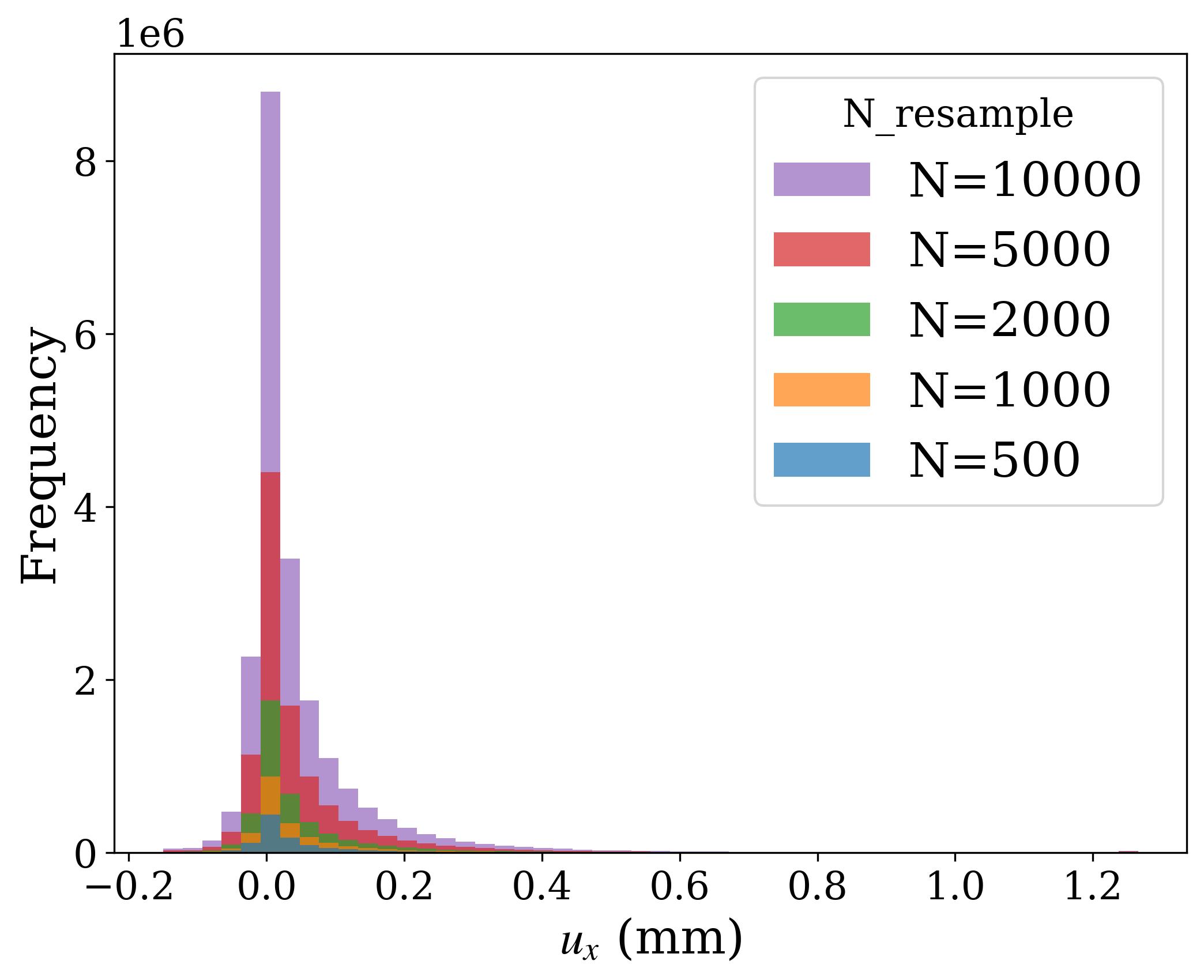}
        \caption{}
        \label{fig:figure_13_a}
    \end{subfigure}%
    \begin{subfigure}{0.245\textwidth}
        \centering
        \includegraphics[width=\linewidth]{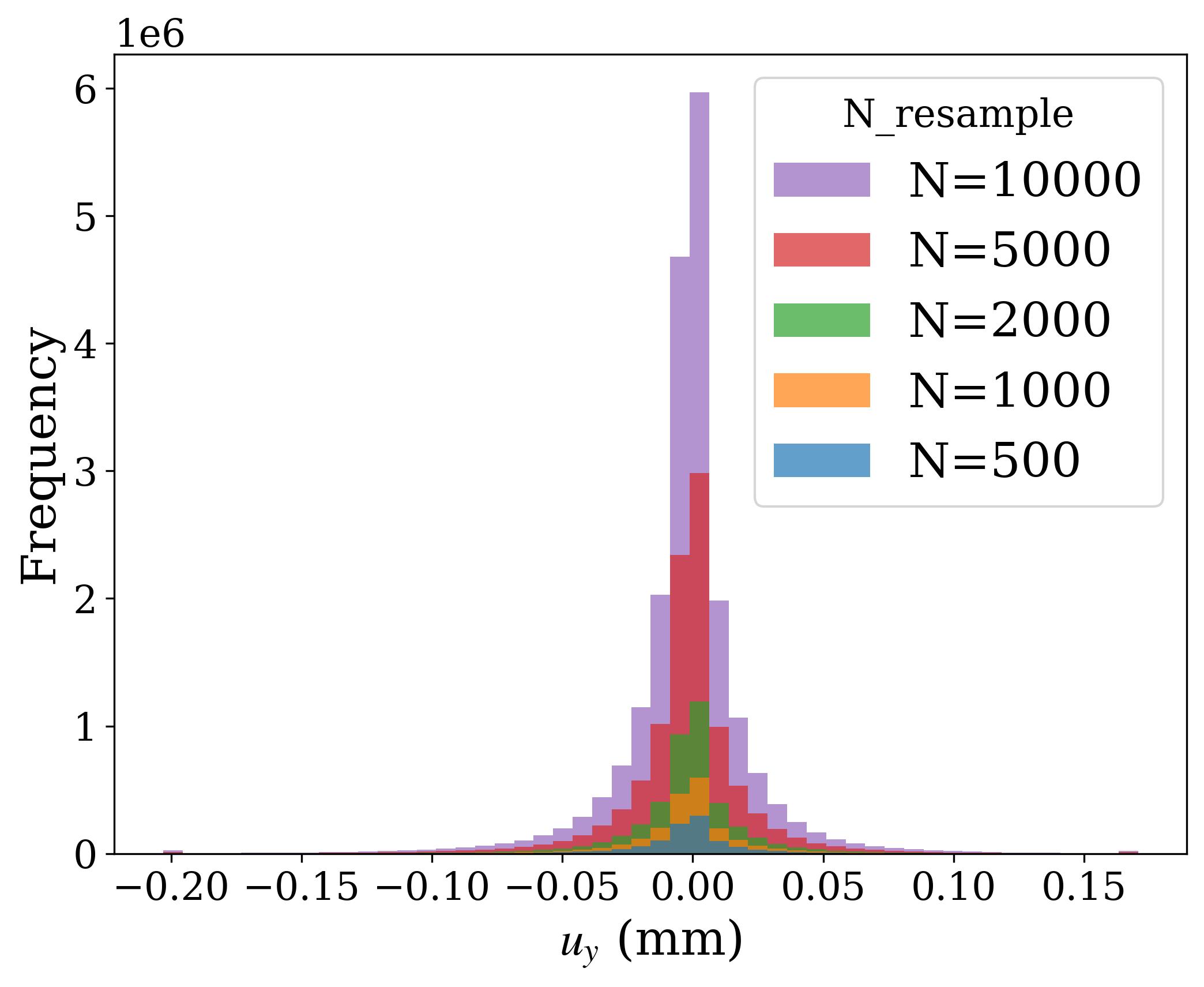}
        \caption{}
        \label{fig:figure_13_b}
    \end{subfigure}
    \begin{subfigure}{0.245\textwidth}
        \centering
        \includegraphics[width=\linewidth]{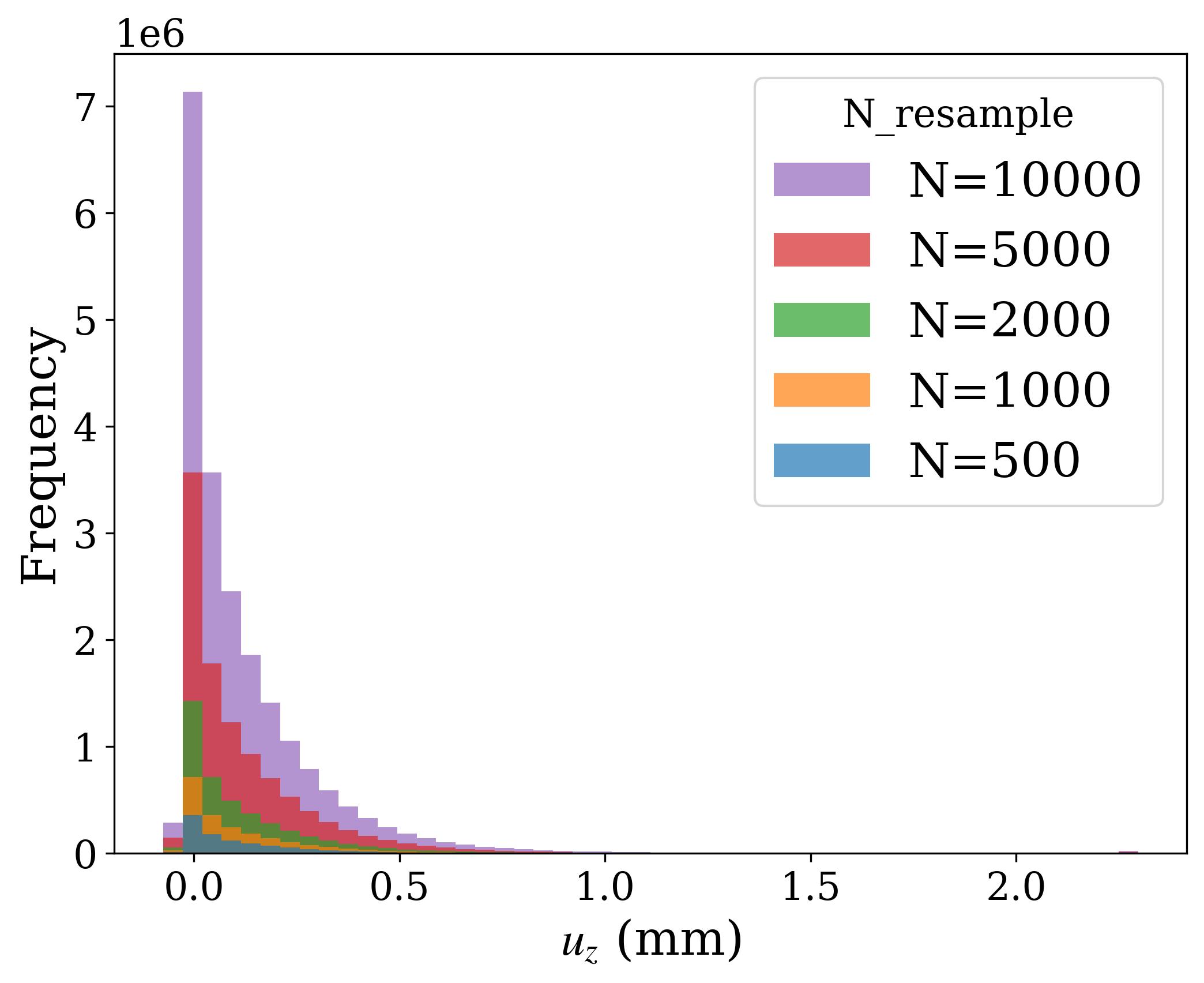}
        \caption{}
        \label{fig:figure_13_c}
    \end{subfigure}
    \begin{subfigure}{0.245\textwidth}
        \centering
        \includegraphics[width=\linewidth]{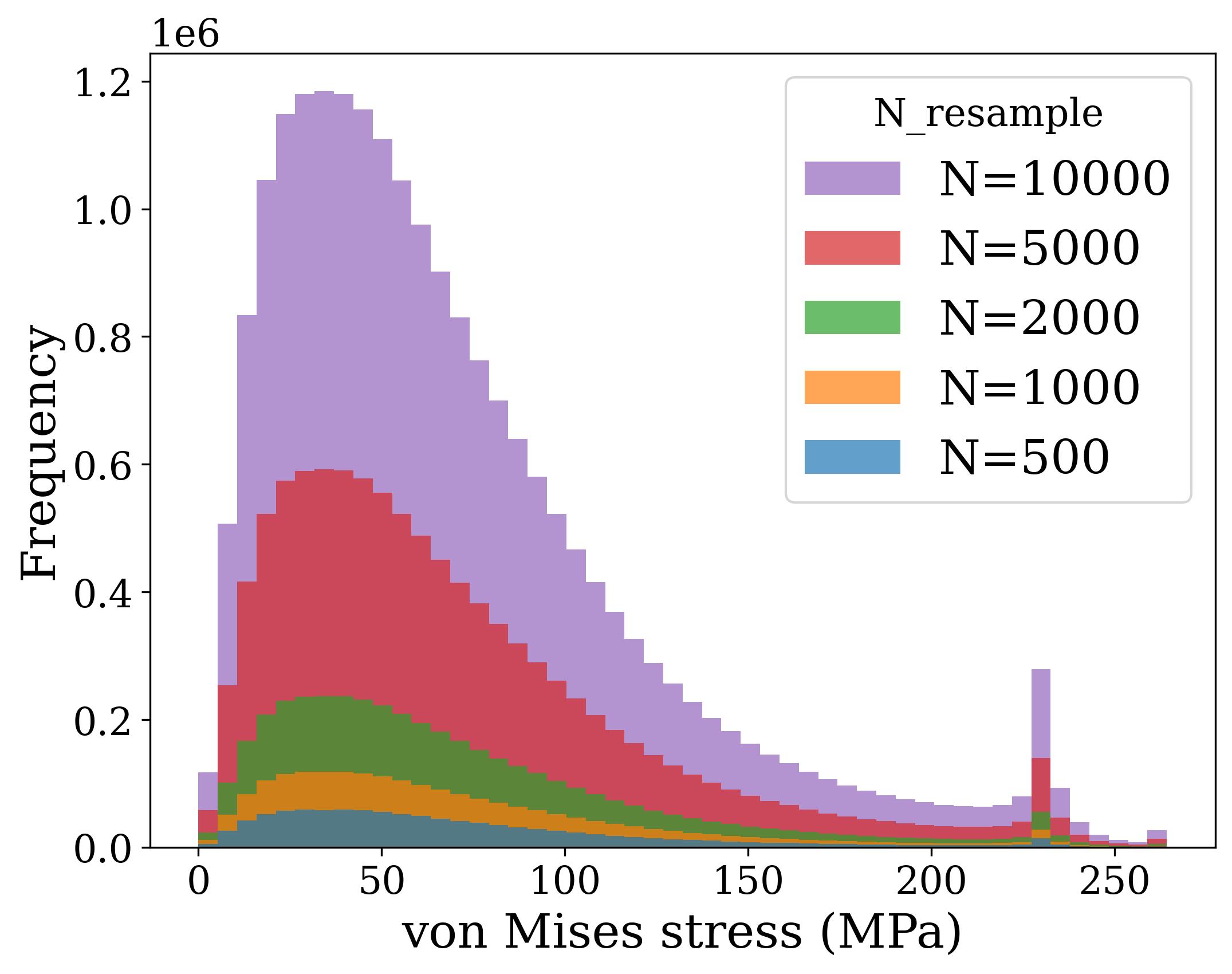}
        \caption{}
        \label{fig:figure_13_d}
    \end{subfigure}
    
    
    \begin{subfigure}{0.245\textwidth}
        \centering
        \includegraphics[width=\linewidth]{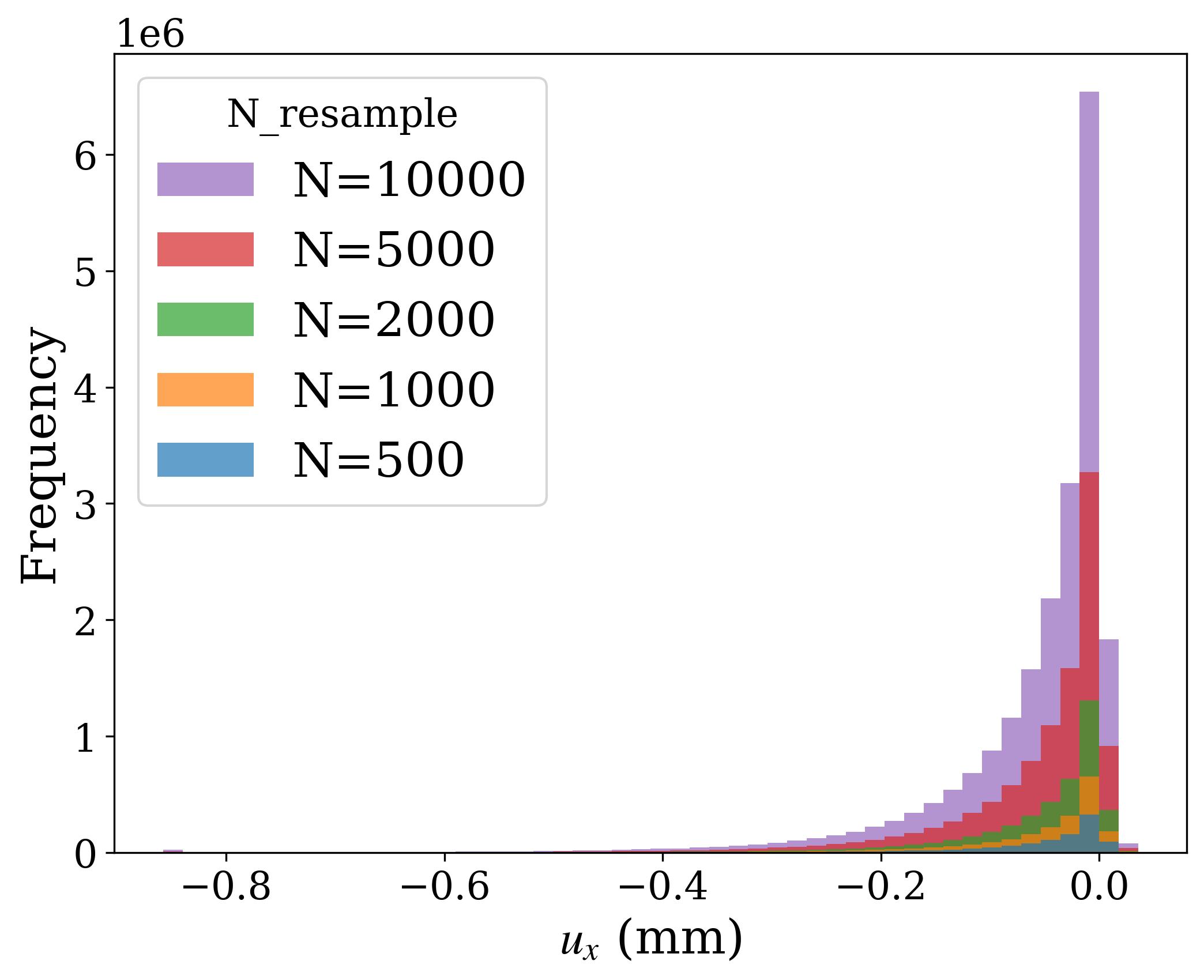}
        \caption{}
        \label{fig:figure_13_e}
    \end{subfigure}%
    \begin{subfigure}{0.245\textwidth}
        \centering
        \includegraphics[width=\linewidth]{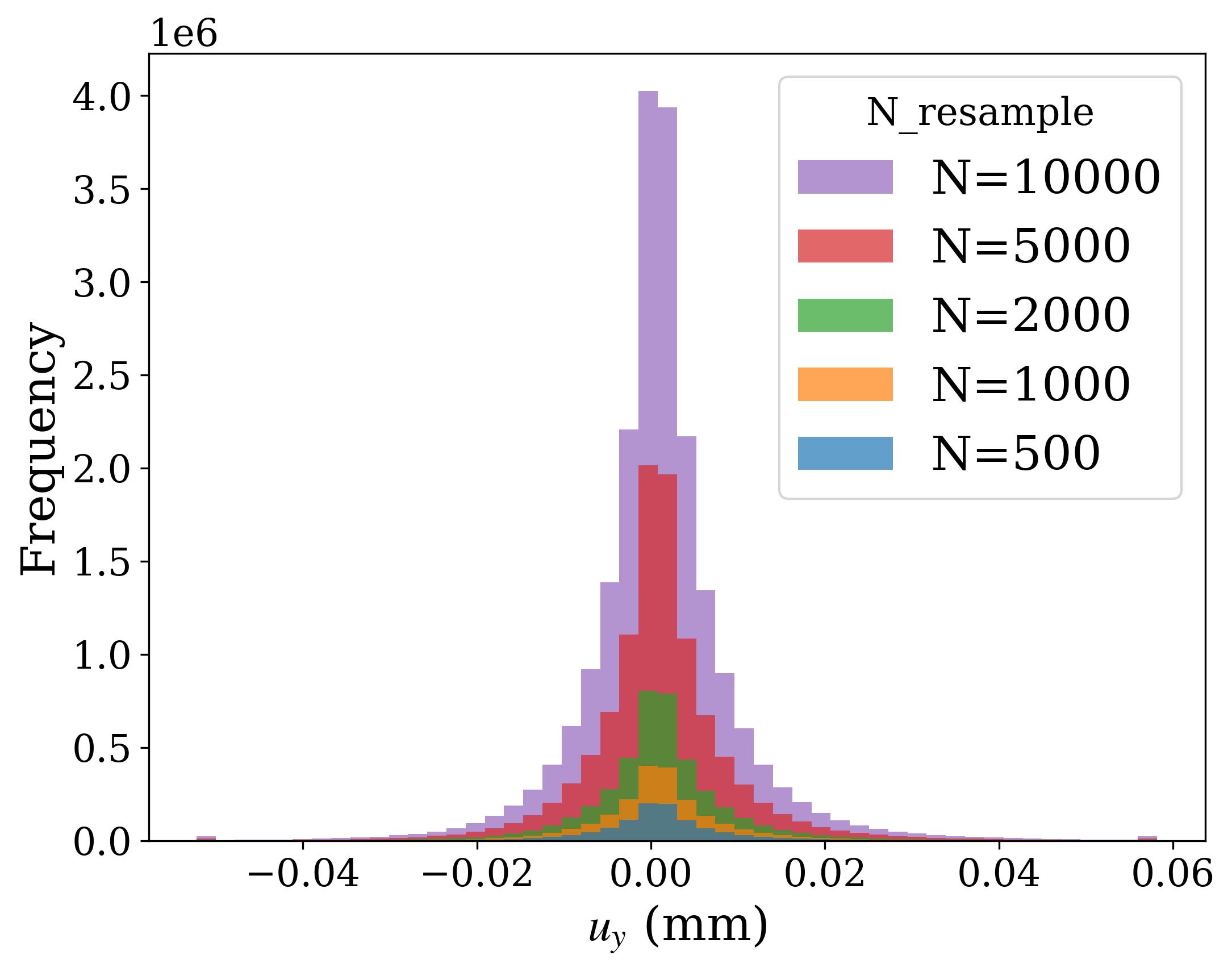}
        \caption{}
        \label{fig:figure_13_f}
    \end{subfigure}%
    \begin{subfigure}{0.245\textwidth}
        \centering
        \includegraphics[width=\linewidth]{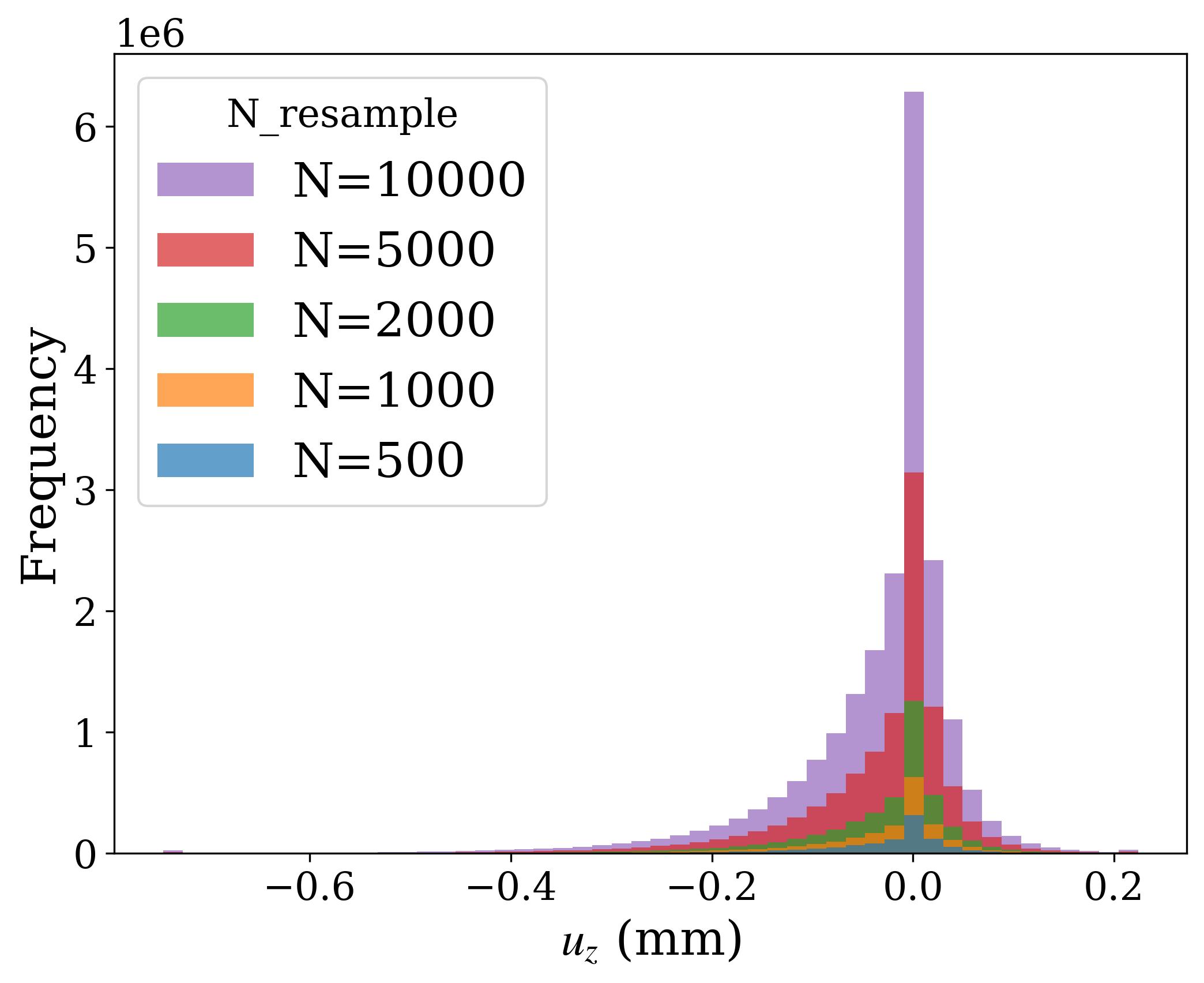}
        \caption{}
        \label{fig:figure_13_g}
    \end{subfigure}%
    \begin{subfigure}{0.245\textwidth}
        \centering
        \includegraphics[width=\linewidth]{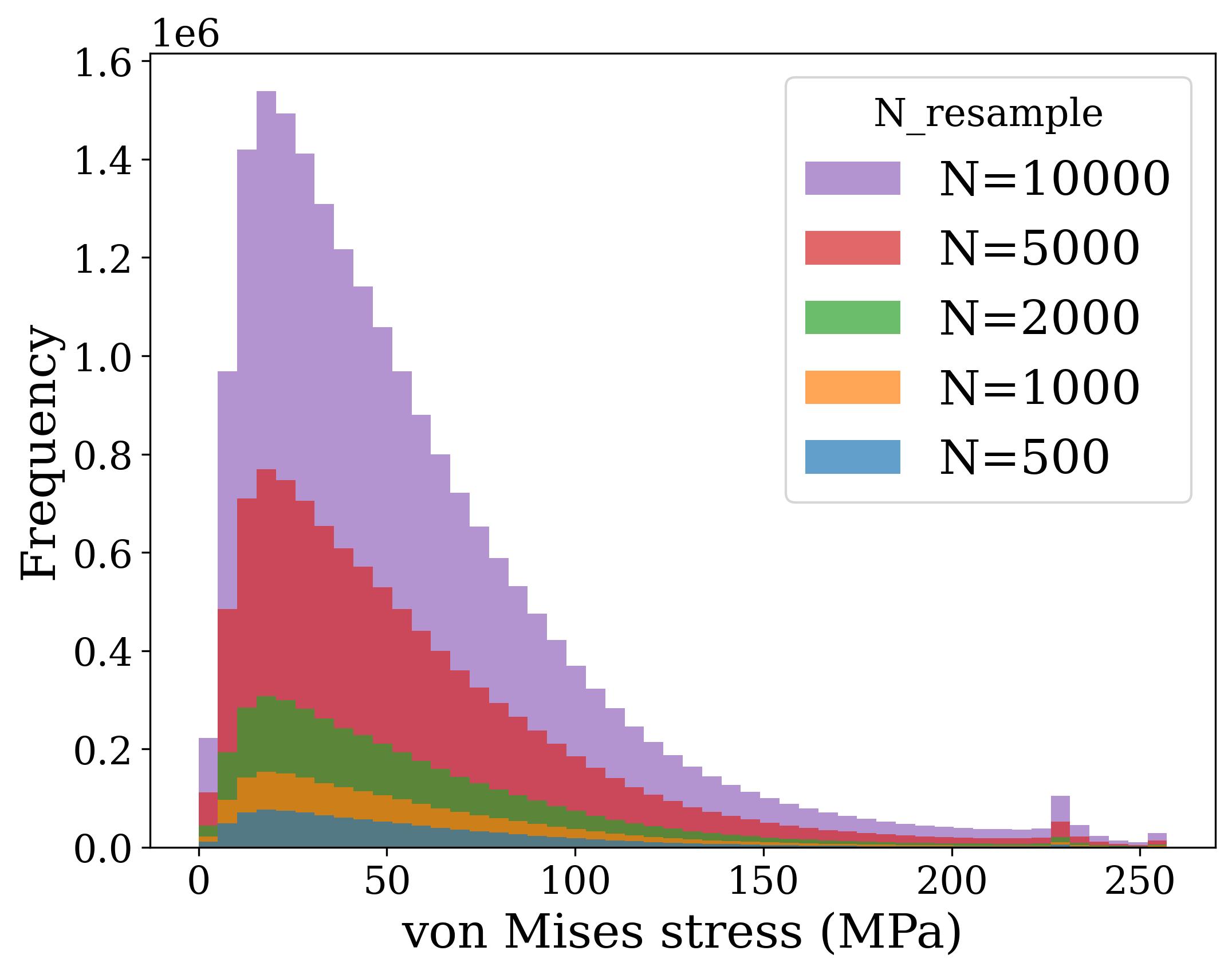}
        \caption{}
        \label{fig:figure_13_h}
    \end{subfigure}
    
    
    \begin{subfigure}{0.245\textwidth}
        \centering
        \includegraphics[width=\linewidth]{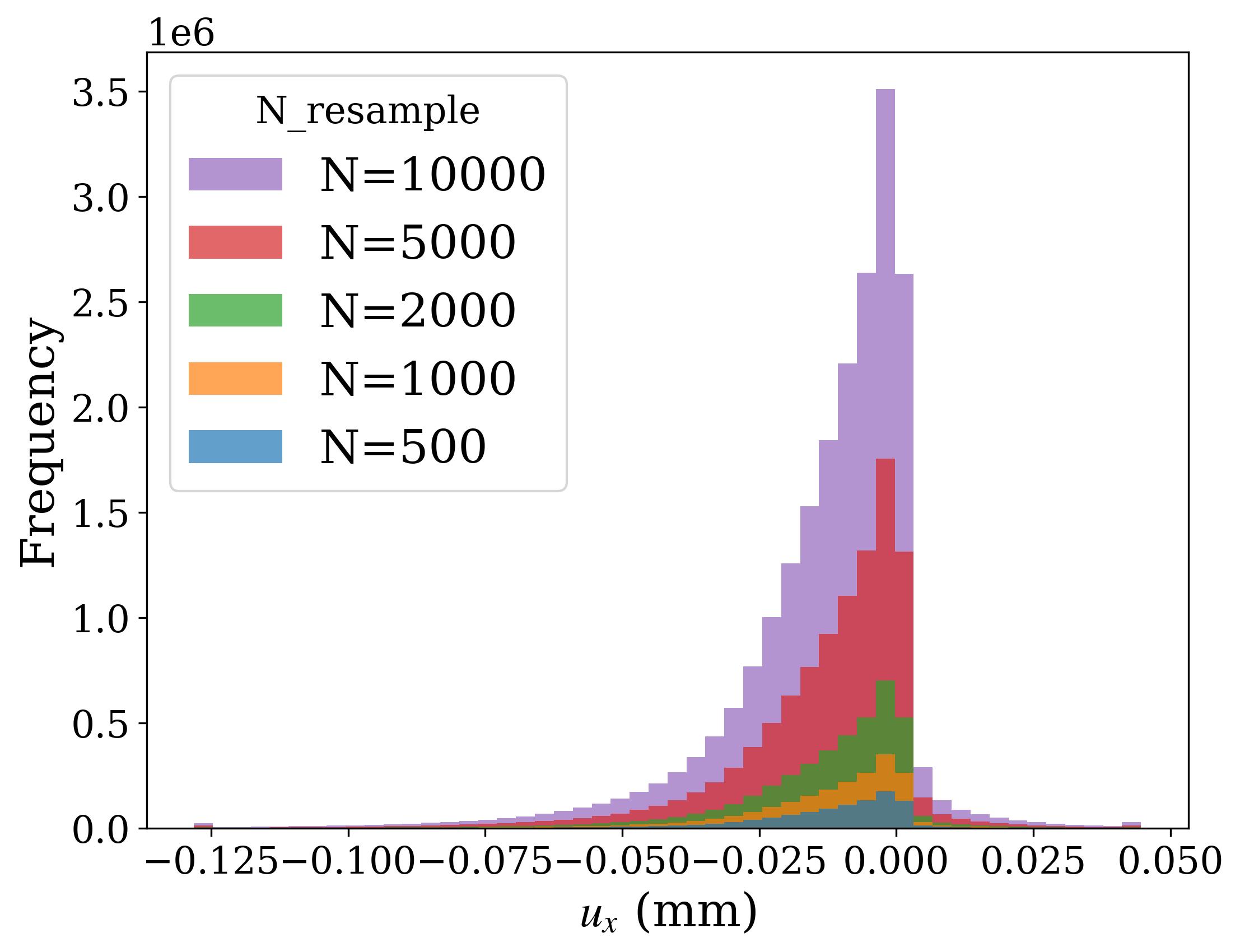}
        \caption{}
        \label{fig:figure_13_i}
    \end{subfigure}%
    \begin{subfigure}{0.245\textwidth}
        \centering
        \includegraphics[width=\linewidth]{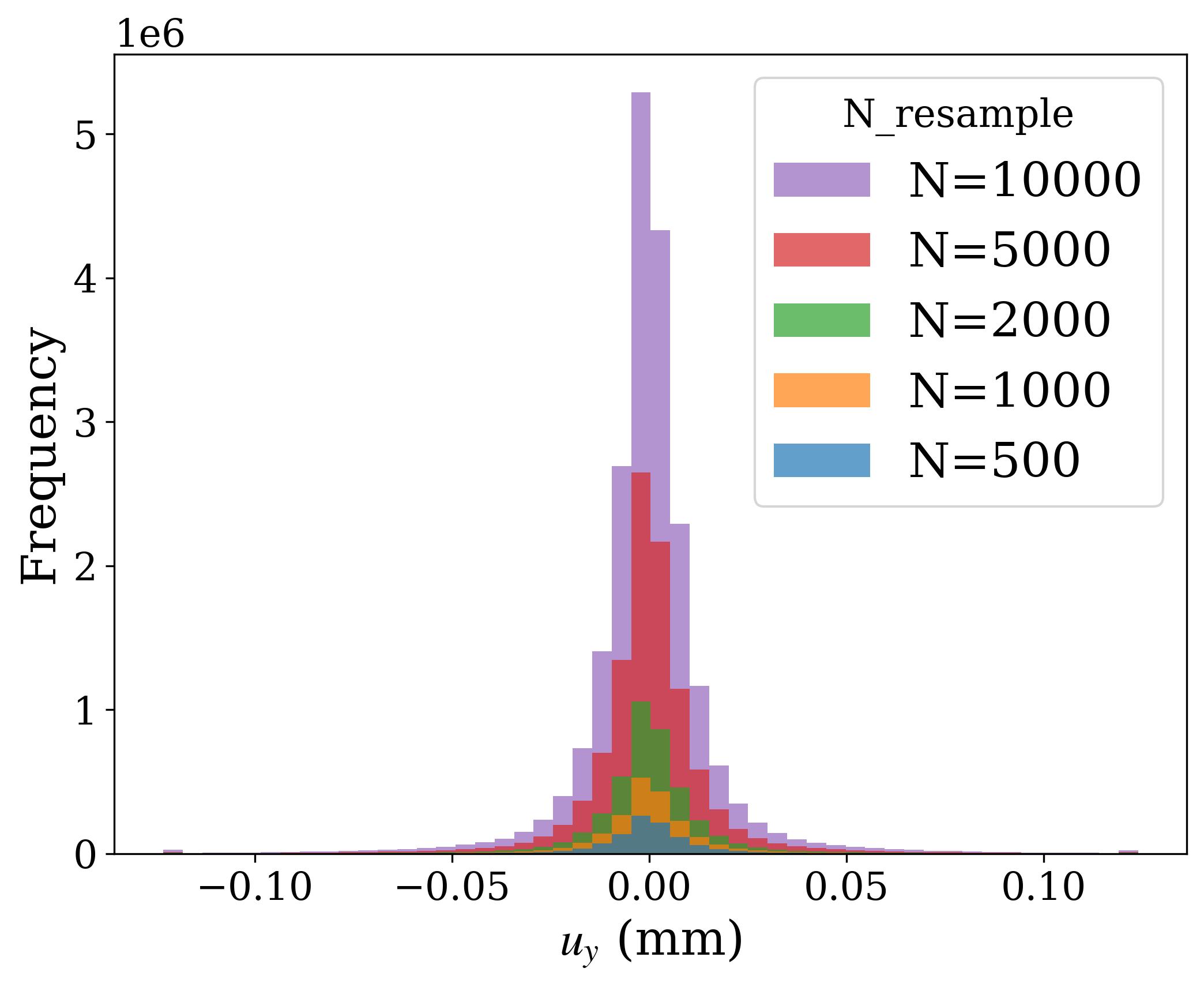}
        \caption{}
        \label{fig:figure_13_j}
    \end{subfigure}%
    \begin{subfigure}{0.245\textwidth}

    \centering
        \includegraphics[width=\linewidth]{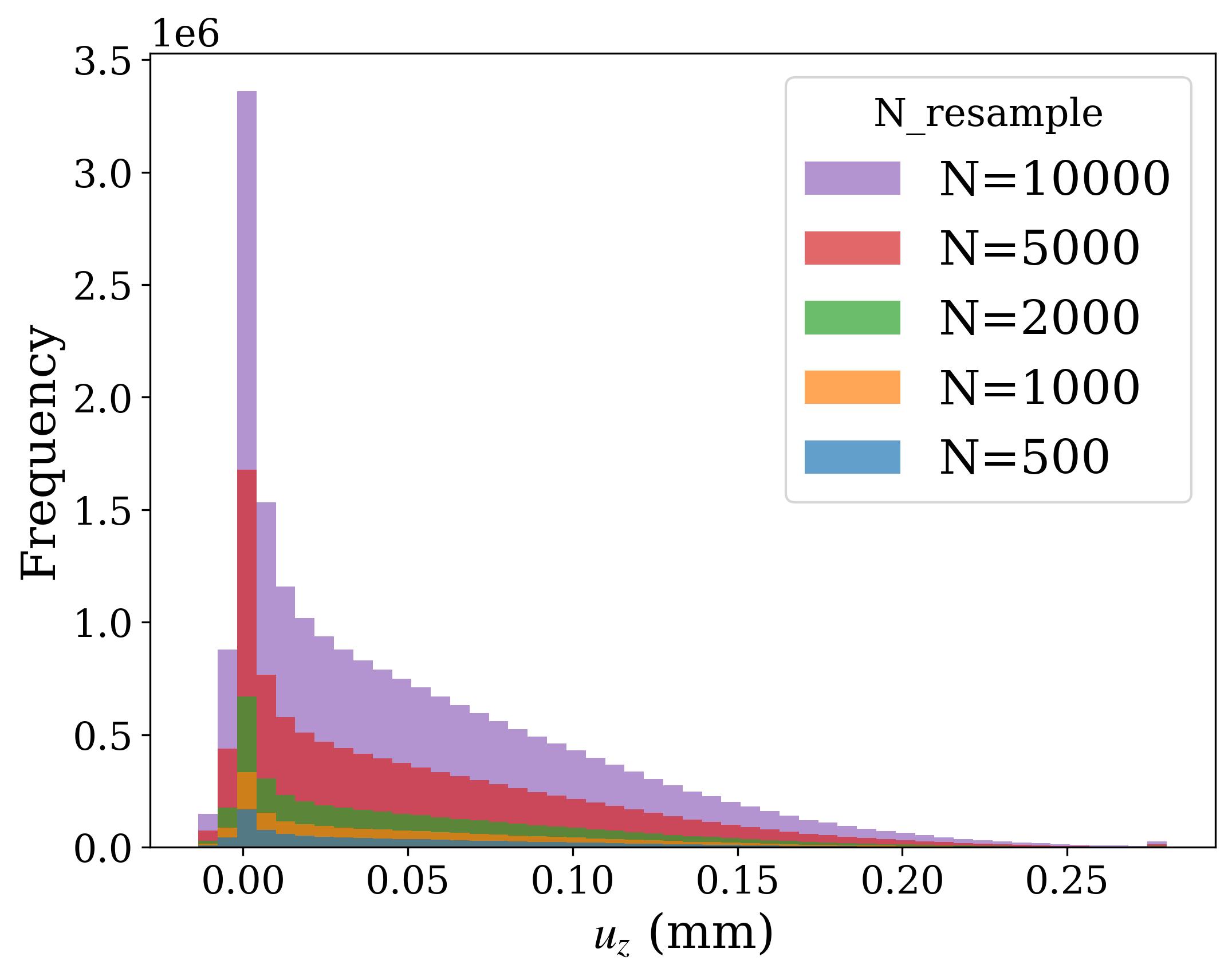}
        \caption{}
        \label{fig:figure_13_k}
    \end{subfigure}%
    \begin{subfigure}{0.245\textwidth}
        \centering
        \includegraphics[width=\linewidth]{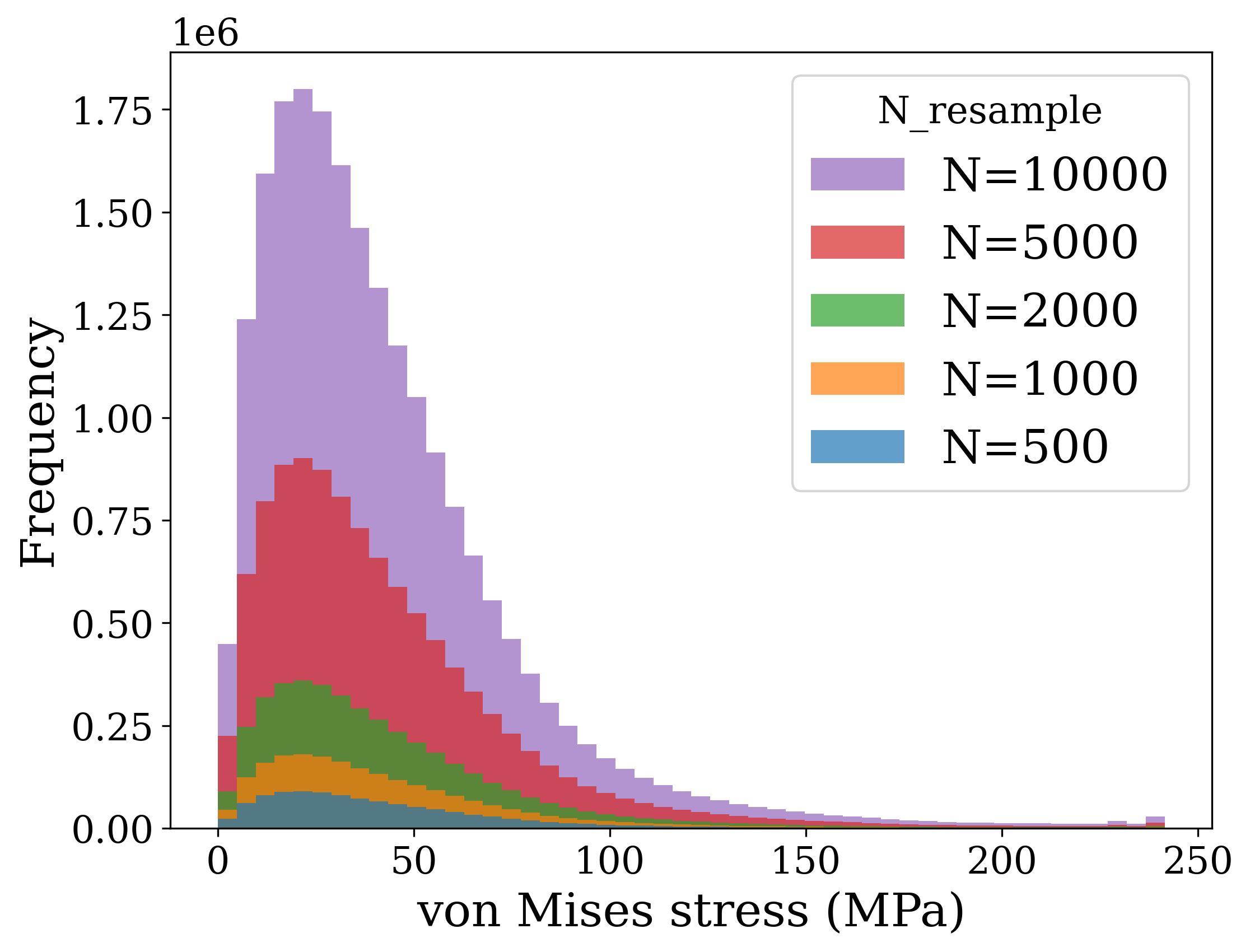}
        \caption{}
        \label{fig:figure_13_l}
    \end{subfigure}
    
    \caption{Displacement ($u_x$, $u_y$, $u_z$) and von Mises stress distributions for different resampling sizes ($N=500$, 1,000, 2,000, 5,000, 10,000) under various load cases. (a)–(d) Vertical load case: $u_x$, $u_y$, $u_z$, and von Mises stress. (e)–(h) Horizontal load case: $u_x$, $u_y$, $u_z$, and von Mises stress. (i)–(l) Diagonal load case: $u_x$, $u_y$, $u_z$, and von Mises stress.}
    \label{fig:figure_13}
\end{figure}

An important aspect of modeling complex geometries is the choice of resampling size $N$, which determines the number of points used to represent the structure in the neural network. We investigated the effect of varying $N$ on the accuracy of the predicted displacement components ($u_x$, $u_y$, $u_z$) and the von Mises stress under different loading conditions. Specifically, we considered resampling sizes of $N = 500$, 1,000, 2,000, 5,000, and 10,000 points. Figure~\ref{fig:figure_13} illustrates the distributions of these quantities for each resampling size and load case, highlighting how changes in sampling density influence the models' predictive performance.

For the vertical load case, Figures~\ref{fig:figure_13_a} to \ref{fig:figure_13_d} depict the histograms of the predicted $u_x$, $u_y$, $u_z$, and von Mises stress values across different resampling sizes. As $N$ increases, the distributions become narrower and more peaked around the mean, indicating that higher sampling densities lead to more consistent and precise predictions. This improvement can be attributed to the models' enhanced ability to capture local geometric details and stress concentrations when provided with a denser point cloud representation.

Similarly, in the horizontal load case (Figures~\ref{fig:figure_13_e} to \ref{fig:figure_13_h}), we observe that increasing $N$ results in tighter distributions for all displacement components and von Mises stress. The reduction in variance and outliers suggests that the models benefit from the additional spatial information, which allows them to better approximate the underlying physical behavior of the structure under horizontal loading.

For the diagonal load case, depicted in Figures~\ref{fig:figure_13_i} to \ref{fig:figure_13_l}, the impact of resampling size is particularly noticeable for the von Mises stress predictions. At lower sampling densities (e.g., $N = 500$), the stress distributions are broader and exhibit significant variability, indicating that the models struggle to accurately capture the complex stress patterns associated with diagonal loading when limited spatial information is available. As $N$ increases, the distributions become more concentrated, reflecting improved predictive accuracy.

\begin{figure}[!htbp]
    \centering
    \begin{subfigure}{0.4\textwidth}
        \centering
        \includegraphics[width=\linewidth]{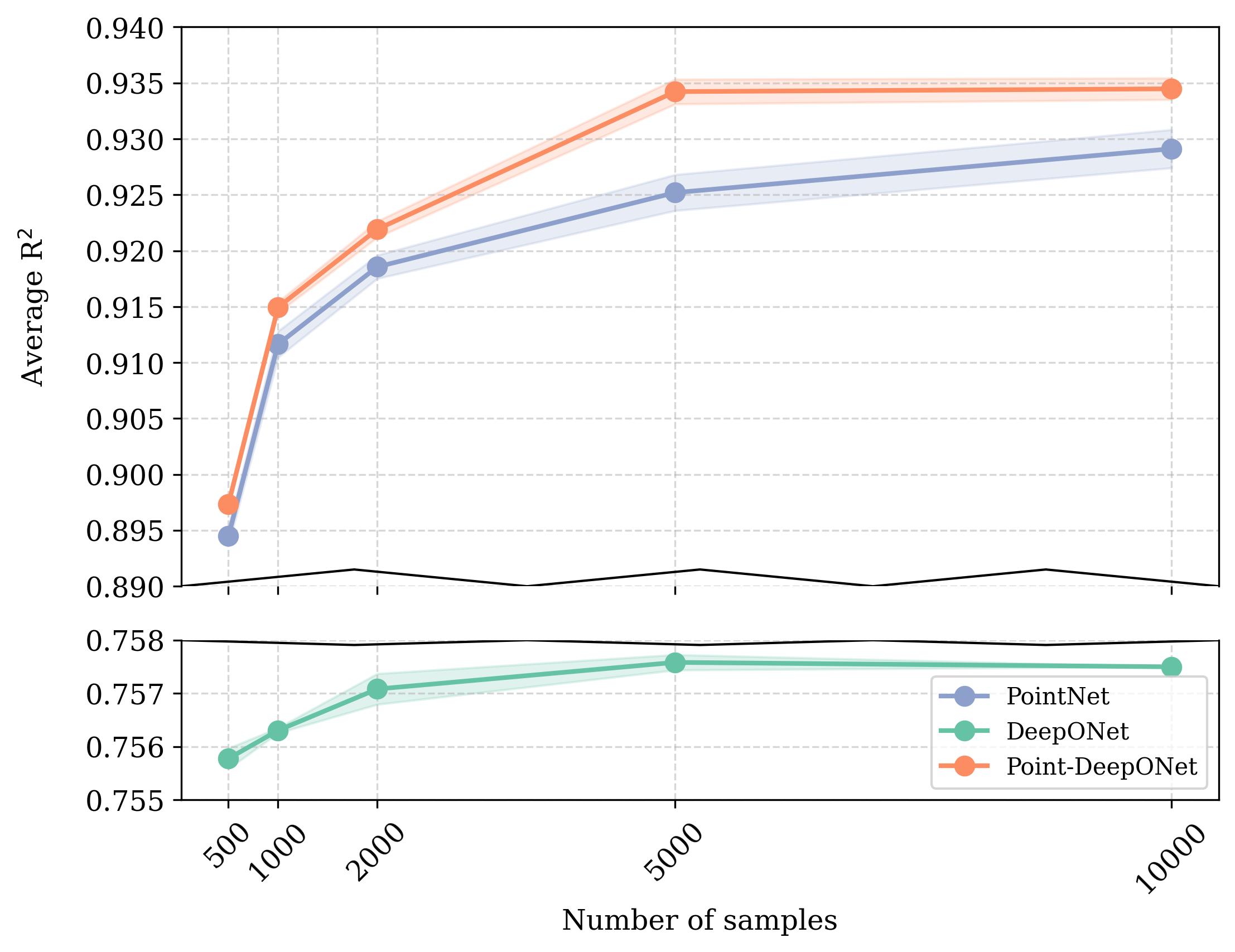}
        \caption{}
        \label{fig:figure_14_a}
    \end{subfigure}
    \begin{subfigure}{0.4\textwidth}
        \centering
        \includegraphics[width=\linewidth]{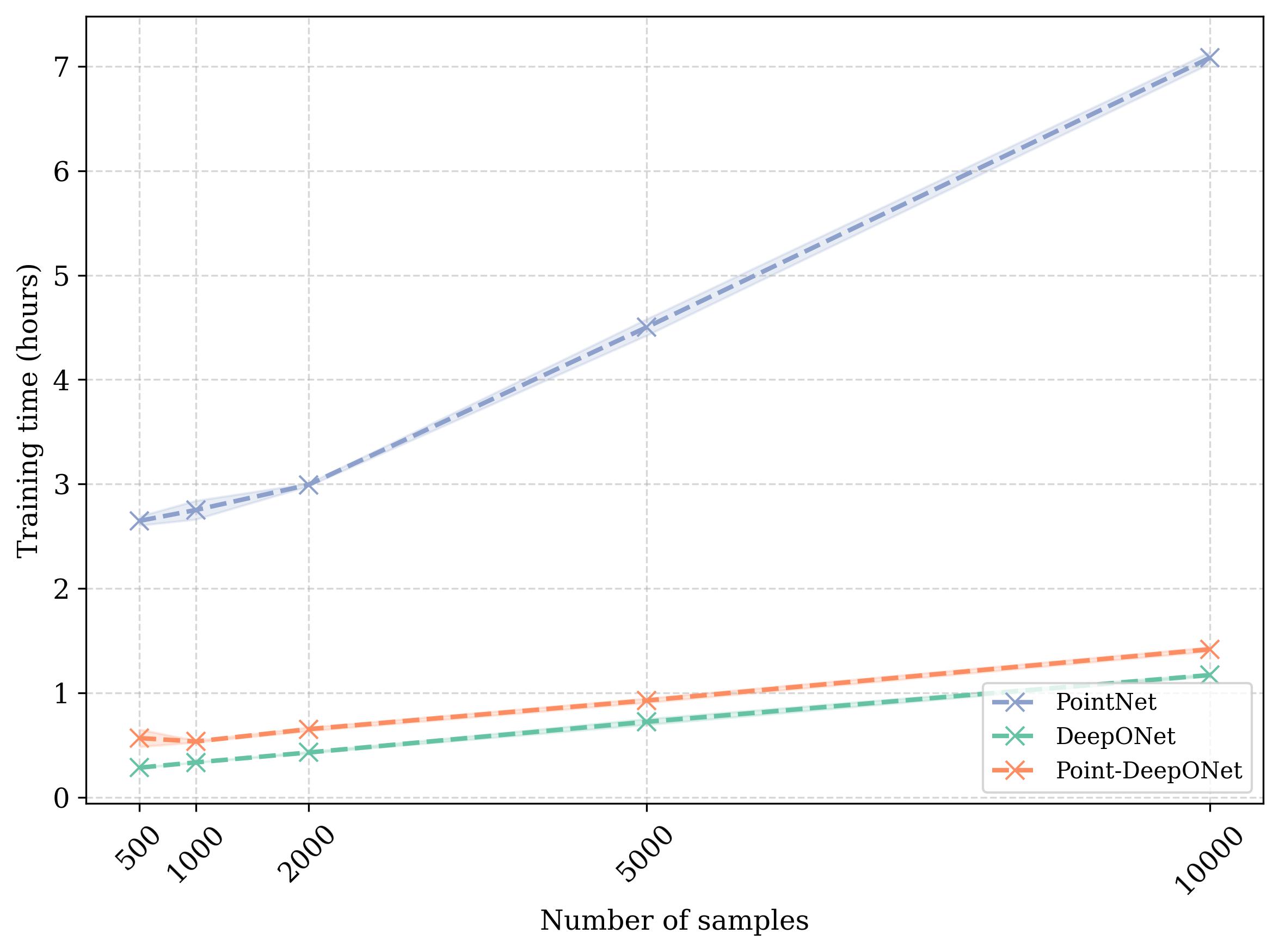}
        \caption{}
        \label{fig:figure_14_b}
    \end{subfigure}
    \caption{Model performance and training time comparison for varying sample sizes. (a) Average $R^2$ indicating accuracy increases with sample size. (b) Training time with computational costs for each model.}
    \label{fig:figure_14}
\end{figure}

To quantitatively assess the influence of resampling size, we computed the average coefficient of determination ($R^2$) for each model and resampling size, as shown in Figure~\ref{fig:figure_14_a}. The results demonstrate a clear trend of increasing $R^2$ values with larger $N$, confirming that higher sampling densities enhance the models' ability to explain the variance in the displacement and stress fields. Notably, the proposed Point-DeepONet generally achieved higher $R^2$ values compared to PointNet and DeepONet across all sampling sizes. At $N = 5{,}000$, Point-DeepONet attained an average $R^2$ of 0.934, surpassing PointNet's 0.925 and significantly outperforming DeepONet's 0.758. This indicates that Point-DeepONet provides superior prediction accuracy even when using only 2.5\% of the average nodes of the original geometries. Furthermore, for resampling sizes beyond $N = 5{,}000$, Point-DeepONet maintained an average $R^2$ of 0.93 or higher, demonstrating its effectiveness in capturing the structural responses with relatively low computational overhead.

However, this improvement in accuracy comes at the cost of increased computational resources. Figure~\ref{fig:figure_14_b} illustrates the corresponding training times for each model and resampling size. Despite the models having a similar number of parameters, PointNet consistently required the longest training times. At $N = 5{,}000$, Point-DeepONet's training time was approximately 57 minutes, which is about  13 minutes longer than DeepONet's 44.4 minutes but about 3 hours 36 minutes shorter than PointNet's 4.5 hours. Importantly, Point-DeepONet achieved higher accuracy than both models, suggesting that it offers a favorable balance between training time and predictive performance. The additional training time relative to DeepONet is justified by the significant gains in accuracy, while it is more efficient than PointNet despite delivering better results.

\begin{figure}[!h]
    \centering
    \begin{subfigure}{0.32\textwidth}
        \centering
        \includegraphics[width=\linewidth]{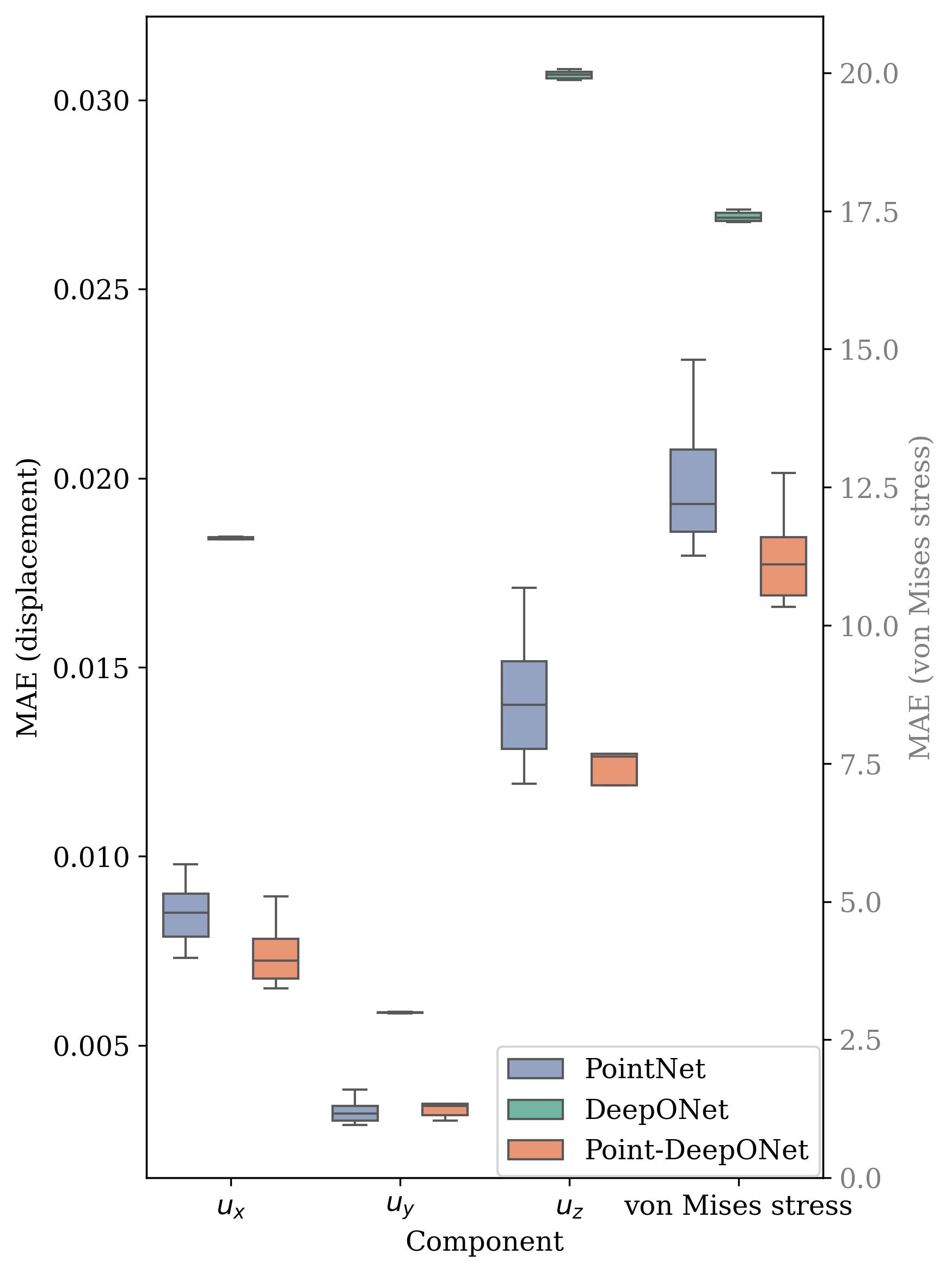}
        \caption{}
        \label{fig:figure_15_a}
    \end{subfigure}
    \begin{subfigure}{0.32\textwidth}
        \centering
        \includegraphics[width=\linewidth]{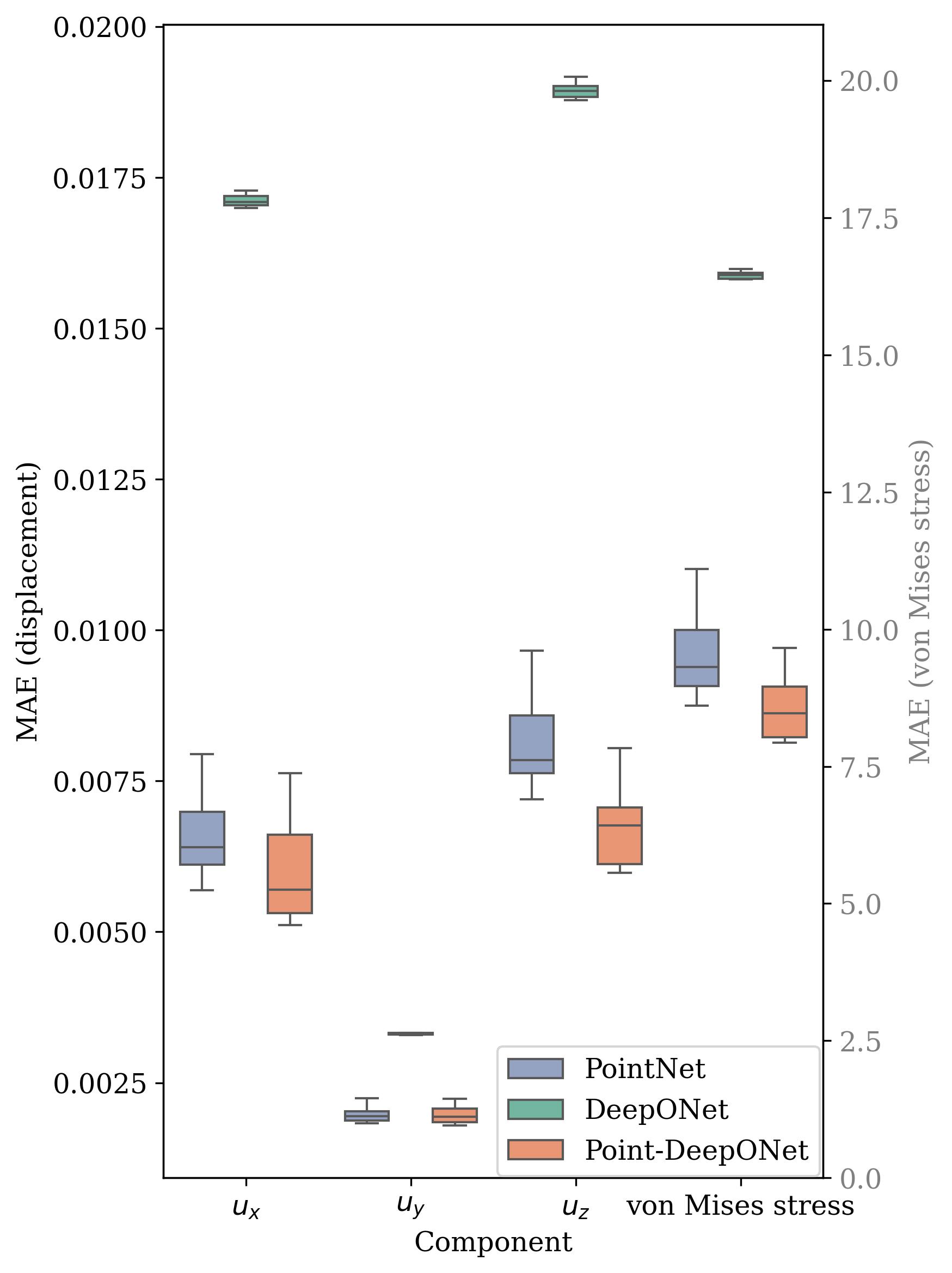}
        \caption{}
        \label{fig:figure_15_b}
    \end{subfigure}
    \begin{subfigure}{0.32\textwidth}
        \centering
        \includegraphics[width=\linewidth]{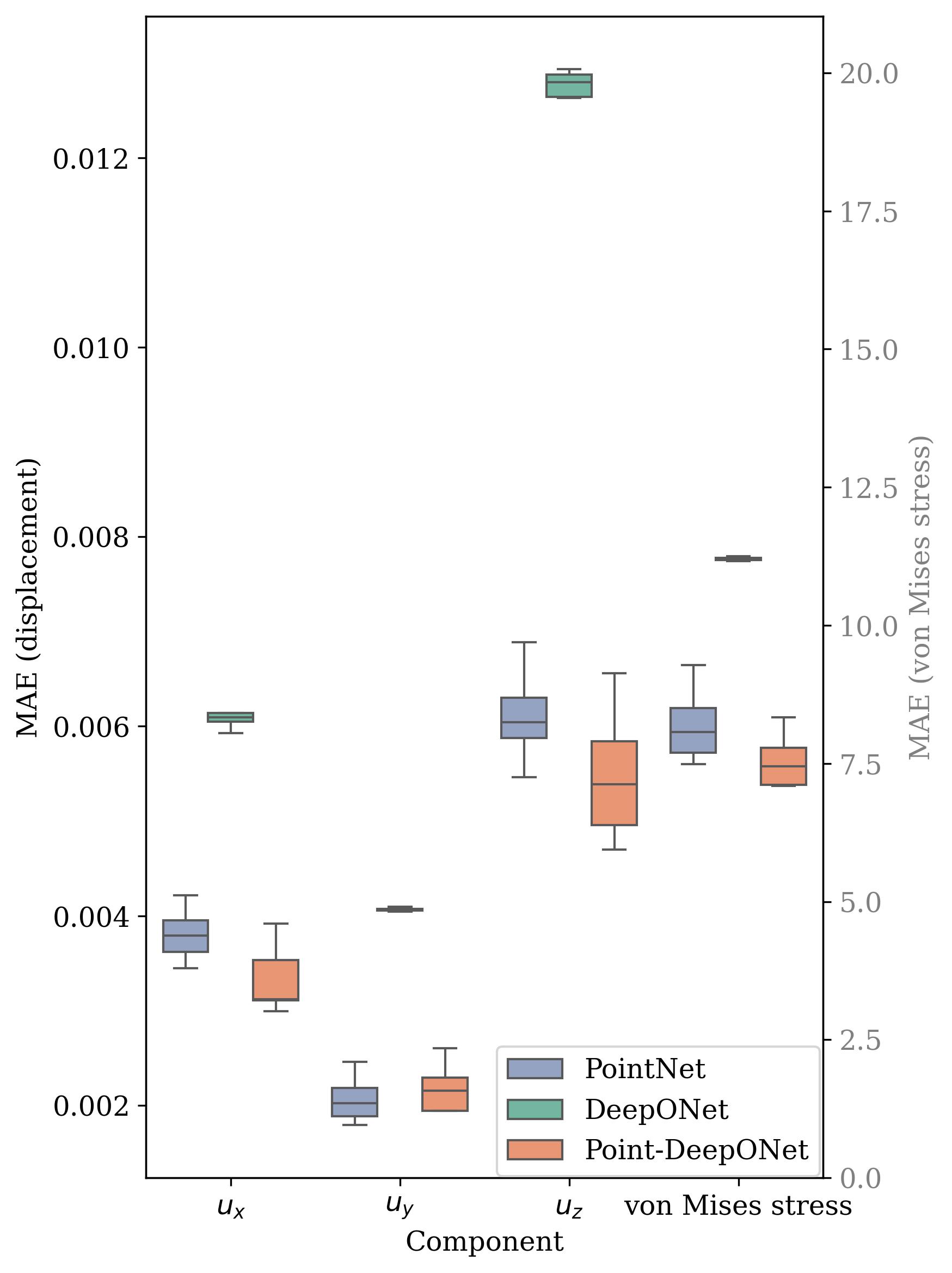}
        \caption{}
        \label{fig:figure_15_c}
    \end{subfigure}    
    \caption{Mean Absolute Error (MAE) comparison for displacement and von Mises stress components across different load directions using PointNet, DeepONet, and Point-DeepONet models: (a) Vertical load case. (b) Horizontal load case. (c) Diagonal load case. Each subplot highlights model performance across $u_x$, $u_y$, $u_z$, and von Mises stress components, illustrating how prediction accuracy varies with load cases.}
    \label{fig:figure_15}
\end{figure}

Moreover, we evaluated the models' performance across different load cases by comparing the mean absolute errors (MAEs) for each displacement component and von Mises stress, as presented in Figure~\ref{fig:figure_15}. The proposed Point-DeepONet consistently outperformed PointNet and DeepONet across all load cases and variables, achieving lower MAE values. This superior performance can be attributed to the integration of geometric encoding and operator learning within Point-DeepONet, which enables it to capture complex spatial relationships and nonlinear material behavior more effectively.

\subsection{Influence of Dataset Size on Model Performance}
\label{subsec:influence_dataset}

In this subsection, we examine how the size of the dataset influences the performance and training time of the PointNet and Point-DeepONet models. Given that DeepONet exhibited lower predictive performance in the previous analysis, we focus here on comparing PointNet and Point-DeepONet as the dataset size varies.

\begin{figure}[h]
    \centering
    \begin{subfigure}[b]{0.32\textwidth}
        \centering
        \includegraphics[width=\textwidth]{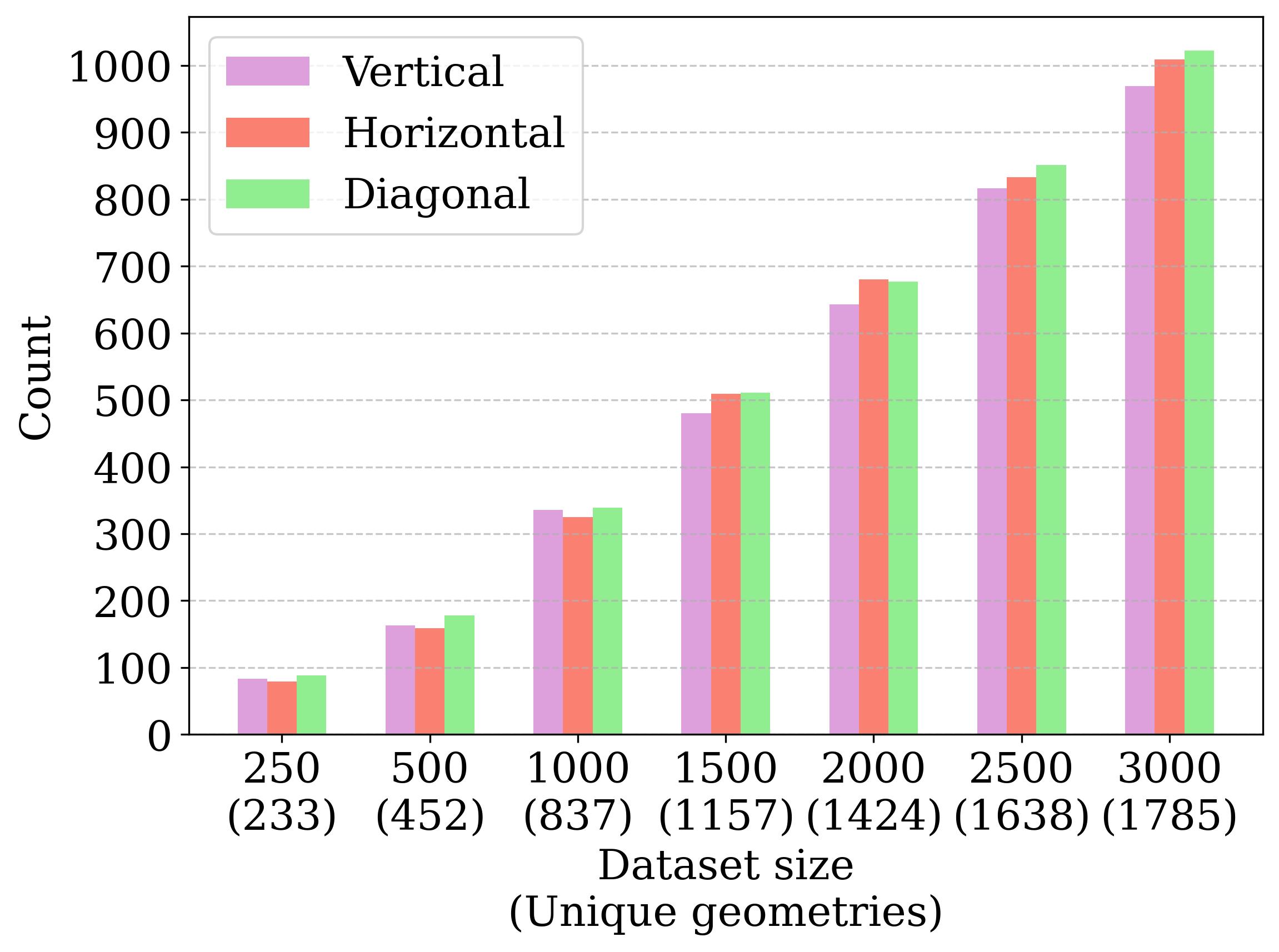}
        \caption{}
        \label{fig:figure_16_a}
    \end{subfigure}
    \hfill
    \begin{subfigure}[b]{0.32\textwidth}
        \centering
        \includegraphics[width=\textwidth]{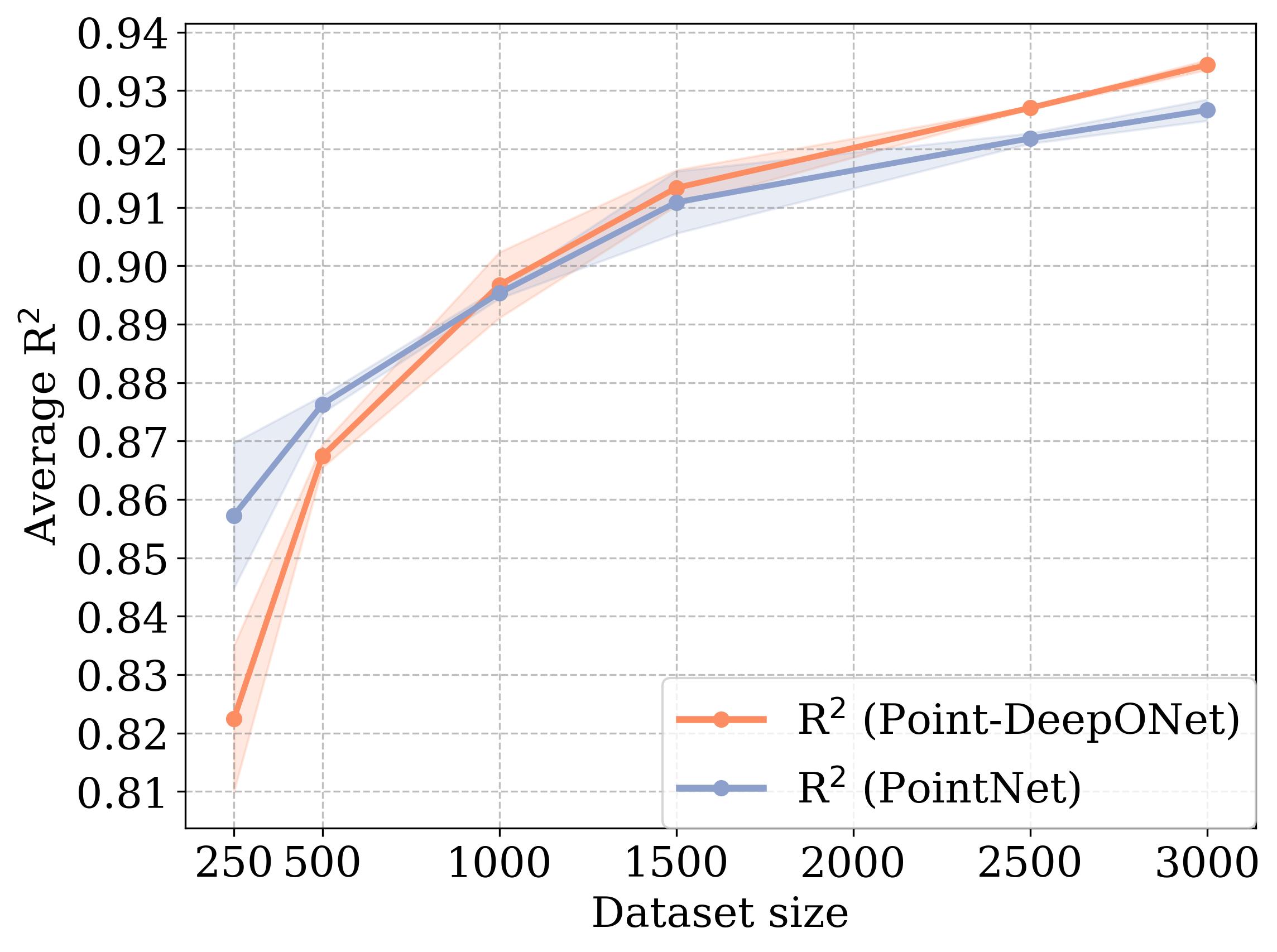}
        \caption{}
        \label{fig:figure_16_b}
    \end{subfigure}
    \hfill
    \begin{subfigure}[b]{0.32\textwidth}
        \centering
        \includegraphics[width=\textwidth]{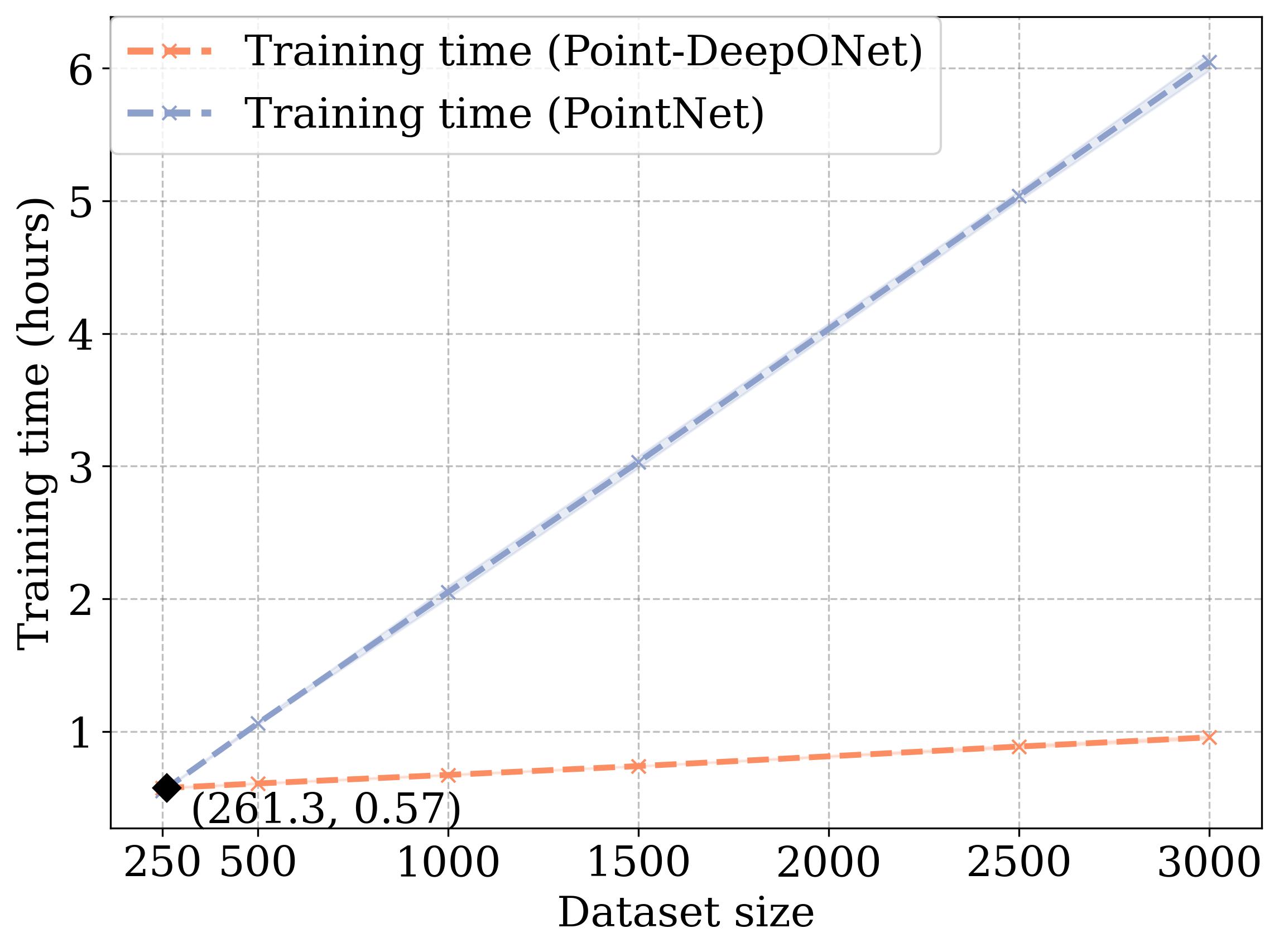}
        \caption{}
        \label{fig:figure_16_c}
    \end{subfigure}
    \caption{
        (a) Visualization of dataset sizes by load case direction (Vertical, Horizontal, Diagonal) across different dataset sizes. The numbers in parentheses denote the counts of unique geometries. (b) Average $R^2$ values for $u_x$, $u_y$, $u_z$ displacements, and von Mises stress across dataset sizes for Point-DeepONet and PointNet models. (c) Training time for each model as a function of dataset size, indicating that Point-DeepONet has a slower increase in training time compared to PointNet.
    }
    \label{fig:figure_16}
\end{figure}

Figure~\ref{fig:figure_16} illustrates the dataset composition, model performance, and training time as functions of dataset size. In Figure~\ref{fig:figure_16_a}, the distribution of the dataset across load case directions—Vertical, Horizontal, and Diagonal—is presented for different dataset sizes ranging from 250 to 3,000 samples. The detailed counts for each load case and the total duplicates removed are summarized in Table~\ref{tab:dataset_distribution}.

\begin{table}[h]
    \centering
    \caption{Dataset distribution across load case directions and dataset sizes, including the number of unique geometries (duplicates removed).}
    \label{tab:dataset_distribution}
    \begin{tabularx}{\linewidth}{Xcccc}
        \toprule
        Dataset size & 
        Vertical & 
        Horizontal & 
        Diagonal & 
        Total duplicates removed \\
        \midrule
        250   & 83  & 79  & 88   & 17   \\
        500   & 163 & 159 & 178  & 48   \\
        1,000 & 336 & 325 & 339  & 163  \\
        1,500 & 480 & 509 & 511  & 343  \\
        2,500 & 816 & 833 & 851  & 862  \\
        3,000 & 969 & 1,009 & 1,022 & 1,215 \\
        \bottomrule
    \end{tabularx}
\end{table}

These counts reflect the increasing diversity and complexity of the dataset as more samples are included, ensuring that the models are trained on a wide variety of geometries and loading conditions.

Figure~\ref{fig:figure_16_b} compares the model performance in terms of the average coefficient of determination ($R^2$) for $u_x$, $u_y$, $u_z$ displacements, and von Mises stress across different dataset sizes. The data indicates that when the dataset size is less than or equal to 1,000 samples, PointNet achieves higher average $R^2$ values than Point-DeepONet. Specifically, at a dataset size of 1,000, PointNet attains an average $R^2$ of 0.895, while Point-DeepONet reaches 0.897. However, beyond a dataset size of 1,000, Point-DeepONet begins to outperform PointNet. For instance, at 2,500 samples, Point-DeepONet achieves an average $R^2$ of 0.927, surpassing PointNet's 0.922. At the maximum dataset size of 3,000 samples, Point-DeepONet attains an average $R^2$ of 0.934, higher than PointNet's 0.927. This trend suggests that Point-DeepONet benefits more from larger datasets, likely due to its enhanced ability to capture complex relationships in the data.

In Figure~\ref{fig:figure_16_c}, the training time for each model as a function of dataset size is presented. Notably, PointNet exhibits a steeper increase in training time with increasing dataset size compared to Point-DeepONet. For PointNet, the training time increases from approximately 0.53 hours at 250 samples to about 6.05 hours at 3,000 samples. In contrast, Point-DeepONet's training time increases from approximately 0.55 hours at 250 samples to about 0.95 hours at 3,000 samples. The reduced slope in training time for Point-DeepONet suggests that its computational complexity increases more slowly with dataset size compared to PointNet.

The intersection point where both models have equal training times occurs at a dataset size of approximately 261 samples, with a training time of around 0.57 hours. Beyond this dataset size, Point-DeepONet not only requires less training time than PointNet but also delivers superior predictive performance, as evidenced by higher $R^2$ values. This indicates that Point-DeepONet scales more efficiently with larger datasets and becomes more advantageous in both accuracy and computational cost when training on larger data volumes.

Overall, the analysis demonstrates that Point-DeepONet outperforms PointNet in terms of both accuracy and training efficiency as the dataset size increases beyond 261 samples. The scalability of Point-DeepONet makes it a more suitable choice for large-scale datasets in structural mechanics applications, where training time and computational resources are critical considerations.

\subsection{Ablation Study and Architectural Validation}
\label{subsec:ablation_input_configs}
To validate that our proposed architecture is a deliberately designed framework rather than a simple stacking of modules, we analyze the performance under various input and architectural configurations. This serves a dual purpose: first, to determine the optimal set of inputs, and second, to perform an ablation study that validates the contribution of key architectural components.

Table~\ref{tab:model_metrics_comparison} presents a comprehensive performance comparison across different input configurations. The results for Point-DeepONet, when compared row by row, effectively serve as an ablation study for the input features. For instance, comparing the configuration with `Trunk: X, SDF` to the one with `Trunk: X` highlights the significant performance gain from including the signed distance function, which provides critical geometric context to the model. Overall, the configuration using all available inputs—mass, force, and direction in the branch, and coordinates with SDF in the trunk—yielded the best performance, confirming that enriching the model with both physical and detailed geometric information is crucial for accuracy.

To specifically isolate and validate the architectural contribution of the sinusoidal representation network (SIREN) layers and the fusion mechanism, we conducted additional focused experiments. First, we compared our full Point-DeepONet model against a variant where the SIREN activations in the trunk network were replaced with standard SiLU activations. The results, summarized in Table~\ref{tab:siren_ablation}, demonstrate the clear advantage of using SIREN. The superior performance of the SIREN-based model underscores its effectiveness in capturing the high-frequency details inherent in complex stress and displacement fields, a task where standard activation functions often fall short.

Second, we investigated the impact of the feature fusion strategy. Our proposed model uses an element-wise multiplication to combine the branch and trunk network outputs, allowing the global context to modulate the local features. We compared this against the standard dot product fusion used in the original DeepONet. As shown in Table~\ref{tab:fusion_ablation}, the element-wise multiplication approach yielded a notable improvement in the average $R^2$ score. This suggests that the richer, point-wise interaction enabled by element-wise multiplication is more effective for capturing complex field responses than the single aggregated value produced by a dot product. These analyses, combined with the input configuration tests, confirm that the proposed Point-DeepONet architecture is not an arbitrary combination of modules but a carefully designed framework where each component plays a vital role.

\begin{table}[!h]
    \centering
    \caption{Ablation study on the effect of SIREN activations. The table shows the average $R^2$ score across all displacement and stress components for the best-performing input configuration.}
    \label{tab:siren_ablation}
    \begin{tabular}{lc}
        \toprule
        Model Variant & Average $R^2$ Score \\
        \midrule
        \textbf{Point-DeepONet (with SIREN)} & \textbf{0.934} \\
        Point-DeepONet (with SiLU instead of SIREN) & 0.905 \\
        \bottomrule
    \end{tabular}
\end{table}

\begin{table}[!h]
    \centering
    \caption{Ablation study on the feature fusion mechanism. The table shows the average $R^2$ score for the best-performing input configuration.}
    \label{tab:fusion_ablation}
    \begin{tabular}{lc}
        \toprule
        Fusion Method & Average $R^2$ Score \\
        \midrule
        \textbf{Element-wise Multiplication (Proposed)} & \textbf{0.934} \\
        Dot Product & 0.912 \\
        \bottomrule
    \end{tabular}
\end{table}

\begin{table}[!h]
    \caption{Performance comparison of PointNet, DeepONet, and Point-DeepONet models across various input configurations for displacement and von Mises stress metrics.}
    \label{tab:model_metrics_comparison}
    \centering
    \scriptsize
    \setlength{\tabcolsep}{2pt}
    \begin{adjustbox}{width=\textwidth}
        \begin{threeparttable}
            \begin{tabular}{@{}lcc|rrr|rrr|rrr|rrr@{}}
                \toprule
                \multirow{2}{*}{Model} & \multirow{2}{*}{Input Parameters} & \multirow{2}{*}{Component} & \multicolumn{3}{c|}{$u_x$ (mm)} & \multicolumn{3}{c|}{$u_y$ (mm)} & \multicolumn{3}{c|}{$u_z$ (mm)} & \multicolumn{3}{c}{von Mises stress (MPa)} \\
                & & & MAE & RMSE & $R^2$ & MAE & RMSE & $R^2$ & MAE & RMSE & $R^2$ & MAE & RMSE & $R^2$ \\
                \midrule
                \multirow{6}{*}{PointNet} 
                & \multirow{3}{*}{$X$, $f$, $\mathbf{d}$}
                & Vertical   & 0.008 & 0.018 & 0.950 & 0.003 & 0.006 & 0.920 & 0.014 & 0.034 & 0.951 & 12.111 & 18.671 & 0.860 \\
                &   & Horizontal & 0.006 & 0.011 & 0.982 & 0.002 & 0.003 & 0.876 & 0.007 & 0.012 & 0.982 & 9.042 & 14.183 & 0.904 \\
                &   & Diagonal   & 0.004 & 0.005 & 0.884 & 0.002 & 0.003 & 0.960 & 0.006 & 0.009 & 0.968 & 8.224 & 12.406 & 0.844 \\
                \cmidrule{2-15}
                & \multirow{3}{*}{$X$, $m$, $f$, $\mathbf{d}$}
                & Vertical   & 0.008 & 0.018 & 0.953 & 0.003 & 0.006 & \textbf{0.927} & 0.013 & 0.034 & 0.952 & 11.666 & 18.291 & 0.866 \\
                &   & Horizontal & 0.006 & 0.010 & 0.983 & 0.002 & 0.003 & 0.878 & 0.008 & 0.012 & 0.980 & 9.073 & 14.208 & 0.904 \\
                &   & Diagonal   & 0.004 & 0.005 & 0.884 & 0.002 & 0.003 & 0.963 & 0.006 & 0.008 & 0.975 & 7.639 & 11.786 & 0.859 \\
                \midrule
                \multirow{9}{*}{DeepONet} 
                & \multirow{3}{*}{\makecell[c]{Branch: $f$, $\mathbf{d}$ \\ Trunk: $X$, $SDF$}} 
                & Vertical   & 0.026 & 0.051 & 0.625 & 0.007 & 0.012 & 0.709 & 0.048 & 0.090 & 0.653 & 21.893 & 31.979 & 0.590 \\
                &   & Horizontal & 0.024 & 0.044 & 0.692 & 0.004 & 0.006 & 0.503 & 0.026 & 0.047 & 0.700 & 20.187 & 29.984 & 0.570 \\
                &   & Diagonal   & 0.006 & 0.010 & 0.578 & 0.005 & 0.007 & 0.752 & 0.016 & 0.024 & 0.773 & 12.607 & 18.743 & 0.644 \\
                \cmidrule{2-15}
                & \multirow{3}{*}{\makecell[c]{Branch: $m$, $f$, $\mathbf{d}$ \\ Trunk: $X$}} 
                & Vertical   & 0.019 & 0.037 & 0.795 & 0.006 & 0.010 & 0.775 & 0.032 & 0.061 & 0.841 & 19.081 & 28.328 & 0.678 \\
                &   & Horizontal & 0.017 & 0.033 & 0.835 & 0.003 & 0.005 & 0.561 & 0.019 & 0.035 & 0.835 & 17.867 & 26.777 & 0.657 \\
                &   & Diagonal   & 0.006 & 0.010 & 0.620 & 0.004 & 0.007 & 0.787 & 0.013 & 0.020 & 0.854 & 11.930 & 17.973 & 0.672 \\
                \cmidrule{2-15}
                & \multirow{3}{*}{\makecell[c]{Branch: $m$, $f$, $\mathbf{d}$ \\ Trunk: $X$, $SDF$}} 
                & Vertical   & 0.026 & 0.050 & 0.626 & 0.007 & 0.012 & 0.709 & 0.048 & 0.090 & 0.654 & 21.842 & 31.901 & 0.592 \\
                &   & Horizontal & 0.024 & 0.044 & 0.694 & 0.004 & 0.006 & 0.503 & 0.026 & 0.047 & 0.699 & 20.091 & 29.935 & 0.572 \\
                &   & Diagonal   & 0.006 & 0.010 & 0.578 & 0.005 & 0.007 & 0.753 & 0.016 & 0.024 & 0.774 & 12.555 & 18.710 & 0.645 \\
                \midrule
                \multirow{9}{*}{\makecell{Point-DeepONet \\ (Our proposed)}} 
                & \multirow{3}{*}{\makecell[c]{Branch: $f$, $\mathbf{d}$ \\ Trunk: $X$, $SDF$}}
                & Vertical   & 0.007 & 0.017 & 0.959 & 0.003 & 0.006 & 0.919 & 0.032 & 0.058 & 0.958 & 10.538 & 17.043 & 0.883 \\
                &   & Horizontal & 0.005 & 0.010 & 0.983 & 0.002 & 0.003 & 0.876 & 0.006 & 0.011 & 0.983 & 8.078 & 12.976 & 0.920 \\
                &   & Diagonal   & 0.003 & 0.005 & 0.905 & 0.002 & 0.003 & 0.950 & 0.005 & 0.009 & 0.972 & 7.180 & 11.468 & 0.867 \\
                \cmidrule{2-15}
                & \multirow{3}{*}{\makecell[c]{Branch: $m$, $f$, $\mathbf{d}$ \\ Trunk: $X$}}
                & Vertical   & 0.007 & 0.017 & 0.958 & 0.003 & 0.006 & 0.920 & 0.012 & 0.033 & \textbf{0.982} & 8.967 & 14.347 & \textbf{0.902} \\
                &   & Horizontal & 0.002 & 0.003 & 0.855 & 0.006 & 0.011 & \textbf{0.982} & 0.008 & 0.012 & 0.980 & 9.073 & 14.208 & 0.904 \\
                &   & Diagonal   & 0.003 & 0.005 & \textbf{0.911} & 0.002 & 0.003 & 0.953 & 0.005 & 0.008 & 0.977 & 7.661 & 12.028 & 0.853 \\
                \cmidrule{2-15}
                & \multirow{3}{*}{\makecell[c]{Branch: $m$, $f$, $\mathbf{d}$ \\ Trunk: $X$, $SDF$}}
                & Vertical   & 0.007 & 0.016 & \textbf{0.964} & 0.003 & 0.006 & 0.923 & 0.012 & 0.030 & 0.961 & 10.541 & 17.148 & 0.882 \\
                &   & Horizontal & 0.005 & 0.010 & \textbf{0.985} & 0.002 & 0.003 & 0.875 & 0.006 & 0.011 & \textbf{0.985} & 7.935 & 12.684 & \textbf{0.923} \\
                &   & Diagonal   & 0.003 & 0.005 & 0.903 & 0.002 & 0.003 & \textbf{0.953} & 0.005 & 0.008 & \textbf{0.978} & 7.090 & 11.208 & \textbf{0.873} \\
                \bottomrule
            \end{tabular}
            \begin{tablenotes}
                \footnotesize
                \item *Input Parameters: $X$: coordinates $(x, y, z)$; $SDF$: signed distance function; $m$: mass; $f$: force magnitude; $\mathbf{d}$: direction vector $(d_x, d_y, d_z)$.
            \end{tablenotes}
        \end{threeparttable}
    \end{adjustbox}
\end{table}

\subsection{Validation of Generalization to Unseen Load Conditions}
\label{subsec:random_load_validation}
To rigorously evaluate the model's ability to generalize across both load directions and magnitudes, we conducted a targeted validation study. A new test set was generated using 20 distinct geometries, a subset of which are shown with IDs in Figure~\ref{fig:figure_4}. For these geometries, we created 50 test cases where load directions were sampled uniformly at random from 0 to 90 degrees, and magnitudes were sampled from a range of 35-45 kN, ensuring the conditions were unseen during training. The Point-DeepONet model, trained on the original large-scale dataset, was then evaluated on these 50 new cases without any fine-tuning. The overall performance, summarized in Table~\ref{tab:random_load_test}, shows only a minor degradation compared to the original validation set, confirming robust generalization.

\begin{figure*}[!htbp] 
    \centering
    \begin{subfigure}{0.33\textwidth}
        \centering
        \includegraphics[width=\linewidth]{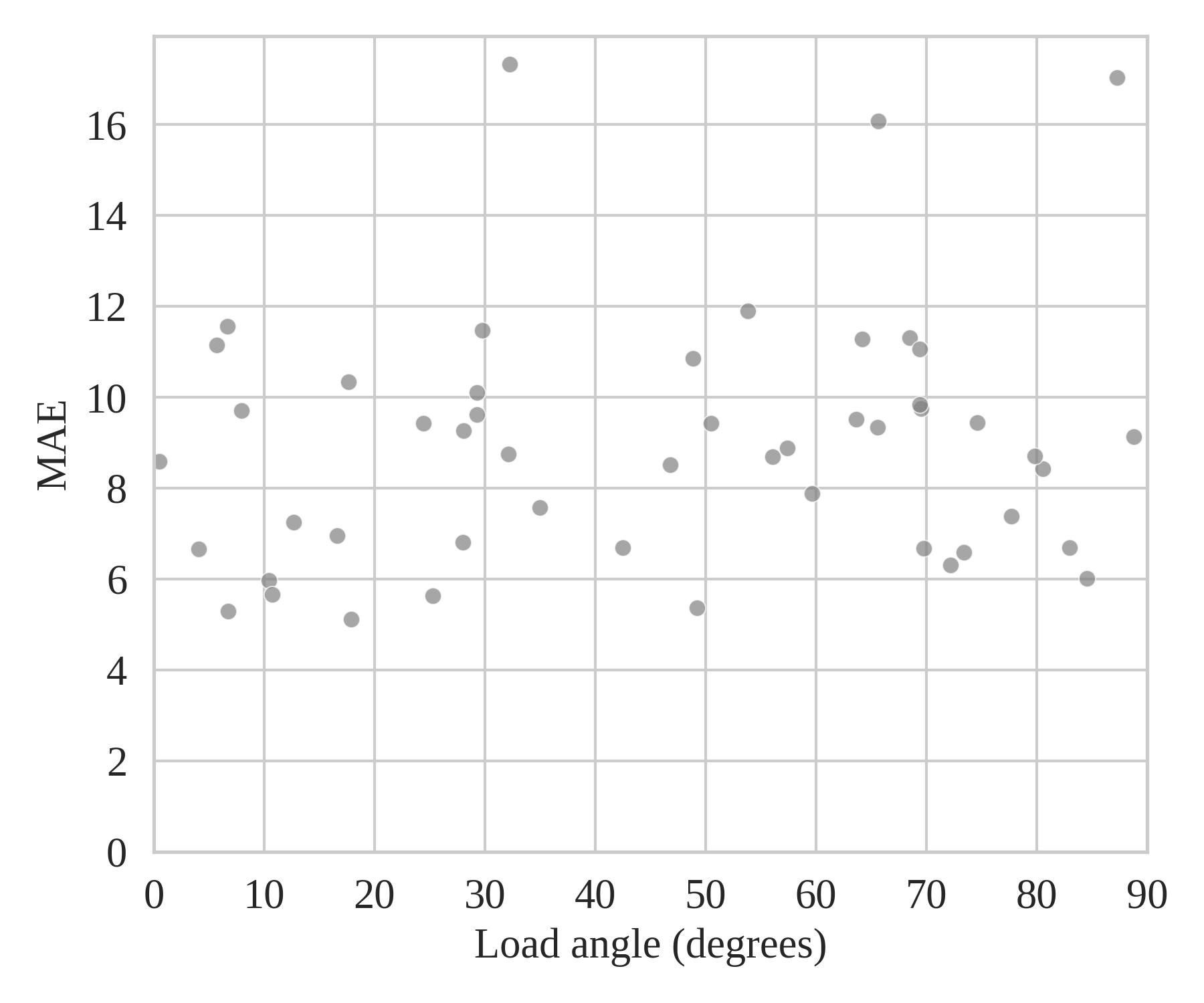} 
        \captionsetup{font=small}
        \caption{}
    \end{subfigure}
    \hfill
    \begin{subfigure}{0.33\textwidth}
        \centering
        \includegraphics[width=\linewidth]{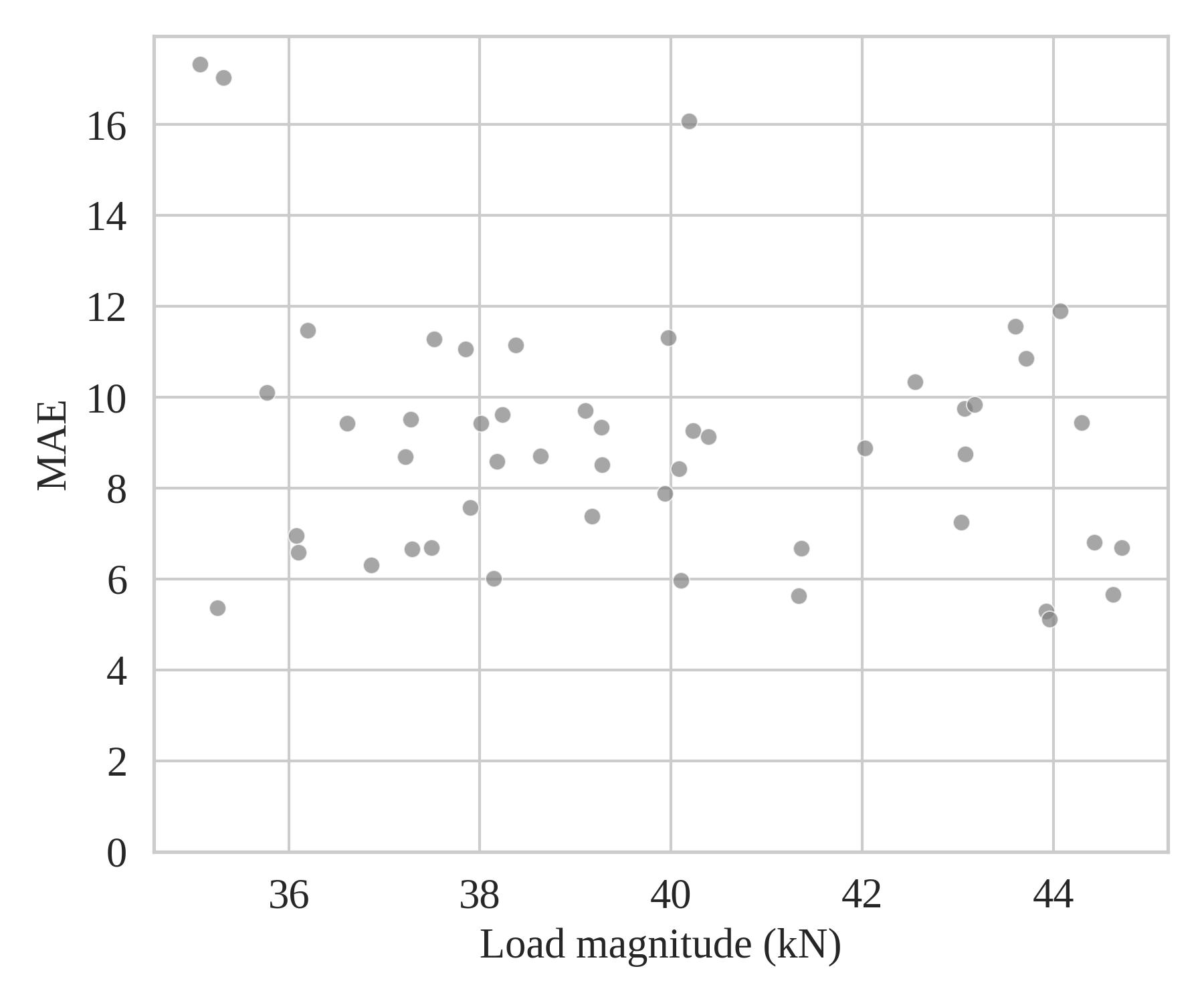} 
        \captionsetup{font=small}
        \caption{}
    \end{subfigure}
    \hfill
    \begin{subfigure}{0.33\textwidth}
        \centering
        \includegraphics[width=\linewidth]{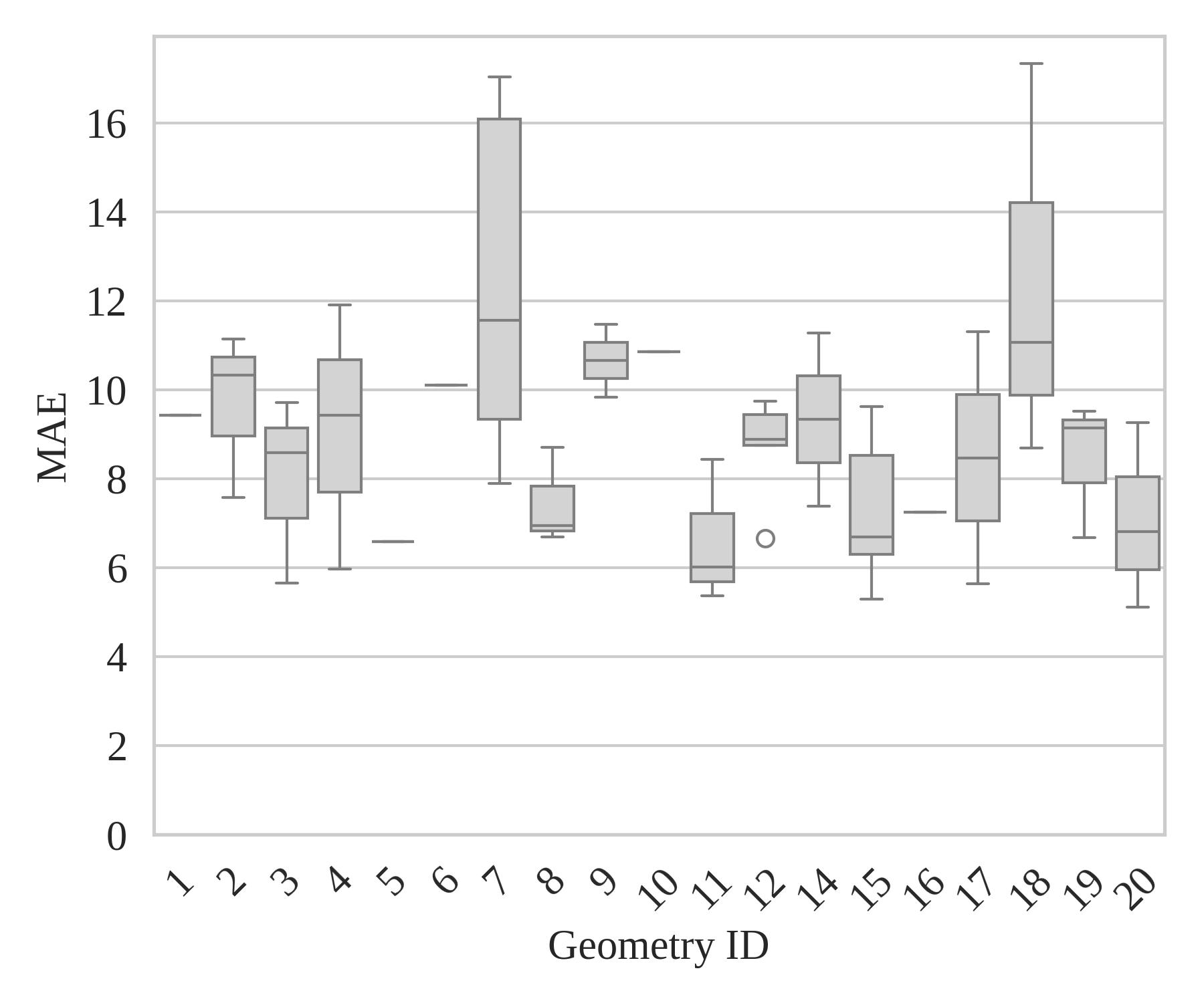} 
        \captionsetup{font=small}
        \caption{}
    \end{subfigure}
    
    \caption{Detailed analysis of error trends on the random load test set. (a) MAE vs. load angle shows no clear correlation, indicating good directional generalization. (b) MAE vs. load magnitude is also randomly distributed. (c) Box plot of MAE per geometry reveals that specific geometries (e.g., ID 7, 18) consistently produce higher errors, highlighting a limitation related to geometric complexity.}
    \label{fig:error_analysis_trends}
\end{figure*}

To further dissect the model's performance and identify its limitations, we analyzed the error trends from these 50 random test cases, as shown in Figure~\ref{fig:error_analysis_trends}. The scatter plot of error versus load angle (Figure~\ref{fig:error_analysis_trends}a) reveals no discernible correlation, suggesting that the model's accuracy is largely independent of the specific load direction. This provides strong evidence that the model has successfully generalized across the entire directional space. Similarly, the relationship between error and load magnitude (Figure~\ref{fig:error_analysis_trends}b) appears random, indicating uniform performance across the tested magnitude range.

In contrast, the box plot of errors per geometry (Figure~\ref{fig:error_analysis_trends}c) highlights that certain geometries (e.g., ID 7, 18) consistently yield higher prediction errors regardless of the load condition. A qualitative review of these "difficult" geometries, by referencing them back to Figure~\ref{fig:figure_4}, reveals they possess challenging features. For example, Geometry 7 has a deep, sharp U-shaped groove known to cause stress concentration, while Geometry 18 features complex intersections of thin reinforcing members. This finding points to a key limitation of our current approach: while robust to load variations, its accuracy can be sensitive to specific geometric complexities where high-frequency stress concentrations are likely to occur. This analysis provides valuable insight for future model improvements.

\begin{table}[h]
    \centering
    \caption{Performance on the validation set (Fixed Directions) vs. the new test set (Random Directions).}
    \label{tab:random_load_test}
    \begin{tabular}{lcc}
        \toprule
        Metric & Validation Set (Fixed Dir.) & Random-Direction Test Set \\
        \midrule
        Avg. $R^2$ (Displacement) & 0.970 & 0.938 \\
        Avg. $R^2$ (von Mises)   & 0.898 & 0.847 \\
        \bottomrule
    \end{tabular}
\end{table}

\section{Conclusion and Future Work}
\label{sec:Conclusion_futurework}

This study introduced \textit{Point-DeepONet}, a novel neural network architecture that integrates PointNet into the DeepONet framework to overcome key challenges in nonlinear elastoplastic analyses of non-parametric, three-dimensional geometries. Our results highlight how the method advances beyond existing surrogate modeling approaches by enabling accurate, efficient, and scalable predictions directly on unstructured finite element meshes. More specifically, our findings support the following conclusions:

\begin{enumerate}
    \item \textbf{Flexible handling of non-parametric 3D geometries:} 
    By leveraging PointNet's capability to process raw point cloud data, \textit{Point-DeepONet} effectively modelled a broad range of complex geometries without any parameterization or regular grid mapping. With a resampling size of $5{,}000$ points (only 2.5\% of the average mesh), our model achieved an average $R^2$ of $0.934$, thereby demonstrating the ability to preserve geometric fidelity even at significantly reduced input sizes.

    \item \textbf{Versatile adaptation to varying load conditions:} 
    Unlike methods limited to load magnitude variations, \textit{Point-DeepONet} seamlessly accommodates changes in both load magnitude and direction. Our extensive validation, including tests on unseen interpolated and completely random load directions, confirmed the model's strong generalization capability, maintaining high accuracy without retraining.

    \item \textbf{High-fidelity field predictions at the source mesh:} 
    \textit{Point-DeepONet} provides direct predictions of displacement and von Mises stress on the original finite element mesh, eliminating the need for interpolation. The model delivered robust high-resolution field outputs that closely match the FEA results across a wide range of conditions.

    \item \textbf{Reduced computational overhead:} 
    Comparisons with conventional nonlinear FEA highlight significant computational savings. While the average FEA runtime exceeded 19 minutes per case, \textit{Point-DeepONet} required only a few seconds for full-mesh inference, offering a more than $4 \times 10^2$-fold speedup. Our quantitative analysis also confirmed that the one-time preprocessing cost for SDF generation is negligible, preserving the model's overall efficiency for iterative applications.

    \item \textbf{Accelerated decision-making processes:} 
    Owing to its rapid inference and high predictive accuracy, \textit{Point-DeepONet} facilitates quick evaluations of numerous design iterations. This advantage was evident when scaling up the dataset size, significantly expediting the decision-making process in engineering applications.

    \item \textbf{Robust scalability with increasing data and complexity:} 
    The model's performance was validated on a comprehensive dataset of 3,000 simulations. The results from this dataset, combined with ablation studies and targeted generalization tests, confirm that the method is robust and that its key architectural components are essential for achieving high accuracy.
\end{enumerate}

Building on these accomplishments, future research can extend \textit{Point-DeepONet} to encompass a wider spectrum of material models, including anisotropic and composite materials. While the current study focuses on static loads, the architectural design naturally extends to time-dependent conditions. The branch network can incorporate temporal information through recurrent layers (LSTM/GRU), sequential operator frameworks such as s-DeepONet \citep{he2024sequential}, or attention mechanisms, enabling transient response predictions under seismic, cyclic, or impact loading. Integrating physics-informed neural networks (PINNs) could further enhance model robustness by embedding physical constraints directly into the loss function, potentially reducing data requirements and increasing accuracy in sparse data regimes. Furthermore, our error analysis indicates that model accuracy is sensitive to certain geometric complexities. Future work could address this limitation by incorporating adaptive point sampling strategies that focus on regions with high geometric gradients or by exploring more advanced geometric feature extractors capable of capturing fine-grained local details more effectively. Additionally, exploring transfer learning techniques may enable the model to adapt to new geometries and load conditions with minimal retraining, further improving computational efficiency.

In summary, \textit{Point-DeepONet} represents a significant step forward in operator learning-based surrogate modeling. Its demonstrated accuracy, computational efficiency, flexibility, and scalability make it well-suited for real-time structural assessment, design optimization, uncertainty quantification, and other engineering scenarios that demand reliable, high-fidelity predictions across diverse and evolving conditions.

\section*{Acknowledgments}
This work was supported by the Ministry of Science and ICT of Koreagrant (GTL24031-000, N10250154) and the Ministry of Trade, Industry\& Energy (RS-2024-00410810, RS-2025-02317327)

\section*{Conflict of interest statement}
The authors declared no potential conflicts of interest.

\section*{Replication of results}
The data and source code that support the findings of this study can be found at \href{https://github.com/jangseop-park/Point-DeepONet}{https://github.com/jangseop-park/Point-DeepONet}.

\bibliographystyle{elsarticle-num-names} 
\bibliography{references}

\end{document}